\documentclass[a4paper,fleqn]{cas-dc}

\usepackage{lipsum}
\usepackage{microtype}

\usepackage{tabularx}
\usepackage[flushleft]{threeparttable}
\usepackage{graphicx}
\usepackage[T1]{fontenc}

\usepackage[ruled,vlined,linesnumbered]{algorithm2e}
\SetAlFnt{\footnotesize}\SetAlCapFnt{\footnotesize}\SetAlCapNameFnt{\footnotesize}
\SetCommentSty{mycommfont}

\usepackage{bookmark}

\usepackage{amsmath}
\DeclareMathOperator{\argmin}{argmin}

\usepackage[bbgreekl]{mathbbol}
\usepackage{amsfonts}

\DeclareSymbolFontAlphabet{\mathbb}{AMSb}
\DeclareSymbolFontAlphabet{\mathbbl}{bbold}

\newdefinition{rmk}{Remark}

\usepackage[numbers]{natbib}

\def\tsc#1{\csdef{#1}{\textsc{\lowercase{#1}}\xspace}}
\tsc{WGM}
\tsc{QE}
\tsc{EP}
\tsc{PMS}
\tsc{BEC}
\tsc{DE}
\begin{document}
\let\WriteBookmarks\relax
\def\floatpagepagefraction{1}
\def\textpagefraction{.001}
\shorttitle{SDCOR: Scalable Density-based Clustering Outlierness Ratio}
\shortauthors{Sayyed~Ahmad~Naghavi~Nozad et~al.}

\title [mode = title]{SDCOR: Scalable Density-based Clustering for Local Outlier Detection in Massive-Scale Datasets}                      
%\tnotemark[1,2]
%
%\tnotetext[1]{This document is the results of the research
%   project funded by the National Science Foundation.}
%
%\tnotetext[2]{The second title footnote which is a longer text matter
%   to fill through the whole text width and overflow into
%   another line in the footnotes area of the first page.}

\author[1]{Sayyed~Ahmad~Naghavi~Nozad}
%[type=author,
%                        auid=000,bioid=1,
%                        prefix=Mr.,
%                        role=Researcher,
%                        orcid=0000-0001-7740-8387]
\cormark[1]
\ead{sa_na33@aut.ac.ir}
%\ead[url]{https://ce.aut.ac.ir/~sann_cv/}

\credit{Project administration, Conceptualization of this study, Investigation, Methodology, Software, Data curation, Formal analysis, Validation, Visualization, Resources, Writing - Original draft preparation}

\address[1]{Department of Computer Engineering, Amirkabir University of Technology, Tehran, Iran}

\author[2]{Maryam~Amir~Haeri}
%[style=chinese]
\ead{m.amirhaeri@utwente.nl}
%\ead[URL]{https://scholar.google.com/citations?user=eqX0aK8AAAAJ&hl=en&oi=ao}

\credit{Initial conceptualization, Data curation, Validation}

\address[2]{Learning, Data-Analytics, and Technology Department, University of Twente, Enschede, Netherlands}

\author[3]{Gianluigi~Folino}
%[%
%   role=Co-ordinator,
%   suffix=Jr,
\ead{gianluigi.folino@icar.cnr.it}
%\ead[URL]{http://www.folino.it/}

\credit{Supervision, Formal analysis, Validation}

\address[3]{ICAR-CNR, Rende, Italy}

%\author%
%[1,3]
%{Rishi T.}
%\cormark[2]
%\fnmark[1,3]
%\ead{rishi@stmdocs.in}
%\ead[URL]{www.stmdocs.in}

%\address[3]{STM Document Engineering Pvt Ltd., Mepukada,
%    Malayinkil, Trivandrum 695571, India}

\cortext[cor1]{Corresponding author}
%\cortext[cor2]{Principal corresponding author}
%\fntext[fn1]{This is the first author footnote. but is common to third
%  author as well.}
%\fntext[fn2]{Another author footnote, this is a very long footnote and
%  it should be a really long footnote. But this footnote is not yet
%  sufficiently long enough to make two lines of footnote text.}

%\nonumnote{This note has no numbers. In this work we demonstrate $a_b$
%  the formation Y\_1 of a new type of polariton on the interface
%  between a cuprous oxide slab and a polystyrene micro-sphere placed
%  on the slab.
%  }

\begin{abstract}
	This paper presents a batch-wise density-based clustering approach for local outlier detection in massive-scale datasets. Unlike the well-known traditional algorithms, which assume that all the data is memory-resident, our proposed method is scalable and processes the input data chunk-by-chunk within the confines of a limited memory buffer. A temporary clustering model is built at the first phase; then, it is gradually updated by analyzing consecutive memory loads of points. Subsequently, at the end of scalable clustering, the approximate structure of the original clusters is obtained. Finally, by another scan of the entire dataset and using a suitable criterion, an outlying score is assigned to each object called SDCOR (Scalable Density-based Clustering Outlierness Ratio). Evaluations on real-life and synthetic datasets demonstrate that the proposed method has a low linear time complexity and is more effective and efficient compared to best-known conventional density-based methods, which need to load all data into the memory; and also, to some fast distance-based methods, which can perform on data resident in the disk.
\end{abstract}

%This template helps you to create a properly formatted \LaTeX\ manuscript.

%\noindent\texttt{\textbackslash begin{abstract}} \dots 
%\texttt{\textbackslash end{abstract}} and
%\verb+\begin{keyword}+ \verb+...+ \verb+\end{keyword}+ 
%which
%contain the abstract and keywords respectively. 

%\noindent Each keyword shall be separated by a \verb+\sep+ command.

%\includegraphics{figs/grabs.pdf}

\begin{highlights}
	\item A novel scalable approach for local outlier detection in massive data is presented.
	\item The input data is not required to be entirely loaded in the memory as it is processed in chunks.
	\item Our assessments prove that the proposed method has a linear time complexity with a low constant.
	\item The proposed approach is superior to the state-of-the-art density- and distance-based methods.
\end{highlights}


\maketitle

\section{Introduction}\label{sec_Intro}
Outlier detection, which is a noticeable and open line of research \cite{hodge2004survey,chandola2009anomaly,zimek2012survey,wang2019progress}, is a fundamental issue in data mining. Outliers refer to rare objects that deviate from the well-defined notions of expected behavior, and discovering them is sometimes compared with searching for a needle in a haystack because the rate of their occurrence is much lower than normal objects. Outliers often interrupt the learning procedure from data for most of the analytical models, and thus, capturing them is very important because it can enhance the model accuracy and reduce the computational load of the algorithm. However, outliers are not always annoying, and sometimes, they become of particular interest for the data analyst in many problems such as controlling cellular phone activity to detect fraudulent usage, like stolen phone airtime. Outlier detection methods could also be considered as a preprocessing step, which is useful before applying any other advanced data mining analytics, and it has a wide range of applicability in many research areas, including intrusion detection, activity monitoring, satellite image analysis, medical condition monitoring, etc. \cite{hodge2004survey,agyemang2006comprehensive}.

Outliers can be generally partitioned into two different categories, namely global and local. Global outliers are objects that show significant abnormal behavior compared to the rest of the data, and thus in some cases, they are considered point anomalies. On the contrary, local outliers only deviate significantly w.r.t. a specific neighborhood of the object \cite{chandola2009anomaly,han2011data}. In \cite{breunig2000lof,de2010finding}, it is noted that the concept of the local outlier is more comprehensive than that of the global outlier; i.e., a global outlier could also be considered as a local one, but not necessarily vice versa. This is the reason that makes finding the local outliers much more cumbersome.

\subsection{Motivation}\label{subSec_Motiv}
In recent years, advances in data acquisition have made massive collections of data, which contain valuable information in diverse fields like business, medicine, society, government, etc. As a result, the common conventional software methods, including but not limited to \cite{zhang2009new,breunig2000lof,kriegel2009loop,jin2006ranking,tang2017local,huang2016non,wahid2019rkdos,wahid2020odra,wahid2020nanod,xie2020local,wu2011information,dang2013local,he2003discovering,duan2009cluster,jobe2015cluster,huang2017novel,moonesignhe2006outlier,moonesinghe2008outrank,wang2018new,wang2018outlier,wang2019vos,amil2019outlier}, for processing and management of such massive amount of data will no longer be efficient, because most of these methods assume that the data is memory-resident and their computational complexity for large-scale datasets is really expensive \cite{wang2019progress}.

One of the usual ways to solve this issue is the use of parallel/distributed computing techniques \cite{zeng2012distributed,januzaj2004scalable,angiulli2012distributed,mao2018outlier,chen2018practical,yan2017distributed1,yan2017distributed2}. In such a strategy in which both the hardware and software facilities are extensively employed, a big or even an intolerable task for a single computing module is divided into a pack of smaller assignments, as each of them is appointed to a distinct processing unit. Therefore, more work can be carried out simultaneously by exerting more devices than the condition in which the task was about to be done at once through an isolated computer.

However, distributed solutions come along with some inevitable downsides like being more complicated to implement and troubleshoot, requiring additional resources, colossal power consumption, and the principal need for hardware cooling technologies in some cases; furthermore, such dispersed resolutions commonly accomplish an activity with a less efficacy compared to their serial counterparts on account of communication and coordination overheads.

The motivation behind this study is to propose a scalable in-memory algorithm that utilizes merely a single stand-alone computer while delivering competing results with the single-run state of the data mining algorithm on the entire dataset.

\subsection{Contribution}\label{subSec_Cntrb}
In this paper, we propose a new scalable and density-based clustering method for local outlier detection in massive-scale datasets that cannot be loaded into memory at once, employing a chunk-by-chunk load procedure. In practice, the proposed approach is a clustering method for huge datasets in which outlier identification comes after that as a side effect, and it is inspired by a scaling clustering algorithm for very large databases named after its authors, BFR\cite{bradley1998scaling}. However, BFR has some weak spots: it needs to know the actual number of original clusters in data; it has a strong assumption on the structure of existing clusters, which ought to be Gaussian distributed with uncorrelated attributes; and more importantly, it is not introducing noise. In short, this clustering algorithm works as follows. First, it reads the data as successive (preferably random) samples so that each sample can be stored in the memory buffer, and then it updates the current clustering model over the contents of the buffer. Based on the updated model, singleton data are classified into three groups: some of them can be discarded over the updates to the sufficient statistics (Discard Set, DS); some can be moderated through compression and abstracted as sufficient statistics (Compression Set, CS); some demand to be retained in the buffer (Retained Set, RS).

Like BFR, our proposed method operates within the confines of a limited memory buffer. Thus, assuming that an interface to the database allows the algorithm to load an arbitrary number of requested data points, whether sequentially or randomized, we are forced to load the data chunk-by-chunk so that there is enough space for both loading and processing each chunk at the same time. The proposed approach is based on clustering, and therefore, it must avoid outliers being able to play any influential role in forming and updating clusters. After processing each chunk, the algorithm should combine its approximate results with those of the previous chunks in a way that the final approximate result will compete with the same result obtained by processing the entire dataset at once. An algorithm, which is capable of handling data in such a gradual way and finally provides an approximate result, from an operational perspective, is called a scalable algorithm \cite{yin2013scalable}. Moreover, from an algorithmic point of view, scalability means that algorithm complexity should be nearly linear or sublinear w.r.t. the problem size \cite{teng2016scalable}. 

In more detail, the proposed method comprises three steps. In the first step, a primary random sampling is carried out to create an abstraction of the whole data on which the algorithm works. Then, an initial clustering model is built, and some information required for the next phase of progressive clustering will be acquired. In the second step, on the basis of the currently loaded chunk of data into the memory, a scalable density-based clustering approach is executed in order to identify dense regions, which leads to building incrementally some clusters, named miniclusters or subclusters. When all chunks are processed, the final clustering model will be built by merging the information obtained through these miniclusters. Finally, by applying a Mahalanobis distance criterion \cite{mahalanobis1936generalized,ro2015outlier} to the entire dataset, an outlying score is assigned to each object.

In summary, the main contributions of our proposed method are listed as follows:

\begin{itemize}\label{item_Cntrb}
	\item \relax There is no need to know the real number of original clusters.
	
	\item \relax It works better with Gaussian clusters having correlated or uncorrelated features, but it also works well with convex-shaped clusters following an arbitrary distribution.
	
	\item \relax It has a linear time complexity with a low constant.
	
	\item \relax In spite of working in a scalable manner and operating on chunks of data, in terms of detection accuracy, it is still able to compete with conventional density-based methods, which maintain all the data in the memory; and also, with some fast distance-based methods, which do not require to load the entire data into the memory at their training stage.
\end{itemize}

The paper is organized as follows: Section~\ref{sec_RelWr} discusses some related work in the field of outlier detection. In Section~\ref{sec_PropAppr}, we present the detailed descriptions of the proposed approach. Section~\ref{sec_ExprEval} delineates the experimental design of our analysis for the proposed method and the other contending techniques. In Section~\ref{sec_ExprResl}, the experimental results on various real and synthetic datasets are provided. Finally, conclusions are given in Section~\ref{sec_Conc}.

\section{Related work}\label{sec_RelWr}
Outlier detection methods can generally be divided into the following eight categories \cite{aggarwal2015data,aggarwal2015outlier,domingues2018comparative,wangnew}: extreme value analysis, probabilistic methods, distance-based methods, density-based methods, clustering-based methods, graph-based methods, information-theoretic methods, and isolation-based methods. Besides, after considering these categories, in a discussion, we will look upon the capabilities of the corresponding methods to operate on massive-scale data.

\subsection{Extreme value analysis}\label{subSec_ExtValAnls}
In extreme value analysis, the overall population is supposed to have a unique probability density distribution, and only those objects at the very ends of it are considered outliers. In particular, these types of methods are useful to find global outliers \cite{aggarwal2015data,cabras2007extreme}.

\subsection{Probabilistic methods}\label{subSec_ProbMeth}
In probabilistic methods, we assume that the data were generated from a mixture of different distributions as a generative model, and we use the same data to estimate the parameters of the model. After determining the specified parameters, outliers will be those objects with a low likelihood of being generated by this model \cite{aggarwal2015data}. Sch{\"{o}}lkopf et al. \cite{scholkopf2001estimating} propose a supervised approach, in which a probabilistic model w.r.t. the input data is provided so that it can fit normal data in the best possible way. In this manner, the goal is to discover the smallest region that contains most of the normal objects; data outside this region are supposed to be outliers. This method is, in fact, an extended version of Support Vector Machines (SVM), which has been improved to cope with imbalanced data; the other name for this method in the literature is one-class SVM, or in brief, OCSVM \cite{tax1999support}. In practice, in this way, a small number of outliers is considered as belonging to the rare class and the rest of the data as belonging to normal objects.

\subsection{Distance-based methods}\label{subSec_DistMeth}
In distance-based methods, distances among all objects are computed to detect outliers. An object $ O $ is assumed to be a distance-based outlier if at least fraction $ p_0 $ of the objects in the dataset has a distance greater than $ d_0 $ from $ O $ \cite{knox1998algorithms,ramaswamy2000efficient}. Another definition for the distance-based outliers denotes that concerning the two input parameters, $ k $ as a positive integer and $ R $ as a positive real number, an object $ O $ is reported as an outlier if less than $ k $ objects in the dataset rest in the distance $ R $ from $ O $ \cite{angiulli2009dolphin}. In \cite{zhang2009new}, a Local Distance-based Outlier Factor (LDOF) is proposed to find outliers in scattered datasets, which employs the relative distance from each data object to its nearest neighbors.

Bay and Schwabacher \cite{bay2003mining} propose an optimized nested-loop algorithm based on the \textit{k} Nearest Neighbors (\textit{k}NN) distances among objects, that has a near-linear time complexity and is shortly named ORCA (Optimal Reciprocal Collision Avoidance). ORCA shuffles the input dataset in random order using a disk-based algorithm and processes it in blocks, as there is no need to load the entire data into the memory. It keeps looking for a set of the user-defined number of data points as potential anomalies, as well as for their anomaly scores. The cut-off value is set as the minimum outlierness score of the set, and it will get updates if there is a data point having a higher score in other blocks of data. For data points that obtain a lower score than the cut-off, they should be pruned; this pruning scheme will only expedite the process of distance computation in the case of data being ordered in an uncorrelated manner. ORCA's worst-case time-complexity is $ O\left( n^2\right) $, and the I/O cost for the data accesses is quadratic. For the anomaly definition, it can use either the distance to the \textit{k}-th nearest neighbor or the average distance of \textit{k}NN.

Angiulli and Fassetti \cite{angiulli2009dolphin} propose DOLPHIN (Detecting OutLiers PusHing objects into an INdex), a distance-based outlier mining method that is particularly dedicated to operating on disk-resident data and functions efficiently in terms of CPU and I/O cost two at a time. Furthermore, both theoretical and practical proofs are presented that the proposed method occupies the memory space as much as a small portion of the dataset. DOLPHIN obtains its efficiency certainly through combining three policies in an integrated plan, namely: 1) careful selection of objects to be retained in the buffer; 2) employing decent pruning strategies; 3) applying effective similarity inspection approaches, which is especially achieved without requiring prior indexing the entire input data, differently from the other competing techniques. Moreover, DOLPHIN is capable of being applied on any kind of records attributed to either metric or non-metric scopes.

\textit{S\textsubscript{p}} \cite{sugiyama2013rapid} is a simple and rapid distance-based method that utilizes the nearest neighbor distance on a small sample from the dataset. It takes a small random sample of the entire dataset and then assigns an outlierness score to each point, as the distance from the point to its nearest neighbor in the sample set. Therefore, this method enjoys a linear time complexity concerning each one of the essential variables, viz the number of objects, the number of dimensions, and the number of samples; furthermore, \textit{S\textsubscript{p}} has a constant space complexity, which makes it ideal for analyzing massive datasets.

\subsection{Density-based methods}\label{subSec_DensMeth}
In density-based methods, the local density of each object is calculated in a specific way and then is utilized to define the outlier scores. Given an object, the lower its local density compared to its neighbors, the more likely it is that the object is an outlier. Density around the points could be calculated by using many techniques, which most of them are distance-based \cite{breunig2000lof,aggarwal2015data}. For example, Breunig et al. \cite{breunig2000lof} propose a Local Outlier Factor (LOF) that uses the distance values of each object to its nearest neighbors to compute local densities. However, LOF has a drawback: the scores obtained through this approach are not globally comparable between all objects in the same dataset or even in different datasets. The authors of \cite{kriegel2009loop} introduce the Local Outlier Probability (LoOP), which is an enhanced version of LOF. LoOP gives each object a score in the interval [0,1], which is the probability of the object being an outlier and is widely interpretable among various situations. Moreover, two distributed versions of LOF are presented in \cite{yan2017distributed1} and \cite{yan2017distributed2} to operate on very large datasets. In \cite{yan2017distributed1}, a distributed LOF pipeline framework (DLOF) along with a data assignment strategy are introduced to efficiently handle the required information exchange among various machines and elevating the independence of every processing unit; this will reduce the number of data replications and save more resources. In \cite{yan2017distributed2}, a novel and cautious removal technique is proposed to effectively eliminate those objects that are not potential to be reported as top-ranked outliers and also not capable of being contemplated as possible neighbors of other points residing on other machines.

An INFLuenced Outlierness (INFLO) score is presented in \cite{jin2006ranking}, which adopts both \textit{k}NN and Reverse Nearest Neighbors (RNN) of an object to estimate its relative density distribution. Furthermore, Tang and He \cite{tang2017local} propose a local density-based outlier detection approach, concisely named RDOS, in which the local density of each object is approximated with the local Kernel Density Estimation (KDE) through nearest neighbors of it. In this approach, not only the \textit{k}NN of an object are taken into account, but also the RNN and the Shared Nearest Neighbors (SNN) are considered for density distribution estimation as well. Moreover, a parameter-free density-based algorithm for both clustering and outlier detection purposes is presented in \cite{rahman2018unique}, which utilizes two novel neighborhood concepts, namely Unique Closest Neighbor (UCN) and Unique Neighborhood set (UNS), depending on the underlying data distribution. This proposed approach does not follow random or repetitive solutions to culminate in the best candidate outcome, and hence, is computationally cost-effective.

A non-parameter outlier detection algorithm based on Natural Neighbors (NaN) is proposed in \cite{huang2016non}. In this parameter-free method, first, a novel search algorithm using KD-tree data structure \cite{bentley1975multidimensional} is introduced for detecting NaN instead of \textit{k}NN for every object. The great benefit of NaN over \textit{k}NN is that the neighbor search method is free of any input parameter, and the calculation is upon the intrinsic manifold of data. In addition, the NaN concept is scale-free, which means that the number of neighbors for various objects is not essentially the same. Furthermore, the new concepts of Natural Influence Space (NIS) and Natural Neighbor Graph (NNG) are proposed to finally compute the Natural Outlier Factor (NOF) for each individual data point. A higher NOF value for an object indicates the higher possibility for it to be identified as an outlier.

Wahid and Rao have recently proposed three different density-based methods, shortly named RKDOS \cite{wahid2019rkdos}, ODRA \cite{wahid2020odra} and NaNOD \cite{wahid2020nanod}. RKDOS is mostly similar to RDOS, in which by using both \textit{k}NN and RNN of each point, the density around it is estimated employing a Weighted Kernel Density Estimation (WKDE) strategy with a flexible kernel width. Moreover, the Gaussian kernel function is adopted to corroborate the smoothness of estimated local densities, and also, the adaptive kernel bandwidth enhances the discriminating ability of the algorithm for better separation of outliers from inliers. ODRA is specialized for finding local outliers in high dimensions and is established on a relevant attribute assortment approach. There are two major phases in the algorithm, which in the preliminary one, unimportant features and data objects are pruned to achieve better performance. In other words, ODRA firstly tries to distinguish those attributes which contribute substantially to anomaly detection from the other irrelevant ones; also, a significant number of normal points residing in the dense regions of data space will be discarded for the sake of alleviating the detection process. In the second phase, like RDOS, a KDE strategy based on \textit{k}NN, RNN, and SNN of each data object is utilized to evaluate the density at the corresponding locations, and finally, assign an outlying degree to each point. In NaNOD, the concept of Natural Neighbors (NaN) is used to adaptively attain a proper value for \textit{k} (number of neighbors), which is called here the Natural Value (NV), while the algorithm does not ask for any input parameter to acquire that. Thus, the parameter choosing challenge could be mitigated through this procedure. Moreover, a WKDE scheme is applied to the problem to assess the local density of each point, as it takes the benefit of both \textit{k}NN and RNN categories of nearest neighbors, which makes the anomaly detection model adjustable to various kinds of data patterns.

Besides, in recent times, Xie et al. \cite{xie2020local} have proposed LGOD, an outlier model founded upon Newton's theory of gravitation \cite{newton1802mathematical}. In this model, every individual point is supposed as an element with mass, associated with a Local Resultant Force (LRF) engendered by its neighboring points. LGOD first determines the LRFs on each data object; then, regarding its tolerance to proximity parameter \textit{k}, by aggregating the varying LRF values out of the fluctuating proximity parameter, outlying scores are deduced.

\subsection{Clustering-based methods}\label{subSec_CustMeth}
Clustering-based methods use a global analysis to detect crowded regions, and outliers will be those objects not belonging to any cluster \cite{aggarwal2015data}. A Cluster-Based Local Outlier Factor (CBLOF) in \cite{he2003discovering}, and a Cluster-Based Outlier Factor (CBOF) in \cite{duan2009cluster} are presented. In both of them, after the clustering procedure is carried out, in regard to a specific criterion, clusters are divided into two large and small groups; it is assumed that outliers lie in the small clusters. At last, the distance of each object to its nearest large cluster is used in different ways to define the respective outlier score. Furthermore, a distributed strategy for both clustering and outlier detection tasks is proposed in \cite{chen2018practical}, in which the classic \textit{k}-means/median clustering techniques are fundamentally studied to become operable on datasets disseminated across various sites. The proposed method is efficient considering both time and required communications among multiple machines and has a satisfactory estimation level, especially on global outliers.

Jobe and Pokojovy \cite{jobe2015cluster} propose a data-intensive computing approach that is established upon cluster analysis and employs a reweighted variant of Rousseeuw’s minimum covariance determinant method \cite{rousseeuw1999fast}. Adopting the powerful squared Mahalanobis distance statistic is the core idea of this study, although the detection capability of this criterion is substantially declined with the increasing number of outliers in data. Thus by applying a clustering-based multi-stage algorithm in this method, the potential outlying points which could be misclassified through the Mahalanobis distance measure are primarily distinguished from true inliers. Furthermore, Huang et al. \cite{huang2017novel} propose a novel outlier cluster detection method called ROCF, which does not require the count of top-N outliers as an input parameter. Using the mutual neighbors concept, ROCF firstly builds a neighborhood graph entitled MUNG (MUtual Neighbors Graph). Then, regarding the truth that outlying clusters are usually smaller in size than the normal clusters, and also by automatically unraveling the approximate outlier ratio of the input dataset, both singular outliers and anomalous minuscule clusters are discovered.

\subsection{Graph-based methods}\label{subSec_GrphMeth}
Graph-based methods are among the most robust approaches in the outlier mining area. They have called immense consideration in recent time because they can substantially express various data conditions \cite{cook2000graph,akoglu2015graph} and extraordinarily grasp the wide-ranging interrelationships within the data objects \cite{ranshous2015anomaly,yu2016survey}. In general, each individual object is characterized as a graph node, and the connections and dependencies among data elements are represented as the linking edges between the nodes. Then by assessing the graph structure \textemdash~whether locally or globally \textemdash~or even through evaluating the changing process of the graph or other suitable criteria, the anomalousness degree for every data object will be defined.

Moonesignhe and TAN \cite{moonesignhe2006outlier,moonesinghe2008outrank} propose OutRank, which is among the very first graph-based approaches for outlier detection in datasets comprising of multidimensional points. OutRank is a stochastic algorithm in which, at first, a graph representation of the input data is constructed based on the similarity among objects and the count of shared neighbors between them. Then, employing the Markov chain model established regarding the attained graph, every object in data will attain an outlying score. One of the major problems with the graph-based methods is that they overlook the local information in each node vicinity, and unfortunately, this entails so many inliers be mistakenly identified as outliers, which is referred to as a high false-positive rate. In \cite{wang2018new}, a local information graph is built upon combining the graph delineation of the input data with the local information around each data object to overcome this issue. By utilizing the local information graph, the mismatching inter-dependencies among different kinds of objects are captured. Finally, the outlying scores are defined by applying a random walk on the graph. Wang et al. \cite{wang2018outlier} propose another graph-based method that utilizes multiple neighborhood graphs to obtain diverse local information from different vantage points. Then, some stationary distribution vectors are acquired by executing a random walk process on these neighborhood graphs that are developed through distinctive neighbor sizes. Finally, outlier scores are derived over a novel designed scoring function established upon various change patterns of the corresponding values in the stationary distribution vectors.

Moreover, in \cite{wang2019vos}, Wang et al. again, by integrating the local information with the latent associations in the graph rendering of the genuine dataset, propose a modern graph-based method called Virtual Outlier Score (VOS). In VOS, at the start, a similarity graph is initiated through employing the top-\textit{k} analogous neighbors of each individual object; furthermore, a \textit{k}-virtual graph is determined using the new concepts of virtual nodes and groups of virtual edges, which captures both local and global information among objects within the dataset. Then, by applying a well-suited version of the Markov random walk process on the highly associated virtual graph, under the fact that for accessing the likely anomalies in data, they should obtain more weight by the random walker than the other inliers, the algorithm reaches equilibrium or a stationary distribution in its operation. Finally, the virtual outlier ranks will be inferred from this stationary situation. Amil et al. \cite{amil2019outlier} propose two methods that begin with a graph structure and are more useful for anomaly detection in high-dimensional data. Both methods depend on a reliable definition for the distance between pairs of data elements; thus, they can deal with non-specific data types. The nodes and weighted edges in the representing graph correspond to data objects and distances among them, respectively. In the first method, an anomaly degree is specified for each item concerning the graph fragmentation. The other method employs the famous non-linear dimensionality reduction approach named IsoMap \cite{tenenbaum2000global} and defines the anomaly scores regarding the variations between the geodesic distances and the distances in the acquired subspace.

\subsection{Information-theoretic methods}\label{subSec_InfrTher}
Information-theoretic methods could be particularly deemed at almost the same level as distance-based and other deviation-based models. The only exception is that in such methods, a fixed concept of deviation is determined at first, and then, anomaly scores are established through evaluating the model size for this specific type of deviation; unlike the other usual approaches, in which a fixed model is determined initially and employing that, the anomaly scores are obtained by means of measuring the size of the deviation from such model \cite{aggarwal2015data}.

Wu and Wang \cite{wu2011information} propose a single-parameter method for outlier detection in categorical data using a new concept of Holoentropy, and by utilizing that, a formal definition of outliers and an optimization model for outlier detection is presented. According to this model, a function for the outlier factor is defined, which is solely based on the object itself, not the entire data, and it could be updated efficiently. There are two proposed greedy Information-Theory-Based (ITB) algorithms for this method, namely ITB-SP (Single Pass) and ITB-SS (Step-by-Step), that essentially can operate only on nominal space. However, the authors claim that they can be adapted to numerical space through either the extension of holoentropy or employing a suitable and viable discretization approach. Moreover, Dang et al. \cite{dang2013local} propose an information-theoretic method named LODI in which besides identifying local outliers in continuous data, the anomalousness reason for any of them is presented too. In LODI, the quadratic entropy is investigated in an appropriate manner to select a neighboring set per each point. Then, following a learning method based on matrix eigen-decomposition, an optimal subspace is defined in which an outlier candidate is as much as possible isolated from its neighborhood. The revealed features in the corresponding subspace are critical to interpreting the outstanding characteristics of outliers.

\subsection{Isolation-based methods}\label{subSec_IsolMeth}
Isolation-based methods were firstly introduced in \cite{liu2008isolation,liu2012isolation} and the authors named their novel method \textit{i}Forest. The main idea behind \textit{i}Forest was elicited from a well-known ensemble method called Random Forests \cite{breiman2001random}, which is mostly employed in classification problems. In this approach, firstly, it is essential to build an ensemble of isolation trees (\textit{i}Trees) for the input dataset, then outliers are those objects with a short average path length on the corresponding trees. For making every \textit{i}Tree, after acquiring a random sub-sample of the entire data, it is recursively partitioned by randomly choosing an attribute and then randomly specifying a split value among the interval of minimum and maximum values of the chosen attribute. Since this type of partitioning can be expressed by utilizing a tree structure, thus the number of times required to partition the data for isolating an object is equal to the path length from the starting node to the ending node in the respective \textit{i}Tree. In such a situation, the tree branches which bear outliers have a remarkably lower depth because these outlying points fairly deviate from the normal conduct in data. Therefore, data instances that possess substantially shorter paths from the root in various \textit{i}Trees are more potential to be outliers.

One chief issue about using such an isolation-based method is that with the increase in the number of dimensions, if an incorrect attribute choice for the splitting matter is made at the higher levels of the \textit{i}Tree, then the probability of detection outcomes getting misled, grows potentially. Nevertheless, the advantage of \textit{i}Forest is that by considering the idea of isolation, it can utilize the sub-sampling in a more sophisticated way than the existing methods and provide an algorithm with a low linear time complexity and a low memory requirement \cite{aggarwal2015outlier}. Moreover, in \cite{bandaragoda2014efficient,bandaragoda2018isolation}, an enhanced version of \textit{i}Forest, named \textit{i}NNE is proposed, which tries to overcome some drawbacks of the primary method, including the incapability of efficiently detecting local anomalies, anomalies with a low amount of pertinent attributes, global anomalies that are disguised through axis-parallel projections, and finally anomalies in a dataset containing various modals.

\subsection*{Discussion}\label{subSec_ReltWrksDisc}\addcontentsline{toc}{subsection}{Discussion}
Almost all of the mentioned studies here for outlier identification assume the input data as static, centralized, and memory-resident. Therefore, they incur an expensive computational cost for large-scale datasets; i.e., they cannot simply scale well to massive data.

Extreme value analysis is commonly applied to some specific cases, where outliers are acknowledged to be existing at the margins, not the sparse internal areas of the dataset; hence, it is not employed for local outlier identification in typical KDD applications \cite{aggarwal2015outlier}. OCSVM is a semi-supervised outlier model and is mostly utilized in ``novelty detection'' applications, where the input data contains only good (normal) samples, and the primary purpose is to investigate whether new incoming observations \textemdash~maybe through a stream \textemdash~match the existing model. Furthermore, there are some cases for OCSVM in which the training data is contaminated with some abnormal instances (anomalies) to evaluate the algorithm robustness with regard to a regularization parameter meaningfully greater than the anticipated outliers portion \cite{domingues2018comparative}. However, this does not change the original method essence and its crucial difference with unsupervised models \textemdash~which are of our main interest here \textemdash~where there is no need for any information about object labels, and the input data may contain outliers that require to be identified. This study focuses on ``outlier detection'' in static numerical applications where the outlier labels are only employed while assessing the final detection outcomes.

Information-theoretic techniques are generally established upon analyzing the individual data attributes and the corresponding correlations between pairs of them to build the detection model; this leads to expensive computations as in the best case, the associated computational complexity is $ O\left( np\right) $, a multiplication of the cardinality by the dimensionality of the data, or even higher \cite{cover1999elements}. Hence, such approaches mostly fail at processing huge datasets unless they are exerted through a sophisticated manner to cope with massive scales.

Distance-, density-, and clustering-based methods all fall under the proximity-based outlier models. In such methods, the key idea is to recognize outliers as specific points far away from the rest of the data. This causes the critical complexity stress be on the nearest neighbors calculations which may require $ O\left(n^2 \right) $ time, unless an indexing strategy is exerted to expedite the computations and reduce the complexity to, e.g., $ O\left(n\log\left( n\right)\right) $ \cite{aggarwal2015outlier}. Nevertheless, even indexing approaches degenerate in high dimensions, and the situation can become aggravated if the underpinning data patterns are not in support of effective pruning rules \cite{bay2003mining,angiulli2009dolphin}. Moreover, for the graph-based methods, as they rely on the computation of nearest neighbors for each data object, too, the same scenario is repeated.

As a density-based technique, ODRA has specifically a complexity equal to the product of cardinality and dimensionality, which is not suitable for large scales. Cluster-analysis-based work in \cite{jobe2015cluster} is introduced as a computer-intensive algorithm, hence not apt for processing huge data sizes. Furthermore, the time complexity of \cite{he2003discovering} is outlined by the authors as linear with solely the dataset cardinality; however, the first step of the proposed clustering-based algorithm \textemdash~named \textit{FindCBLOF} \textemdash~is merely applying the \textit{Squeezer} clustering approach \cite{he2002squeezer} to the data, with a computational complexity being linear with the number of points, the number of dimensions, and the final count of clusters. Consequently, \textit{FindCBLOF} suffers from much more complexity and thus cannot scale favorably with the problem size. Moreover, the distributed works in \cite{yan2017distributed1,yan2017distributed2,chen2018practical} require multiple resources for execution; thus, they are out of our concern in this study as we are concentrating on operating on a single processing unit.

Isolation-based methods enjoy a linear time complexity with a low constant while requiring a minimal amount of memory. However, they are out of the context of the proximity-based models as being solely based upon isolating instances without employing any distance or density gauge to calculate the anomaly scores; hence, they do not fit in our experimental analysis. ORCA, DOLPHIN, and \textit{S\textsubscript{p}} are among the enhanced distance-based techniques for large-scale and disk-resident data that, while requiring a relatively small portion of memory space, claim competing linear arithmetical complicacies. Here, we present a scalable proximity-based algorithm to function on massive data that enjoys a low linear time complexity and operates reasonably well over a restricted memory space.

\subsection*{Remarks}\label{sec_Rmrks}\addcontentsline{toc}{subsection}{Remarks}

%\textbf{Remark 1. }
\begin{rmk}
	According to the fact that during scalable clustering, we use the Mahalanobis distance measure to assign each object to a minicluster, and besides, the size of temporary clusters is much smaller than that of original clusters, it would be worth mentioning an important matter here. Concerning \cite{ro2015outlier,filzmoser2008outlier,hubert2005robpca,ayyildiz2012short}, in the case of high-dimensional data, classical approaches based on the Mahalanobis distance are usually not applicable. Because when the cardinality of a cluster is less than or equal to its dimensionality, the sample covariance matrix will become singular and not invertible \cite{ledoit2004honey}; hence, the corresponding Mahalanobis distance will no longer be reliable.
	
	Therefore, to overcome such a problem, we need to resort to dimensionality reduction approaches in a preprocessing step. However, due to the serious dependence of some dimensionality reduction methods like PCA \cite{pearson1901liii} to the original attributes and the consequent high computational load because of the huge volume of the input data, we need to look for alternative methods to determine a basis for data projection.
	
	A simple and computationally inexpensive alternative is the exercise of random basis projections \cite{johnson1984extensions,dasgupta1999elementary,achlioptas2001database}. The main characteristic of these types of methods is that they will approximately preserve pairwise Euclidean distances between data points; in addition, the dimension of the transformed space is independent of the original dimension and only relies logarithmically on the number of data points. Finally, after such a preprocessing step, we can be optimistic that the singularity problem will not be present during the clustering procedures; in the event of happening, we would have a suitable mechanism to handle it.
\end{rmk}

%\textbf{Remark 2. }
\begin{rmk}
	As stated earlier, our proposed approach is inspired by BFR. However, BFR, by default, uses the K-means algorithm \cite{forgey1965cluster} in almost all of its clustering procedures. In addition to this drawback of the K-means algorithm, which is being enormously dependent on foreknowing the actual number of original clusters, in the case of outliers presence, K-means performs poorly; therefore, we need to resort to a clustering approach that is resistant to anomalies in data. Here in this paper, we prefer to employ the density-based clustering method, DBSCAN \cite{ester1996density}\footnote{However, we will demonstrate that sometimes throughout scalable clustering, even DBSCAN may produce some miniclusters that cannot abide outliers because of some data assignment restraints; hence, we will be forced to use the same K-means algorithm to fix the issue.}. Other noise-tolerant clustering algorithms \cite{rodriguez2014clustering,rahman2020clustering,rahman2018unique,bryant2017rnn,lotfi2020density,liu2018shared,xie2016robust,mehmood2016clustering,liu2019constraint,liu2019clustering,bie2016adaptive,chen2011apscan,zhou2018robust,pavan2006dominant,hou2016dsets,hou2017parameter} could be utilized as a substitute option, too, but definitely with their own specific configurations.
	
	However, DBSCAN is strongly reliant upon the choice of its parameters, namely the minimal number of neighbors including itself, \textit{MinPts}, within the range \textit{Eps} \textemdash~with a random distance measure, which herein is chosen as the Euclidean distance. Thus, we are forced to utilize, e.g., some heuristic or optimization algorithm to find the optimal values for these parameters. Here, we prefer to use an evidence-based approach proposed firstly in the DBSCAN original paper, or a combination of this approach with the Particle Swarm Optimization (PSO) algorithm \cite{kennedy1995particle} to locate the optimal parameters. More details can be found in Appendix \ref{appSec_DBSCANoptmlParamSampl}.
\end{rmk}

%\textbf{Remark 3. }
\begin{rmk}
	Another important difference between the proposed method and BFR concerns the volume of structural information, which they need to store for clustering procedures. As the proposed method, differently from BFR, can handle Gaussian clusters with correlated attributes too, thus the covariance matrix will not always be diagonal and could have many non-zero elements. Therefore, the proposed method will consume more space than BFR for building the clustering structures.
	
	Since the Mahalanobis distance criterion is crucially based on the covariance matrix, this matrix will be literally the most prominent property of each subcluster. However, according to the high computational expense of computing the Mahalanobis distance in high-dimension spaces, as in \cite{filzmoser2008outlier,maronna2002robust}, we will use the properties of Principal Components (PCs) in the transformed space. Therefore, the covariance matrix of each minicluster will become diagonal, and by transforming each object to the new space of the minicluster, like BFR, we can calculate the Mahalanobis distance without the need to use matrix inversion.
	
	According to \cite{leskovec2014mining}, when the covariance matrix is diagonal, the corresponding Mahalanobis distance becomes the same normalized Euclidean distance. Moreover, we can establish a threshold value for defining the Mahalanobis radius. If the value of this threshold is, e.g., 4, it means that all points on this radius are as far as four standard deviations from the mean; if we just denote the number of dimensions by $ p $, the size of this Mahalanobis radius is equal to $ 4 \sqrt p $.
\end{rmk}

\section{Proposed approach}\label{sec_PropAppr}
The proposed method consists of three major phases. In the first phase, a preliminary random sampling is conducted in order to obtain the main premises on which the algorithm works, i.e., some information on the original clusters and some parameters useful for the progressive clustering. In the second phase, a scalable density-based clustering approach is carried out in order to recognize dense areas on the basis of the currently loaded chunk of data points in the memory. Clusters built incrementally in this way are called miniclusters or subclusters, and they form the temporary clustering model. In more detail, after loading each chunk of data, according to the points already loaded in the memory and those undecided from the previous chunks, and by employing the Mahalanobis distance measure and in respect to the density-based clustering criteria, we update the temporary clustering model, which consists of making some changes to the existing miniclusters or adding new subclusters.

Note that, in the whole scalable clustering procedure, our endeavor aims not to let outliers participate actively in forming and updating any minicluster, and thus, after processing the entire chunks, there will be some objects in the buffer remained undecided. Some of these data are true outliers, while others are inliers, which, due to constraints, have failed to play an influential role in forming a subcluster. Finally, all these undecided points are cleared from the buffer, while only the structural information of the temporary clusters is maintained in the memory. Then, at the last part of scalable clustering, depending on the miniclusters associated with each initial minicluster out of the ``Sampling'' stage, we combine them to obtain the final clusters, which their structure will be approximately the same as of the original clusters.

\begin{figure}[pos=!ht]
	\centering
	\includegraphics[width=1\linewidth]{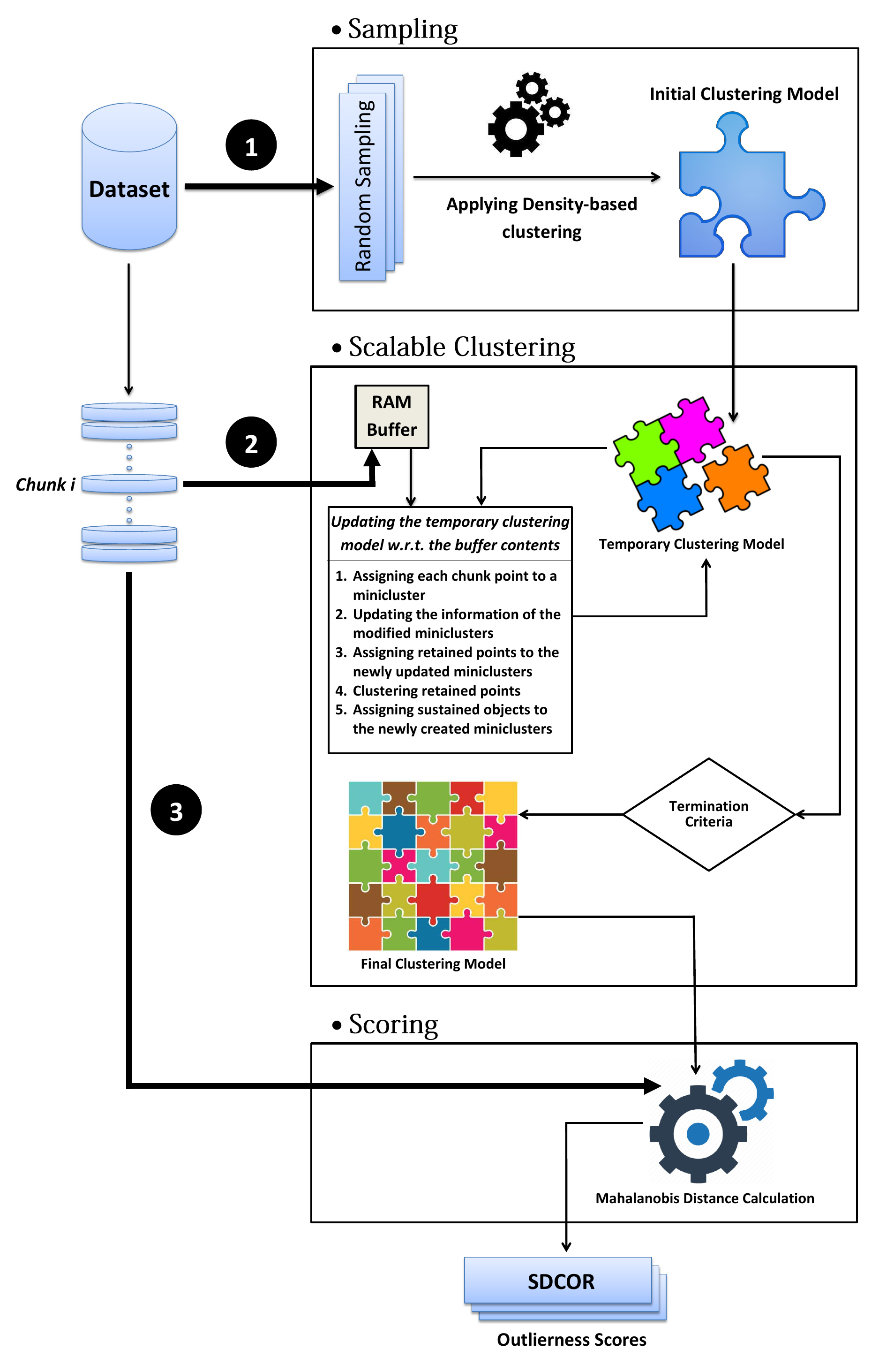}
	\caption{Software architecture of the proposed approach}
	\label{figure-SoftArch}
\end{figure}

At last, in the third phase of the proposed approach, w.r.t. the final clustering model gained out of the second phase, once again, we process the entire dataset in chunks to give each object an outlying score, according to the same Mahalanobis distance criterion. Fig.~\ref{figure-SoftArch} illustrates the software architecture of the approach. Moreover, in Table~\ref{table-majNot}, the main notations used in the paper are summarized.

\begin{table}[pos=!ht]
	\caption{Major notations}
	\label{table-majNot}
	%	\tiny
	\scriptsize
	\centering
	\setlength{\tabcolsep}{.9pt}
	
	\begin{tabular}{p{2cm}p{6.1cm}}
		\toprule
		\bfseries Notation & \bfseries Description\\
		\midrule
		$ \left[ \mathcal{X}\right]_{n\times p} $ & Input dataset $ \mathcal{X} $ with $ n $ objects and $ p $ dimensions\\
		$ x\in \mathcal{X} $ & An instance in $ \mathcal{X} $\\
		$ \mathcal{S}\subset \mathcal{X} $ & A random sample of $ \mathcal{X} $\\
		$ \mathbb{X}\subset \mathcal{X} $ & A set of some points in the buffer\\
		$ \mathcal{Y}\subset \mathcal{X} $ & A chunk of data\\
		$ \Omega $ & A partition of some data points in the memory into some miniclusters as $ \left\lbrace \mathbb{X}_{1},\cdots,\mathbb{X}_{\mathbbl{k}}\right\rbrace $\\
		$ n_\mathfrak{C} $ & Number of objects in $ \mathcal{Y} $\\
		$ \left\lbrace \Gamma\right\rbrace_{8\times \mathcal{L}} $ & Information array of temporary clusters with 8 properties for each minicluster\\
		$ \left\lbrace \Gamma_{i}\right\rbrace_{8\times 1}\in \Gamma $ & Information standing for the $ i $-th minicluster in $ \Gamma , 1\le i\le \mathcal{L}$\\
		$ X_{i} $ & Minicluster points associated with $ \Gamma_{i} $, which are removed from the buffer\\
		$ \Re $ & Retained set of objects in the RAM buffer\\
		$ \Re_{i} $ & The $ i $-th minicluster of retained points in the buffer, discovered through DBSCAN\\
		$ \Re_{\zeta} $ & Retained set of objects in the buffer introduced as noise by DBSCAN\\
		$ \gamma $ & List of indices to recently created or updated miniclusters associated with $ \Gamma $, to be checked on for membership\\
		$ \gamma' $ & Temporary list of indices to recently created or updated miniclusters associated with $ \Gamma $, to be checked on for membership\\
		$ m $ & Current number of objects associated with $ \Gamma_{i} $\\
		$ p' $ & Current Number of superior components associated with $ \Gamma_{i} $\\
		$ e_i $ & The $ i $-th PC coefficient\\
		$ \lambda_i $ & The $ i $-th PC variance\\
		$ \mathcal{L} $ & Current number of temporary clusters associated with $ \Gamma $\\
		$ \mathbbl{k} $ & Number of miniclusters which are about to be added to the temporary clustering model\\
		$ \mathbb{K} $ & K-means parameter for the number of clusters\\
		$ \mathbb{K}' $ & Number of obtained subclusters out of the retained set, discovered through DBSCAN\\
		$ \mathbb{K}'' $ & Number of obtained subdivided subclusters out of the retained set, discovered through K-means\\
		$ \mathcal{T} $ & True number of original clusters in $ \mathcal{X} $\\
		$ \nu \in \left\lbrace 1,\cdots,\mathcal{T} \right\rbrace $ & Index to the nearest initial minicluster for a newly discovered subcluster\\
		$ \eta $ & Random sampling rate\\
		$ \Lambda $ & PC total variance ratio for temporary clusters\\
		$ \alpha $ & Membership threshold for temporary clusters\\
		$ \beta $ & Pruning threshold for final clusters\\
		$ \textit{Eps} $ & DBSCAN parameter for neighborhood radius\\
		$ \textit{MinPts} $ & DBSCAN parameter for minimum number of neighbors involving the query point itself\\
		$ \mu_{X} $ & Mean location of cluster $ X $\\
		$ \Sigma_{X} $ & Covariance structure of cluster $ X $\\
		$ \mathbb{S}_{X} $ & Scatter matrix of cluster $ X $\\
		$ \mathcal{A}_X $ & Transformation matrix of cluster $ X $\\
		$ z $ & Object $ x $ in the space of eigenvectors\\
		$ \mu'_{X} $ & Mean location of cluster $ X $ in the space of eigenvectors\\
		$ MD\left( x,X\right) $ & Mahalanobis distance of object $ x $ from cluster $ X $\\
		$ SingCheck\left( \Sigma_{X}\right) $ & A function that checks on the singularity of $ \Sigma_{X} $, and outputs 1 in case of being singular and 0 otherwise\\
		$ CohrCheck\left( X\right) $ & A function that checks on the coherence of the input data $ X $, so that it checks whether only one dense cluster will be discovered through DBSCAN. Its output is 1 in case of being coherent, and 0 otherwise\\
		$ det_{\Sigma_{X}} $ & Covariance determinant of cluster $ X $\\
		$ \vec{\delta} $ & Vector of maximum covariance determinant condition for miniclusters discovered through scalable clustering\\
		$ \left\lbrace \digamma\right\rbrace_{2\times\mathcal{T}} $ & Information array of the final clustering model with 2 properties for each final cluster\\
		$ \mathcal{M}_{i} $ & The $ i $-th final cluster, comprising of the associated temporary clusters\\
		$ \mu_{f} $ & Mean location of a final cluster\\
		$ \Sigma_{f} $ & Covariance structure of a final cluster\\
		$ \mathcal{U} $ & A set of regenerated points\\
		$ \left| \cdot\right| $ & Cardinality of a set of objects\\
		$ \Phi $ & The empty set\\
		$ \kappa $ & A low constant near zero\\
		\bottomrule
	\end{tabular}
\end{table}

The framework of the proposed approach is presented in Algorithm~\ref{algo_SDCOR-Fram}, which consists of three main phases, including: 1) Sampling; 2) Scalable Clustering; and 3) Scoring. All these phases will be described in detail in the next subsections.

\begin{algorithm}
	\DontPrintSemicolon
	\SetAlgoLined
	
	\caption{Framework of SDCOR}
	\label{algo_SDCOR-Fram}
	
	\SetKwInOut{Input}{Input}\SetKwInOut{Output}{Output}
	\Input{$ \left[ \mathcal{X}\right]_{n\times p} $ - The $ n $ by $ p $ input dataset $ \mathcal{X} $; $\eta $ - Random sampling rate; $\Lambda $ - PC total variance ratio; $\alpha $ - Membership threshold; $\beta $ - Pruning threshold}
	\Output{Outlying scores for every object in $ \mathcal{X} $}
	
	\BlankLine
	\textbf{Phase 1 {\textemdash} Sampling:}
	\BlankLine
	
	\textit{Step 1.} Take a random sample $ \mathcal{S} $ from $ \mathcal{X} $ according to the sampling rate $\eta $\;
	\textit{Step 2.} Find the optimal values for the DBSCAN parameters, \textit{Eps} and \textit{MinPts}, required for clustering $ \mathcal{S} $, w.r.t. Appendix \ref{appSec_DBSCANoptmlParamSampl}\;
	\textit{Step 3.} Run DBSCAN on $ \mathcal{S} $ using the obtained optimal parameters, and reserve the count of discovered miniclusters as $ \mathcal{T} $, as the true number of the original clusters in data\;
	\textit{Step 4.} Build the very first array of miniclusters information (temporary clustering model) out of the result of step 3, w.r.t. Algorithm~\ref{algo_MiniClustMake}\;
	\textit{Step 5.} Reserve the covariance determinant values of the initial subclusters as the vector $ \vec{\delta} = \left[ \delta_{1},\cdots,\delta_{\mathcal{T}} \right] $, for the maximum covariance determinant condition\;
	\textit{Step 6.} Clear $ \mathcal{S} $ from RAM and maintain the initial temporary clustering model in the buffer\;
	
	\BlankLine
	\textbf{Phase 2 {\textemdash} Scalable Clustering:}
	\BlankLine
	
	Prepare the input data to be processed chunk by chunk so that each chunk could be fit and processed in the RAM buffer at the same time\;
	\textit{Step 1.} Load the next available chunk of data into RAM\;
	\textit{Step 2.} Update the temporary model of clustering over the contents of the buffer, w.r.t. Algorithm~\ref{algo_MemoProcess}\;
	\textit{Step 3.} If there is any available unprocessed chunk, go to step 1\;
	\textit{Step 4.} Build the final clustering model, w.r.t. Algorithm~\ref{algo_FinalClustBuild}, using the temporary clustering model acquired out of the previous steps\;
	
	\BlankLine
	\textbf{Phase 3 {\textemdash} Scoring:}
	\BlankLine
	
	According to the final clustering model, for each data point $ x\in \mathcal{X} $, employ the Mahalanobis distance criterion to find the closest final cluster, and lastly, assign $ x $ to that cluster and use the criterion value as the object outlierness score\;
	
\end{algorithm}

\subsection{Sampling}\label{subSec_SampPhas}
In this phase, we generate a random sample of the entire dataset. As it might initially seem simple, but in actuality, it could be a complicated task since it is not guaranteed that in certain massive databases, records are not following a specific order in some attribute; this could lead to costly scans of the complete dataset to build an efficient sampling. Above all, without a practical random sampling, the modeled clustering structure may not represent the original one; after that, outliers could be misclassified over scalable clustering. In the following, the percentage of the sampled data is indicated as $\eta $.

\subsubsection{Clustering the sampled data}\label{subSubSec_ClstSmplData}
After obtaining the sampled data, we conduct the DBSCAN algorithm to cluster them w.r.t. the calculated optimal parameters explained in detail in Appendix \ref{appSec_DBSCANoptmlParamSampl}. Here, as we apply ``uniform random sampling'' \textemdash~also known as ``simple random sampling'' or ``random sampling without replacement'' \textemdash~on the entire data, hence there is equal probability for every individual point to be incorporated in the sample \cite{thompson1992sampling}; for this reason, we assume that the number of sampled clusters through this sampling scheme is the same as the number of original clusters, $ \mathcal{T} $, in the main dataset\footnote{We reserve $ \mathcal{T} $ for later use.}. Most of all, this is an accepted truth as also observed in \cite{bradley1998scaling,palmer2000density,kollios2003efficient,wu2006outlier,angiulli2009dolphin,liu2008isolation,liu2012isolation,sugiyama2013rapid,bandaragoda2014efficient,bandaragoda2018isolation}. Besides, we presume that the location (centroid) and the shape (covariance structure) of such subclusters are so close to the original ones. In Section~\ref{sec_ExprResl}, we will show that even by using a low rate of random sampling, the mentioned properties of a sampled cluster could be quite similar to those of the original one\footnote{As we are working in massive scale, there is no need to be worried about minuscule clusters in data which may not have enough sampled points to represent a relatively similar structure to the genuine one; this case generally happens with not very large datasets. Besides, in most cases, tiny bunches of points tend to be outlier classes that should not be able to initiate a cluster.}. The idea behind making these primary subclusters \textemdash~that are so similar to the original clusters in the input dataset in terms of the basic characteristics \textemdash~is that we intend to determine a Mahalanobis radius, which collapses a specific percentage of objects belonging to every original cluster and let other subclusters be created around this folding area during successive memory loads of points. Ultimately, by merging these miniclusters, we will obtain the approximate structure of the original clusters.

\subsubsection{Building the initial clustering model}\label{subSubSec_InitClstModl}
To build the initial clustering model, we need to extract some information from the sampled clusters obtained through DBSCAN and store them in a special array. As stated earlier about the benefit of using the properties of principal components for high-dimensional data, we need to find those PCs that give higher contributions to the cluster representation. To this aim, we sort the PCs on the basis of their corresponding variances in descending order, and then we choose the topmost PCs having a share of the total variance at least equal to $\Lambda $ percent. We call these PCs superior components and denote their number as $p'$. Let $ x $ be an object among the total $ n $ objects in the dataset $ \left[ \mathcal{X}\right]_{n\times p} $, belonging to the temporary cluster $ \left[ \mathbb{X}_{i}\right]_{m\times p} $, then the information about this subcluster as $ \left\lbrace \Gamma_{i}\right\rbrace_{8\times 1} $, in the array of temporary clustering model $ \left\lbrace \Gamma\right\rbrace_{8\times \mathcal{L}} $, is as follows:

\begin{enumerate}\label{enum_MiniClstInfo}
	\item \relax Mean vector in the original space, $\mu_{\mathbb{X}_{i}}=\frac{1}{m}\displaystyle\sum\nolimits_{x\in \mathbb{X}_{i}}x $
	\item \relax Scatter matrix in the original space, $\mathbb{S}_{\mathbb{X}_{i}}=\displaystyle\sum\nolimits_{x\in \mathbb{X}_{i}} (x-\mu_{\mathbb{X}_{i}})^{t}(x-\mu_{\mathbb{X}_{i}}) $
	\item \relax $p'$ superior components, $\left[e_1,\cdots, e_{p'}\right] $, derived from the covariance matrix $\sum_{\mathbb{X}_{i}}=\frac{1}{m-1} {\mathbb{S}}_{\mathbb{X}_{i}} $, which form the columns of the transformation matrix $\mathcal{A}_{\mathbb{X}_{i}} $
	\item \relax Mean vector in the transformed space, $\mu'_{\mathbb{X}_{i}}=\mu_{\mathbb{X}_{i}}\mathcal{A}_{\mathbb{X}_{i}} $
	\item \relax Square root of the top $p'$ PC variances, $\left[\sqrt{\lambda_1},\cdots, \sqrt{\lambda_{p'}}\right] $
	\item \relax Size of the minicluster, $ m $
	\item \relax Value of $p'$
	\item \relax Index to the nearest initial minicluster, $ \nu \in \left\lbrace 1,\cdots,\mathcal{T} \right\rbrace $
\end{enumerate}

Algorithm~\ref{algo_MiniClustMake} demonstrates the process of obtaining and adding this information per each minicluster attained through either of the ``Sampling'' or ``Scalable Clustering'' stages to the temporary clustering model. As for the eighth property of the information array, for every sampled cluster, it is set as the corresponding original cluster number, and for any of the subclusters achieved out of scalable clustering, it is set as the corresponding index to the nearest initial subcluster\footnote{It is clear that all those sampled points, which the initial clustering model is built upon them, will be met again throughout scalable clustering. Nevertheless, as they are randomly sampled, it could be asserted that they will not impair the final clustering model structure.}.

\begin{algorithm}
	\DontPrintSemicolon
	\SetAlgoLined
	
	\caption{$ \left[ \Gamma\right] = $ MiniClustMake($ \Gamma,\Omega,\Lambda,\nu $)}
	\label{algo_MiniClustMake}
	
	\SetKwInOut{Input}{Input}\SetKwInOut{Output}{Output}
	\Input{$ \Gamma $ - Current array of miniclusters information; $ \Omega=\left\lbrace \mathbb{X}_{1},\cdots,\mathbb{X}_{\mathbbl{k}}\right\rbrace $ - A partition of some data points in the memory into some miniclusters; $\Lambda$ - PC share of the total variance; $ \nu $ - Index to the nearest initial minicluster}
	\Output{$ \Gamma $ - Updated temporary clustering model}
	\BlankLine
	
	$ c \leftarrow \mathcal{L} $\;
	\ForEach{minicluster $ \mathbb{X}_{i}, 1\le i \le\mathbbl{k} $}{
		Apply PCA on $\mathbb{X}_i $ and obtain its PC coefficients and variances. Then choose $ p' $ as the number of those top PC variances for which their share of the total variance is at least $\Lambda $ percent\;
		
		$ \Gamma\left\{1,c+i\right\} \leftarrow$ Mean vector of $ \mathbb{X}_i $\;
		$ \Gamma\left\{2,c+i\right\} \leftarrow$ Scatter matrix of $ \mathbb{X}_i $\;
		$ \Gamma\left\{3,c+i\right\} \leftarrow$ Top $ p' $ PC coefficients corresponding to the top $ p' $ PC variances\;
		$ \Gamma\left\{4,c+i\right\} \leftarrow$ Transformed mean vector, as $ \Gamma\left\{1,c+i\right\}\cdot \Gamma\left\{3,c+i\right\} $\;
		$ \Gamma\left\{5,c+i\right\} \leftarrow$ Square root of the top $ p' $ PC variances\;
		$ \Gamma\left\{6,c+i\right\} \leftarrow$ Number of objects in $ \mathbb{X}_i $\;
		$ \Gamma\left\{7,c+i\right\} \leftarrow$ Value of $ p' $\;
		
		\uIf(\tcc*[f]{``Sampling'' stage}){$ c \equiv 0 $}{
			$ \Gamma\left\{8,c+i\right\} \leftarrow i $\;
		}
		\Else(\tcc*[f]{``Scalable Clustering'' stage}){
			$ \Gamma\left\{8,c+i\right\} \leftarrow \nu $\;
		}
	}
	
\end{algorithm}

\subsubsection{Establishing the scalable clustering criteria}\label{subSubSec_IncrClstCrit}
Regarding the major deviation between the sampling and original distributions, DBSCAN cannot be applied to both situations with the same setting. Hence, as it is detailed in Appendix \ref{appSec_DBSCANoptmlParamSampl}, w.r.t. the insignificant effect of the \textit{MinPts} parameter on the final clustering results, we use the same value for it in both sampling and original conditions; however, \textit{Eps} is far more alert to small changes, and thus as it is experimentally justified, we decide to use half of the utilized quantity in the sampling case for the original one during scalable clustering.

Moreover, according to \cite{hubert2010minimum,johnstone2009sparse}, when a multivariate Gaussian distribution is contaminated with some outliers, then the corresponding covariance determinant is no longer robust and is significantly more than that of the main cluster. Following this contamination, the corresponding (classical) Mahalanobis contour line\footnote{Since the terms ``Mahalanobis contour line'' and ``tolerance ellipse'' are the same in essence, thus we will use them indifferently in this paper.} will also become broader than that of the pure cluster (robust one), as it contains also abnormal data. So, it makes sense that there is a direct relationship between the value of covariance determinant of an arbitrary cluster and the wideness of its tolerance ellipse, which could also be referred to as the spatial volume of the cluster. Moreover, by being contaminated, this volume could increase and become harmful.

Since during scalable clustering, new objects are coming over time, and miniclusters are growing gradually, so, it is possible for a minicluster with an irregular shape to accept some outliers. Then, the following covariance matrix will no longer be robust, and the corresponding Mahalanobis contour line will keep getting wider too, which could cause the absorption of many other outliers. Therefore, to impede the creation process of these voluminous non-convex subclusters, which could be contaminated with myriad outliers, we have to put a limit on the covariance determinant of every subcluster, which is discovered through scalable clustering. Here, we follow a heuristic approach and employ the covariance determinant of the nearest initial subcluster obtained out of the ``Sampling'' stage as the limit. %Thus, we reserve the covariance determinant values of the initial miniclusters for later use.

This problem that outliers, through a detection procedure, could be included in normal clusters and no longer be identified is called the ``masking effect''. Note that we are using the mentioned constraints only when subclusters are created for the first time, not while growing over time. The justification is that while an object is about to be assigned to a minicluster, the other constraint on the Mahalanobis radius, within reason, is hindering outliers from being accepted as a member. In other words, when the Mahalanobis distance of an outlier is more than the predefined radius threshold, it cannot be assigned to the subcluster, if and only if that threshold is set to a fair value\footnote{The three-sigma rule of thumb or the same empirical 68-95-99.7 rule can be taken into account to determine the requisite reasonable thresholds.}. Hence, we do not check on the covariance determinant of subclusters while they are growing.

Here, the first phase of the proposed approach is finished, and we need to clear the RAM buffer of any sampled data and only maintain the very initial information obtained about the existing original clusters.

\subsection{Scalable Clustering}\label{subSec_ScalClstPhas}
During this phase, we have to process the entire dataset chunk-by-chunk, as for each chunk, there is enough space in the memory for both loading and processing it all at the same time. After loading each chunk, we update the temporary clustering model according to the data points currently loaded in the memory from the present chunk or retained from the other previously loaded chunks. Finally, after processing the entire chunks, the final clustering model is built out of the temporary clustering model. A detailed description of this phase is provided as follows.

\subsubsection{Updating the temporary clustering model w.r.t. the contents of the buffer}\label{subSubSec_UpdTmpClstModlContBuff}
After loading each chunk into the memory, the temporary clustering model is updated upon the objects belonging to the currently loaded chunk and other ones sustained from the formerly loaded data. First of all, the algorithm checks for the possible membership of each point of the present chunk to any existing minicluster in the temporary clustering model\footnote{After this step, the unassigned objects of the lastly and previously loaded chunks will be altogether considered as the retained or sustained objects in the buffer.}.

Then, after the probable assignments of the current chunk points, some primary and secondary information of the modified subclusters shall be updated. After this update, the structure of the altered subclusters will change, and thus they might still be capable of absorbing more inliers. Therefore, the algorithm checks again for the likely memberships of sustained points in the memory to the updated subclusters. This updating and assignment checking cycle will keep going until there is not any retained point that could be assigned to an updated minicluster.

When the membership evaluation of the present chunk and retained objects is carried out, the algorithm tries to cluster the remaining sustained objects in the memory, regarding the density-based clustering criteria, which have been constituted at the ``Sampling'' phase. After the new miniclusters were created out of the last retained points, there is this probability that some sustained inliers in the buffer might not be capable of actively participating in forming new subclusters because of the density-based clustering standards, though could be assigned to them, considering the firstly settled membership measures. Hence, the algorithm goes another time in the cycle of assignment and updating procedure, like what was done in the earlier steps.

Algorithm~\ref{algo_MemoProcess} demonstrates the steps necessary for updating the temporary clustering model, w.r.t. an already loaded chunk of data and other undecided objects retained in the memory from before. The following subsections will explain the details of this algorithm.

\begin{algorithm}
	\DontPrintSemicolon
	\SetAlgoLined
	
	\caption{$ \left[ \Gamma,\Re\right] = $ MemoProcess($ \mathcal{Y},\Gamma,\alpha,\Re,\vec{\delta},\Lambda $)}
	\label{algo_MemoProcess}
	
	\SetKwInOut{Input}{Input}\SetKwInOut{Output}{Output}
	\Input{$ \mathcal{Y} $ - A chunk of data; $ \Gamma $ - Current array of miniclusters information; $ \alpha $ - Membership threshold; $ \Re $ - Retained set; $ \vec{\delta} $ - Covariance determinant threshold; $\Lambda$ - PC share of the total variance}
	\Output{$ \Gamma $ - Updated temporary clustering model; $ \Re $ - Modified retained set}
	\BlankLine
	
	\tcc{Trying to assign each tuple of a chunk to a minicluster}
	$ \gamma \leftarrow \left\lbrace 1,\cdots,\mathcal{L}\right\rbrace $\;
	$ \left[ \Gamma,\gamma,\Re\right] = $ MiniClustUpdate($ \mathcal{Y},\Gamma,\gamma,\alpha,\Re $)\;
	
	\tcc{Checking out the retained set}
	\If{$ \left| \Re\right|\neq 0 $}{
		\tcc{Checking on the retained set membership for the recently updated miniclusters}
		$ \left[ \Gamma,\Re\right] = $ RetSetMemb($ \Re,\Gamma,\gamma,\alpha $)\;
		
		\tcc{Clustering the retained set}
		\If{$ \left| \Re\right|\neq 0 $}{
			$ l\leftarrow \mathcal{L} $\;
			$ \left[ \Gamma,\Re\right] = $ RetSetClust($ \Re,\Gamma,\vec{\delta},\Lambda $)\;
			$ \gamma \leftarrow \left\lbrace l+1,\cdots,\mathcal{L}\right\rbrace $\;			
			
			\tcc{Checking on the retained set membership for the recently created miniclusters}
			\If{$ \left| \Re\right|\neq 0 $}{
				$ \left[ \Gamma,\Re\right] = $ RetSetMemb($ \Re,\Gamma,\gamma,\alpha $)\;
			}
		}
	}

\end{algorithm}

\paragraph{Trying to assign each tuple of the chunk to a minicluster}\label{parg_AsgnDatmChnkMinClst}
After loading each chunk of data into the buffer, we need to use the properties of PCs for each minicluster and transform each data tuple into the new space of that minicluster; then, like BFR, we calculate the Mahalanobis distance using the mean vector and the square root of variances, but in the space of eigenvectors. That is,

\begin{equation}
\label{equ_MahalDistObjTempClust}
MD(x,X_{i})=\sum_{j=1}^{p'}(\frac{z_{j}-\mu'_j}{\sqrt{\lambda_j}})^{2}
\end{equation}

Where $ MD(x,X_{i}) $ is the Mahalanobis distance of object $ x $ from minicluster $ X_{i} $; $ z=x \mathcal{A}_{X_{i}} $ is the object in the eigenvector space of the minicluster, and $ z_j $ is its $ j $-th component; $ \mu'_j $ and $ \lambda_j $ are respectively the $ j $-th components of the mean vector and the variance vector in the space of eigenvectors; finally, $ p' $ is the number of superior components associated with the minicluster. As stated above, in this style, the amount of computations is sensibly less than if we would have used matrix inversion to calculate the distance. Moreover, w.r.t. \cite{leskovec2014mining}, the accepted Mahalanobis radius in the eigenvector space of the relevant subcluster will be the product of the membership threshold and the square root of the number of dimensions in the transformed space, as $\alpha\sqrt{p'} $.

For each data point, w.r.t. Eq. (\ref{equ_MahalDistObjTempClust}), we need to find the closest minicluster and check whether or not it falls in the accepted Mahalanobis threshold of that minicluster; if it does, some information connected to the corresponding subcluster shall be updated.

\begin{algorithm}
	\DontPrintSemicolon
	\SetAlgoLined
	
	\caption{$ \left[ \Gamma,\gamma',\Re\right] = $ MiniClustUpdate($ \mathbb{X},\Gamma,\gamma,\alpha,\Re $)}
	\label{algo_MiniClustUpdate}
	
	\SetKwInOut{Input}{Input}\SetKwInOut{Output}{Output}
	\Input{$ \mathbb{X} $ - A set of some points in the buffer; $ \Gamma $ - Current array of miniclusters information; $ \gamma $ - List of indices to the recently created or updated miniclusters associated with $ \Gamma $, to be checked on for membership; $ \alpha $ - Membership threshold; $ \Re $ - Retained set}
	\Output{$ \Gamma $ - Updated temporary clustering model; $ \gamma'\subseteq\gamma $ - Modified list of indices to the recently updated miniclusters; $ \Re $ - Modified retained set}
	\BlankLine
	
	\tcc{Updating the primary information of subclusters}
	\ForEach{$ x\in\mathbb{X} $}{
		$ b \leftarrow \argmin_{i\in \gamma} MD(x,X_{i}) $\;
		%		$ b \leftarrow \underset{i\in \gamma}{\mathrm{argmin}} $ $ MD(x,X_{i}) $\;
		\uIf{$ MD(x,X_{b})\le \alpha\sqrt{\Gamma_{b}\left\lbrace 7\right\rbrace} $}{
			$ \Gamma_{b}\left\lbrace 2\right\rbrace \leftarrow \Gamma_{b}\left\lbrace 2\right\rbrace + x'x$\;
			$ \Gamma_{b}\left\lbrace 6\right\rbrace \leftarrow \Gamma_{b}\left\lbrace 6\right\rbrace + 1$\;
			Remove $ x $ from the RAM buffer\;
		}
		\Else{
			$ \Re \leftarrow \Re\cup x $\;
		}
	}
	
	\tcc{Updating the secondary information of subclusters}
	$ \gamma' \leftarrow \Phi $\;
	\ForEach{$ X_{i},i\in \gamma $}{
		\If{$ X_{i} $ has accepted any new members}{
			Obtain its updated covariance matrix $ \sum_{X_{i}} $, through normalizing its updated scatter matrix, w.r.t. the current size of the minicluster as $ \left( \frac{1}{\Gamma_{i}\left\lbrace 6\right\rbrace -1 }\right)\cdot \Gamma_{i}\left\lbrace 2\right\rbrace $\;
			Apply PCA on $ \sum_{X_{i}} $ to acquire its eigenvalues and eigenvectors, and then, update $ \Gamma_{i} $ as follows:\;
			\Indp $ \Gamma_{i}\left\lbrace 7\right\rbrace \leftarrow$ Value of $ p' $ as the updated number of superior components\;
			$ \Gamma_{i}\left\lbrace 3\right\rbrace \leftarrow$ Updated superior coefficients\;
			$ \Gamma_{i}\left\lbrace 4\right\rbrace \leftarrow$ Updated transformed mean vector, as $ \Gamma_{i}\left\lbrace 1\right\rbrace\cdot \Gamma_{i}\left\lbrace 3\right\rbrace $\;
			$ \Gamma_{i}\left\lbrace 5\right\rbrace \leftarrow$ Square root of the updated superior variances\;
			\Indm$ \gamma' \leftarrow \gamma'\cup i $\;
		}
	}

\end{algorithm}

\paragraph{Updating primary and secondary information of the modified temporary clusters}\label{parg_UpdPrmSecInfoTempClst}
For every subcluster that any new objects have been assigned to it, there are two types of information that need to be updated; primary and secondary. Primary information comprises the scatter matrix of the subcluster and its cardinality, which should be updated after each individual assignment. To update the scatter matrix, the outer product of the belonged data point with itself is added to the current scatter matrix; for the cardinality, the number of objects assigned to the subcluster is increased by one. Every object, after joining a subcluster, is removed from the buffer; otherwise, it will be retained to be decided on later.

After checking on the membership of all points and updating the primary information of the modified subclusters, for each minicluster that has accepted any new members, its PC properties, which are considered as its secondary information, must be updated too. For this purpose, according to the minicluster size, we normalize its scatter matrix in an unbiased manner to acquire its covariance matrix. Then, by applying PCA on this matrix, we update the transformation matrix, the mean vector in the space of eigenvectors, and the superior PC variances of the minicluster. Algorithm~\ref{algo_MiniClustUpdate} demonstrates the required steps for finding the closest subcluster per every query point w.r.t. the relevant limitations, and updating the essential information of every modified subcluster.

\paragraph{Trying to assign retained objects in the buffer to the newly updated miniclusters}\label{parg_AssgnRetObjBuffNewUpdMinClst}
After updating the secondary information of each altered subcluster, the corresponding tolerance ellipse will rotate and might also become skewed a bit around the centroid; in other words, the following accepted Mahalanobis neighborhood is modified. Hence, w.r.t. Algorithm~\ref{algo_RetSetMemb}, it would be necessary to check on the objects retained in the buffer, i.e., whether they can belong to a modified minicluster; and if it is so, the corresponding minicluster information needs to be updated, w.r.t. Algorithm~\ref{algo_MiniClustUpdate}. Nevertheless, this is not the end; by keeping up this cycle of membership checking and minicluster updating, more and more objects could be assigned to miniclusters and then be discarded. In this iterative manner, after each iteration, memory contents should be evaluated using only the updated subclusters from the last iteration; thus, the list of updated miniclusters will be shrinking over time until it becomes an empty list. This means that there is not any other sustained object in the buffer, which falls in the accepted Mahalanobis threshold of any of the updated miniclusters, or every retained object has been eventually assigned to an updated minicluster. Here, by employing this procedure, it seems that the tolerance ellipses of miniclusters are sweeping inliers through the cycle of assignment and updating.

\begin{algorithm}[t!]
	\DontPrintSemicolon
	\SetAlgoLined
	
	\caption{$ \left[ \Gamma,\Re\right] = $ RetSetMemb($ \Re,\Gamma,\gamma,\alpha $)}
	\label{algo_RetSetMemb}
	
	\SetKwInOut{Input}{Input}\SetKwInOut{Output}{Output}
	\Input{$ \Re $ - Retained set; $ \Gamma $ - Current array of miniclusters information; $ \gamma $ - List of indices to the recently created or updated miniclusters associated with $ \Gamma $, to be checked on for membership; $ \alpha $ - Membership threshold}
	\Output{$ \Gamma $ - Updated temporary clustering model; $ \Re $ - Modified retained set}
	\BlankLine
	
	\If{$ \left| \gamma\right| \neq 0 $}{
		\While{true}{
			$ \left[ \Gamma,\gamma,\Re\right] = $ MiniClustUpdate($ \Re,\Gamma,\gamma,\alpha,\Phi $)\;
			\If{$ \left| \gamma\right| \equiv 0 $}{break\;}
		}
	}
	
\end{algorithm}

\begin{figure}[pos=t!]
	\centering
	\includegraphics[width=1\linewidth]{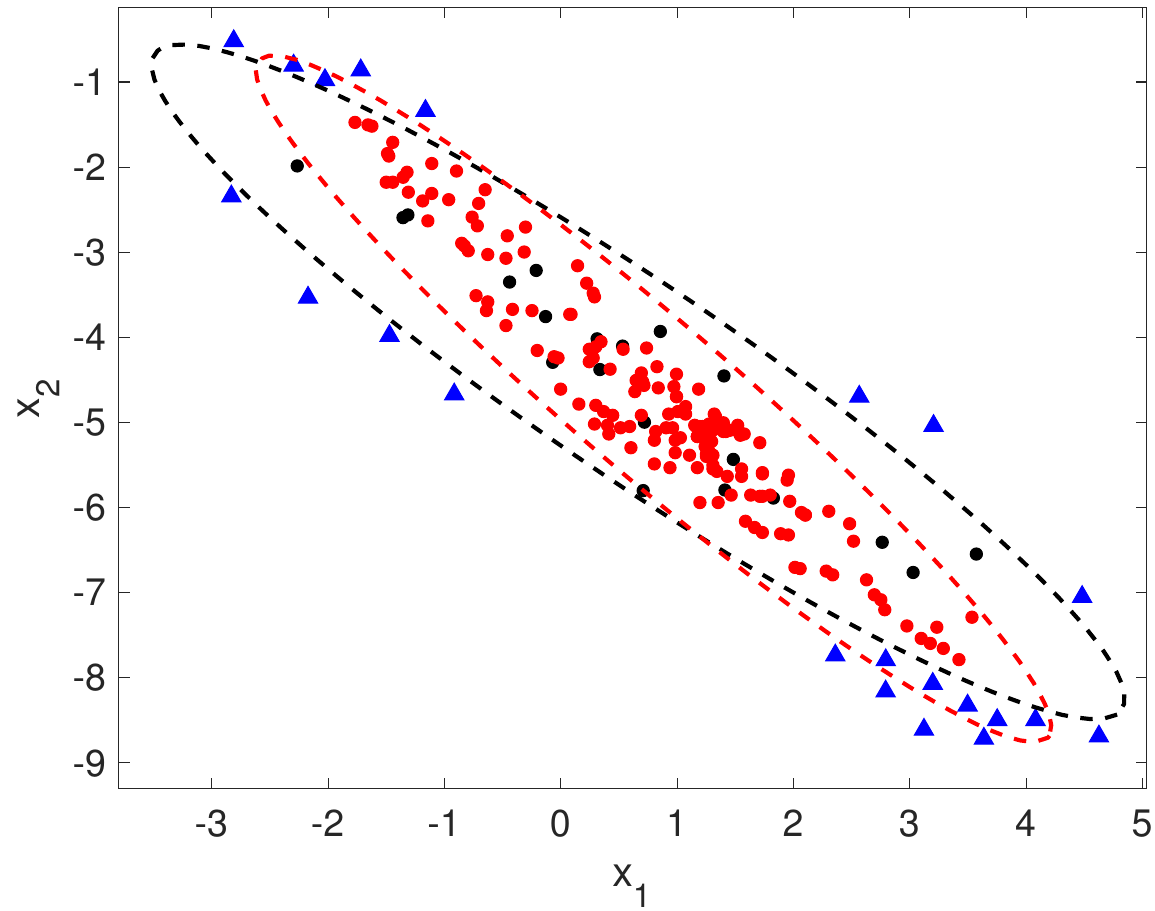}
	\caption{Assigning retained objects in the buffer to the newly updated subclusters}
	\label{figure-AssgnRetObjsRAMUpdSubClust}
\end{figure}

Fig.~\ref{figure-AssgnRetObjsRAMUpdSubClust} shows the scenario in which, after updating the core information of a subcluster, its Mahalanobis neighborhood is modified; some objects which were not able to belong to this subcluster now are capable of being assigned to it. Black circle points and the black dashed line tolerance ellipse respectively represent a subcluster and its Mahalanobis neighborhood. Red circle points represent objects assigned to the subcluster during the last memory process, as they reside in its accepted neighborhood. The red dashed line represents the updated tolerance ellipse of the updated subcluster. Moreover, blue triangle points represent objects which could be assigned to the updated subcluster if they would lie in its updated Mahalanobis radius.

\paragraph{Clustering retained objects in the buffer}\label{parg_ClstRetObjBuff}
Here, after checking on the membership of every data tuple stored in the memory, we afford to cluster retained objects in the RAM buffer, using the DBSCAN algorithm again. However, as emphasized at the ``Sampling'' stage, according to the significant difference in the density of the sampled and original data, we have to employ a specific configuration for each situation.

Furthermore, as it was described before, it is possible that some miniclusters could be discovered during scalable clustering by DBSCAN, which are suffering from the singularity problem. Thus, for handling such a situation, there are some ways. One is to use the pseudoinverse of the covariance structure, but it is not totally accurate. The better way is to disregard such minicluster and let its points still be in the memory, to be resolved later, among other coming points. Therefore, for every discovered subcluster, we shall check on its covariance matrix, whether or not it is singular; in case of singularity, we disregard that subcluster.

Now, w.r.t. this prementioned matter that, we have to put a limit on the boundaries of the miniclusters which are being created throughout scalable clustering, we are going to demonstrate with an intuitive example that if a minicluster with a non-convex shape is formed, how outliers could be absorbed to such irregular minicluster and cause severe damage to the final clustering results.

\begin{figure}[pos=t!]
	\centering
	\includegraphics[width=1\linewidth]{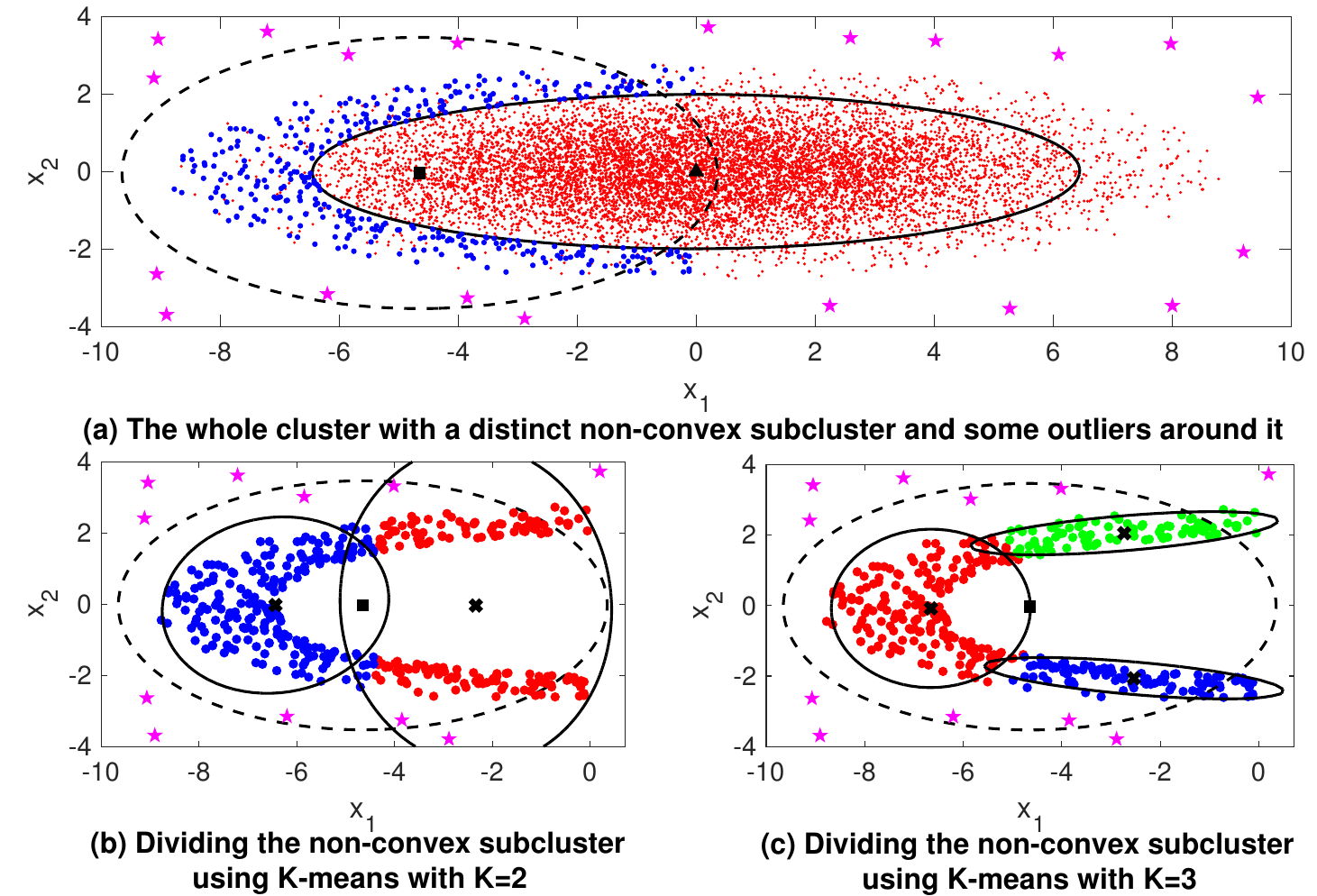}
	\caption{Breaking a non-convex minicluster with a very wide tolerance ellipse into smaller pieces by the K-means algorithm\protect\footnotemark}
	\label{figure-BreakNonConvxClustSmallPiecKmns}
\end{figure}
\footnotetext{Regarding the three-sigma rule of thumb, the Mahalanobis radii equal to 1 and 2, in order, cover roughly 68 and 95 percent of the total objects in a Gaussian distribution. For convex-shaped clusters of other distributions, the amount of coverage might vary, but for non-convex-shaped clusters, it could contain objects not belonging to the distribution. Here, in all subfigures, the presented radius is equal to 1.5.}

Fig.~\ref{figure-BreakNonConvxClustSmallPiecKmns}a illustrates the structure of an original cluster represented with red dots, with a newly discovered non-convex subcluster shown with blue dots, and a black square and a black dashed line as its centroid and accepted tolerance ellipse respectively. The irregular minicluster is formed around the initial minicluster, while its centroid and accepted Mahalanobis radius are denoted as a black triangle and a solid black line, respectively. There are also some local outliers around the original cluster, which are illustrated with magenta pentagons\footnote{Here, for challenging the performance of our method, we are taking outliers so close to the original cluster. However, in reality, it is not usually like that, and outliers often have a significant distance from every normal cluster in data.}.

As it is evident, the irregular minicluster can absorb some local outliers, as its tolerance ellipse is covering a remarkable space out of the containment area by the original cluster. Moreover, the covariance determinant value for the irregular minicluster is equal to 15.15, which is almost twice that of the initial minicluster equal to 7.88. Thus, to fix this concern, by considering the proportion between the covariance determinant of an arbitrary cluster and its spatial volume, we decide to divide the irregular minicluster into smaller coherent pieces, with more minor covariance determinants and more limited Mahalanobis radii as well. Therefore, we heuristically set the threshold value for the covariance determinant of any newly discovered subcluster or subdivided subcluster, as that of the nearest initial minicluster\footnote{This threshold is denoted as $ \delta_i, 1 \le i \le \mathcal{T} $, for any subcluster discovered near the $ i $-th initial minicluster, through scalable clustering. The proximity measure for this nearness is the Mahalanobis distance of the newly discovered subcluster centroid from the initial miniclusters. We assume any subcluster with a covariance determinant greater than such threshold, as a candidate for a non-convex subcluster, whose spatial volume could cover some significant space out of the scope of the related original cluster.}.

For the division process of an irregular subcluster, we prefer to adopt the K-means algorithm. However, K-means can cause some incoherent subdivided subclusters in such cases\footnote{As K-means focuses solely on finding the best locations for the means and does not consider the cohesion of the output clusters.}, as shown in Fig.~\ref{figure-BreakNonConvxClustSmallPiecKmns}b. In Fig.~\ref{figure-BreakNonConvxClustSmallPiecKmns}b, two smaller miniclusters, produced as the K-means result, are represented in different colors, with covariance determinants of 1.63 and 7.71 for the coherent blue and incoherent red miniclusters, respectively. The associated centroids and tolerance ellipses are denoted as black crosses and black solid lines, respectively. It is clear that even incoherent subdivided subclusters, with a covariance determinant less than or equal to the predefined threshold though, could be hazardous as non-convex subclusters with a high value of covariance determinant, as their tolerance ellipses could get out of the scope of the following main cluster, and suck outliers in\footnote{However, the divided minicluster could be significantly smaller in size and determinant, though because of the lack of coherency, some PC variances could be substantially larger than others; therefore, the regarding tolerance ellipse will be more stretched in those PCs, and thus harmful.}.

An alternative for this is to use hierarchical clustering algorithms, which typically present a higher computational load than K-means. For this matter, we decide to adopt a K-means variant, which, for every subdivided subcluster obtained through K-means, we apply DBSCAN again to verify its cohesion.
% that no more than one coherent cluster will be discovered.
Finally, we increase the value of $ \mathbb{K} $ for K-means, from 2 till a value for which\footnote{Here, the upper bound for $ \mathbb{K} $ is $ \lfloor \left|\Re_i\right|/\left( p+1\right)\rfloor $, to avoid the singularity problem for every subdivided subcluster of retained objects. $ \left|\Re_i\right| $ stands for the cardinality of the $ i $-th minicluster of retained points.}, three conditions for every subdivided subcluster are met: to be coherent, not to be singular, and having a covariance determinant less than or equal to $ \delta_i $.

Fig.~\ref{figure-BreakNonConvxClustSmallPiecKmns}c illustrates a scenario in which three smaller miniclusters are represented as the K-means output in different colors; centroids and tolerance ellipses are shown as in Fig.~\ref{figure-BreakNonConvxClustSmallPiecKmns}b. The covariance determinant values are equal to 0.12, 1.02, and 0.10 for respectively the green, red and blue subdivided subclusters. As it is obvious, all subdivided subclusters are coherent and not singular, with much smaller determinants than the threshold and much tighter spatial volumes as such. Ultimately, after attaining acceptable subclusters, it is time to update the temporary clustering model w.r.t. them, in regard to Algorithm~\ref{algo_MiniClustMake}.

Algorithm~\ref{algo_RetSetClust} shows all the steps required to cluster the data elements retained in the memory buffer. First, DBSCAN is applied to the retained set with the optimal parameters concerning the original distribution; some miniclusters, maybe along with some noisy points, will be identified. Then for every detected minicluster out of the sustained objects, we evaluate whether it qualifies for being considered as a valid minicluster conforming with the three conditions of being coherent, not being singular, and not exceeding the determinant threshold. At first, the singularity term is checked out, and in the case of being singular, the related minicluster points are discarded; otherwise, the closest initial minicluster will be found through Mahalanobis distance, and the corresponding covariance determinant will be exerted as the required constraining threshold.

If the determinant limitation is not violated, then such minicluster is qualified for being added to the temporary clustering model, and thereafter, its points in the memory will be obviated. Conversely, in the instance of determinant violation, the K-means variant is utilized for the division strategy, as for every subdivided subcluster, the mentioned three indispensable conditions shall be met. In case that K-means fails, the correspondent unbroken minicluster is discarded; oppositely, if K-means succeeds, the authentic subdivided subclusters are appended to the temporary clustering model. At last, the initial retained set passed to Algorithm~\ref{algo_RetSetClust} is replaced with the entire undecided objects after the described procedure.

\begin{algorithm}
	\DontPrintSemicolon
	\SetAlgoLined
	
	\caption{$ \left[ \Gamma,\Re\right] = $ RetSetClust(
		$ \Re,\Gamma,\vec{\delta},\Lambda $)}
	\label{algo_RetSetClust}
	
	\SetKwInOut{Input}{Input}\SetKwInOut{Output}{Output}
	\Input{$ \Re $ - Retained set; $ \Gamma $ - Current array of miniclusters information; $ \vec{\delta} $ - Covariance determinant threshold; $\Lambda$ - PC share of the total variance}
	\Output{$ \Gamma $ - Updated temporary clustering model; $ \Re $ - Modified retained set}
	\BlankLine
	
	Apply the DBSCAN algorithm to $ \Re $ concerning the optimum parameters for the original distribution explained in Appendix \ref{appSec_DBSCANoptmlParamSampl}. Consider the result of such clustering as $ \left\lbrace \Re_1,\cdots, \Re_{\mathbb{K}'}\right\rbrace \cup \Re_{\zeta} $\;
	
	\tcc{Adding the information of the newly discovered miniclusters to the temporary clustering model}
	\ForEach{$ \Re_i,1\le i\le \mathbb{K}' $}{
		\If(\tcc*[f]{Singularity check}){$ SingCheck\left( \Sigma_{\Re_i}\right) \equiv 1 $}{
			$ \Re_{\zeta} \leftarrow \Re_{\zeta} \cup \Re_i $\;
			continue\;
		}
		
		$ \nu \leftarrow \argmin_{h \in \left\lbrace 1,\cdots,\mathcal{T} \right\rbrace} MD(\mu_{\Re_i},X_{h}) $\;
		\uIf(\tcc*[f]{Irregular minicluster}){$ det_{\sum_{\Re_i}} > \vec{\delta}\left( \nu \right) $}{
			Apply K-means with the number of clusters $ 2 \le \mathbb{K}''\le \lfloor \left|\Re_i\right|/\left( p+1\right)\rfloor $ to $ \Re_i $. Find the minimum value for $ \mathbb{K}'' $, as by which for every subdivided subcluster $ \Re_{i,j},1\le j\le \mathbb{K}'' $, we have $ CohrCheck\left( \Re_{i,j}\right) \equiv 1 $, $ SingCheck\left( \Sigma_{\Re_{i,j}}\right) \equiv 0 $ and $ det_{\sum_{\Re_{i,j}}} \le \vec{\delta}\left( \nu \right) $\;
			\If{such $ \mathbb{K}'' $ is not found}{
				$ \Re_{\zeta} \leftarrow \Re_{\zeta} \cup \Re_i $\;
				continue\;
			}
			$ \left[ \Gamma\right] = $ MiniClustMake($ \Gamma,\left\lbrace \Re_{i,1},\cdots, \Re_{i,\mathbb{K}''}\right\rbrace,\Lambda,\nu $)\;
			Remove $ \left\lbrace \Re_{i,1},\cdots, \Re_{i,\mathbb{K}''}\right\rbrace $ from the RAM buffer\;
		}
		\Else(\tcc*[f]{Regular minicluster}){
			$ \left[ \Gamma\right] = $ MiniClustMake($ \Gamma,\Re_i,\Lambda,\nu $)\;
			Remove $ \Re_{i} $ from the RAM buffer\;
		}
	}
	
	\tcc{Setting the unresolved points as the retained set}
	$ \Re \leftarrow \Re_{\zeta} $\;
	
\end{algorithm}

\paragraph{Trying to assign retained objects in the buffer to the newly created miniclusters}\label{parg_AssgnRetObjBuffNewCrtMinClst}
After checking on retained objects in the buffer in case of being capable of forming a new minicluster, it would be necessary to examine the remaining retained objects once more, w.r.t. Algorithm~\ref{algo_RetSetMemb}; i.e., it ought to be inspected whether or not they could be assigned to the newly created miniclusters, in the same cycle of membership checking and minicluster updating, like what was done in Subsection~\ref{parg_AssgnRetObjBuffNewUpdMinClst}. The reason for this concern is that, due to limitations connected to the utilized density-based clustering algorithm, such objects may not have been capable of being an active member of any of the newly created subclusters, even though they lie in the associated accepted Mahalanobis radius. Hence, it becomes essential to check again the assignment of these latter retained objects.% to those latter temporary clusters, which have been discovered through the latter process of the memory contents.

\begin{figure}[pos=t!]
	\centering
	\includegraphics[width=1\linewidth]{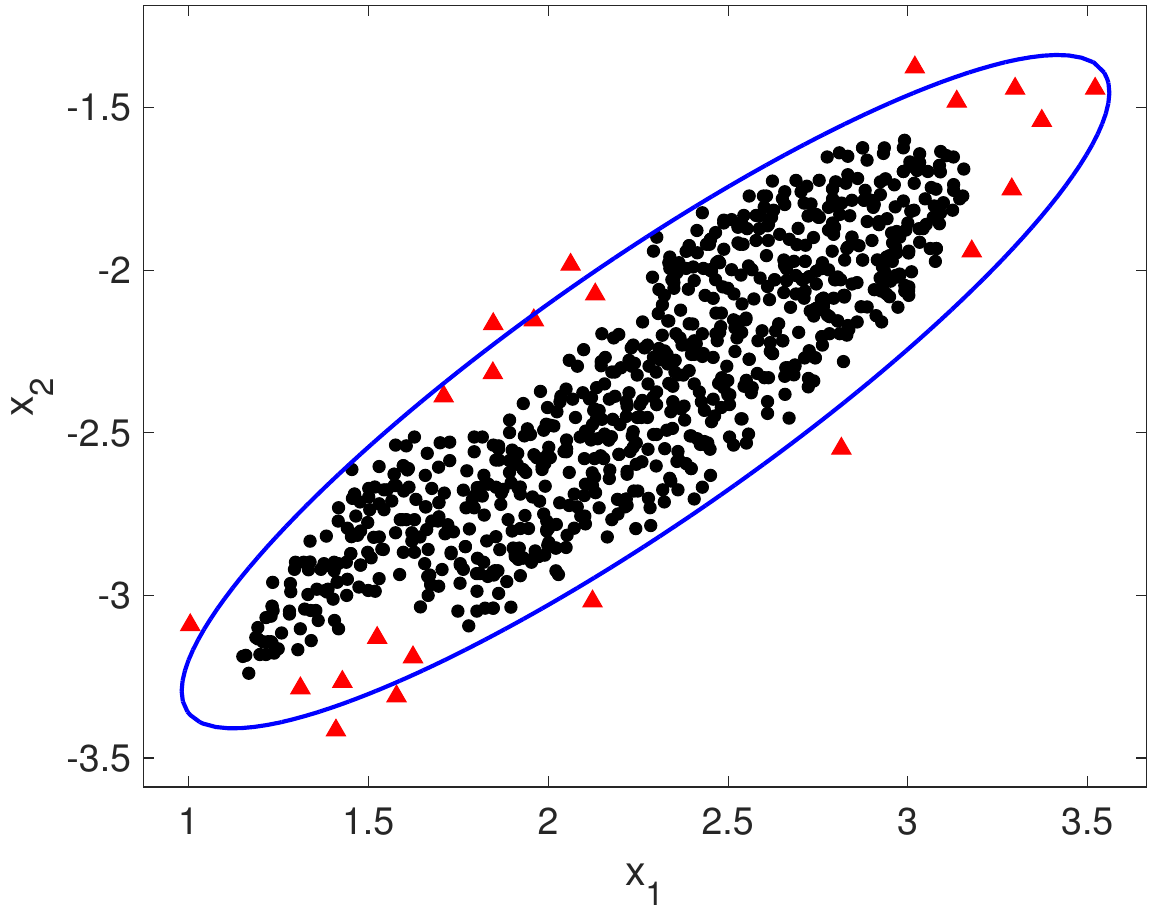}
	\caption{Assigning retained objects in the buffer to the newly created subclusters}
	\label{figure-CheckUltRetPnts}
\end{figure}

Fig.~\ref{figure-CheckUltRetPnts} demonstrates an intuitive example of such a situation in which some objects according to the density restrictions cannot be assigned to a cluster, even though they lie in the accepted Mahalanobis radius of that cluster. Objects that have had the competence to form a cluster are shown with solid black circles; those that are not a part of the cluster, but reside in its accepted Mahalanobis radius represented by a solid blue line, are denoted as red triangles.

Thus, if retained objects could belong to a newly created subcluster, the corresponding information of that subcluster will be modified w.r.t. Algorithm~\ref{algo_MiniClustUpdate}; otherwise, such data will be still retained in the buffer for further process. However, if it were the last chunk that was processed, all these retained objects would be marked as temporary outliers. But, all of these temporary outliers are not true outlying points; as stated earlier, some of them are normal objects that have not found the competence of forming a minicluster or being assigned to one due to applied restrictions. Anyway, at last, all of these true and untrue anomalies which are maintained in the buffer will be discarded, and this is only the temporary clustering model which is remained after all.

\subsubsection{Building the final clustering model}\label{subSubSec_BuldFinlClstModl}
Here, at the end part of scalable clustering, we follow Algorithm~\ref{algo_FinalClustBuild} to construct the final clustering model or the same approximate structure of $ \mathcal{T} $ original clusters, w.r.t. the temporary clustering model with $ \mathcal{L} $ miniclusters.

\begin{algorithm}
	\DontPrintSemicolon
	\SetAlgoLined
	
	\caption{$ \left[ \digamma\right] = $ FinalClustBuild($ \Gamma,\eta,\beta $)}
	\label{algo_FinalClustBuild}
	
	\SetKwInOut{Input}{Input}\SetKwInOut{Output}{Output}
	\Input{$ \Gamma $ - Current array of miniclusters information; $ \eta $ - Random sampling rate; $ \beta $ - Pruning threshold}
	\Output{$ \digamma $ - Final clustering model}
	\BlankLine
	
	Consider the arrangement of the entire temporary clusters $ X_{j}\text{'s},1\le j\le \mathcal{L} $ w.r.t. the original clusters in the main distribution as $ \left\lbrace \mathcal{M}_1,\cdots,\mathcal{M}_\mathcal{T}\right\rbrace $; $ \mathcal{M}_i, 1\le i\le \mathcal{T} $ stands for the $ i $-th final cluster, consisting of the associated temporary clusters as every temporary cluster possesses a property declaring the original cluster it belongs to\;
	$ \digamma \leftarrow \Phi $\;
	
	\ForEach{$ \mathcal{M}_i,1\le i\le\mathcal{T} $}{
		\uIf(\tcc*[f]{Isolated minicluster}){$ \left| \mathcal{M}_i\right| \equiv 1 $}{Use the same mean location and covariance structure of the isolated subcluster, as for those of the final cluster\;}
		\Else(\tcc*[f]{Group of miniclusters}){
			\tcc{Calculating the final mean location}
			$ \mu_{f} \leftarrow \frac{\sum\nolimits_{j\colon X_{j}\in \mathcal{M}_i} \left[ \Gamma_{j}\left\lbrace 6\right\rbrace\cdot\Gamma_{j}\left\lbrace 1\right\rbrace \right]}{\sum\nolimits_{j\colon X_{j}\in \mathcal{M}_i}\Gamma_{j}\left\lbrace 6\right\rbrace} $\;
			
			\tcc{Calculating the final covariance structure}
			$ \mathcal{U} \leftarrow \Phi $\;
			\ForEach{$ X_{j}\in \mathcal{M}_i $}{
				Regenerate $ \eta\cdot\Gamma_{j}\left\lbrace 6\right\rbrace $ number of fresh data points, with Gaussian distribution, with regard to $ \Gamma_{j}\left\lbrace 1\right\rbrace $ as the temporary mean and $ \left( \frac{1}{\Gamma_{j}\left\lbrace 6\right\rbrace - 1}\right)\cdot\Gamma_{j}\left\lbrace 2\right\rbrace $ as the temporary covariance matrix\;
				Add these regenerated points to $ \mathcal{U} $\;
			}
			Calculate the Mahalanobis distance of the points in $ \mathcal{U} $ based upon the sample mean $ \mu_{\mathcal{U}} $, and the sample covariance matrix $ \Sigma_{\mathcal{U}} $\;
			Prune $ \mathcal{U} $ by discarding those points with a Mahalanobis distance more than $ \beta\sqrt{p} $, and recalculate $ \Sigma_{\mathcal{U}} $\;
			$ \Sigma_{f} \leftarrow \Sigma_{\mathcal{U}} $\;
			
			Remove $ \mathcal{U} $ from the buffer\;			
			
			\tcc{Adding information to the final clustering model}
			$ \digamma\left\lbrace 1,i\right\rbrace \leftarrow \mu_{f} $\;
			$ \digamma\left\lbrace 2,i\right\rbrace \leftarrow \Sigma_{f} $\;
			
		}
	}
	
\end{algorithm}

For every individual minicluster in the temporary clustering model, we have already realized to which original cluster in the main distribution it belongs; thus, to obtain the final clustering model, we just need to merge the information of the relevant miniclusters to each of the original clusters to acquire the final clusters. Here, we presume that the core information of each final cluster only consists of a centroid $ \mu_{f} $, and a covariance matrix $ \Sigma_{f} $. Hence, w.r.t. Algorithm~\ref{algo_FinalClustBuild}, if a final cluster contains only one temporary cluster, then the final centroid and the final covariance matrix will be the same as for the temporary cluster. Otherwise, in the case of containing more than one temporary cluster, we utilize the sizes and centroids of the associated miniclusters to obtain the final mean.

For acquiring the final covariance matrix of a final cluster comprised of multiple temporary clusters, for each of the associated miniclusters and w.r.t. its centroid and covariance structure, we afford to regenerate a specific amount of fresh data points with Gaussian distribution. We define the regeneration size of each minicluster equal to the product of the sampling rate \textemdash~which was used at the ``Sampling'' stage \textemdash~and the cardinality of that minicluster, as $\eta\left|X_{j}\right| $; this is necessary for saving some free space in the memory while regenerating the approximate structure of an original cluster. We consider all regenerated objects of all subclusters belonging to a final cluster as a unique and coherent cluster and afford to obtain the final covariance structure out of it.

But before using the covariance matrix of such regenerated cluster, we need to mitigate the effect of some generated outliers, which could be created unavoidably during the regeneration process, and can potentially prejudice the final accuracy outcomes. For this purpose, we need to prune this transient final cluster, according to the Mahalanobis threshold $\beta $, obtained through the user. Thus, regenerated objects having a Mahalanobis distance of more than $\beta\sqrt p $ from the regenerated cluster will be obviated. Now, we can compute the ultimate covariance matrix out of such pruned regenerated cluster and then remove this transient cluster. This procedure is conducted in sequence for every final cluster, consisting of more than one temporary cluster.

\begin{figure*}[pos=t!]
	\centering
	\includegraphics[width=1\linewidth]{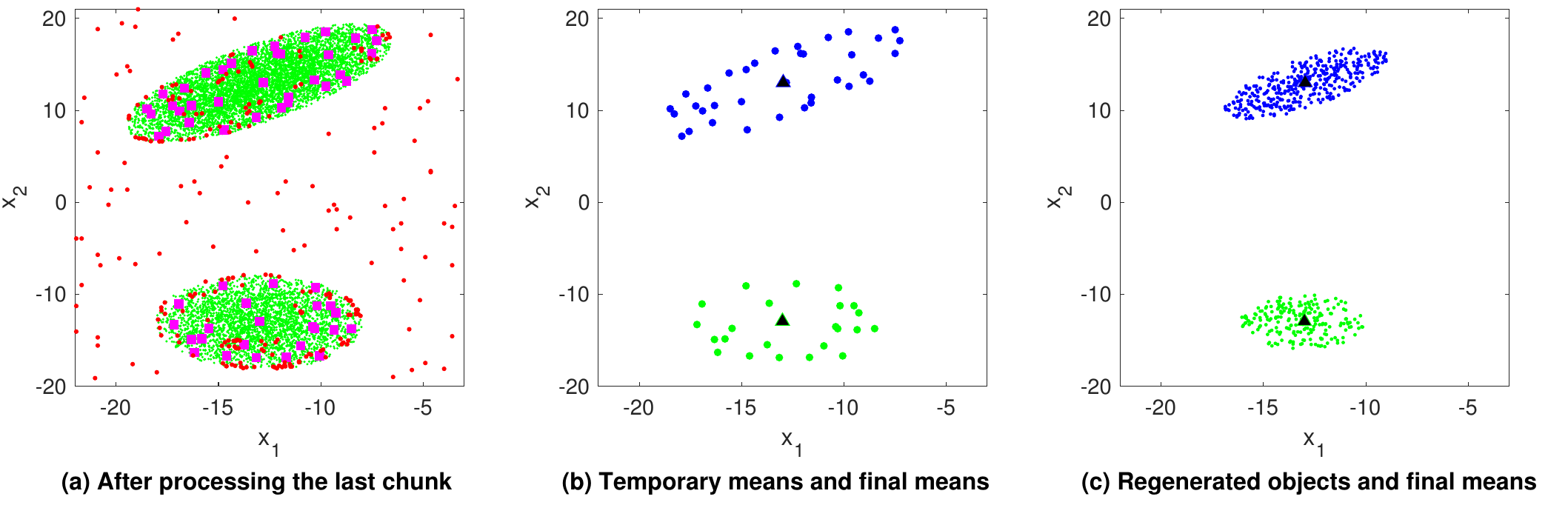}
	\caption{Proposed method appearance at the last steps of scalable clustering}
	\label{figure-PropMethApprncLastStepScalClust}
\end{figure*}

Fig.~\ref{figure-PropMethApprncLastStepScalClust}a demonstrates a dataset consisting of two dense Gaussian clusters with some local outliers around them. This figure is, in fact, a sketch of what the proposed method looks like at the final steps of scalable clustering and before building the final clustering model. Green dots are normal objects belonged to a minicluster, and red dots are temporary outliers; magenta square points represent the temporary centroids. Fig.~\ref{figure-PropMethApprncLastStepScalClust}b demonstrates both temporary means and final means, represented by solid circles and triangles respectively, with a different color for each final cluster. Fig.~\ref{figure-PropMethApprncLastStepScalClust}c colorfully demonstrates the pruned regenerated data points for every final cluster besides the final means, denoted as dots and triangles, respectively.

Now, after obtaining the final clustering model, the second phase of the proposed approach is finished. In the following subsection, the third phase, named ``Scoring'' is presented, and we will describe how to give every object a score of outlierness, w.r.t. the final clustering model obtained out of scalable clustering.

\subsection{Scoring}\label{subSec_ScorPhas}
In this phase, w.r.t. the final clustering model obtained through scalable clustering, we give each data point an outlying rank. Therefore, like phase two, once more, we need to process the whole dataset in chunks and use the same Mahalanobis distance criterion to find the closest final cluster to each object. This local Mahalanobis distance \cite{aggarwal2015data} is assigned to the object as its outlying score; the higher the distance, the more likely it is that the object is an outlier. Here, we name such score obtained out of our proposed approach, SDCOR, which stands for ``Scalable Density-based Clustering Outlierness Ratio''\footnote{Due to the high computational load associated with calculating the Mahalanobis distance in high dimensions, one can still gain the benefit of using the properties of principal components for computing the outlying scores, like what was done during scalable clustering.}.

\subsection{Algorithm complexity}\label{subSec_AlgoCmplx}
Here, at first, we analyze the time complexity of the proposed approach. For the first two phases, ``Sampling'' and ``Scalable Clustering'', the most expensive operations are the application of DBSCAN to the objects residing in the memory after loading each chunk, and the application of PCA to each minicluster to obtain and update its secondary information.

Let $n_\mathfrak{C} $ be the number of data points contained in a chunk. Considering the three-sigma rule of thumb, in every memory load of points, the majority of these points lies in the accepted Mahalanobis neighborhood of the current temporary clusters and is being assigned to them \textemdash~and this will escalate over time by the increasing number of subclusters, which are being created during every memory process \textemdash~and also, by utilizing an indexing structure for the \textit{k}NN queries, the time complexity of the DBSCAN algorithm will be $O\left(\kappa n_\mathfrak{C}\log\left(n_\mathfrak{C}\right)\right) $; where $ \kappa $ is a low constant close to zero. Note that exercising an indexing structure to decrease the computational complexity of answering the \textit{k}NN queries is only applicable in lower dimensions; for high-dimensional data, we need some sort of sequential scans to handle the \textit{k}NN queries, which leads to average time complexity of $O\left(\kappa n_\mathfrak{C}^2\right) $ for DBSCAN.

Applying PCA on miniclusters is $O\left(min\left(p^{3},n_\mathfrak{C}^{3}\right)\right) $ \cite{johnstone2009sparse}, as $ p $ stands for the dimensionality of the input dataset. But according to our necessary assumption that $p<n_\mathfrak{C} $, thus applying PCA will be $O\left(p^{3}\right) $. Hence, the first two phases of the algorithm will totally take $O\left(max\left(\kappa n_\mathfrak{C}^2,p^{3}\right)\right) $.

The last phase of the algorithm, w.r.t. this concern that only consists of calculating the Mahalanobis distance of the total \textit{n} objects in the input dataset to $ \mathcal{T} $ final clusters, is $O\left(n\mathcal{T}\right) $; regarding that $\mathcal{T}\ll n $, hence, the time complexity of this phase will be $O\left(n\right) $. The overall time complexity is thus at most $O\left(max\left(\kappa n_\mathfrak{C}^2,p^{3}\right)+n\right) $. However, in addition to $ \kappa $ being a truly tiny constant, it is evident that both $ p $ and $n_\mathfrak{C} $ values are negligible w.r.t. \textit{n}; therefore, we can state that the time complexity of our algorithm is linear.

Analysis of the algorithm space complexity is twofold. First, we consider the stored information in the memory for the clustering models, and second, the amount of space required for processing the resident data in RAM. Concerning the fact that the most voluminous parts of the temporary and of the final clustering models are the scatter and covariance matrices, respectively, and that $ \mathcal{L}\gg\mathcal{T} $, thus, the space complexity of the first part will be $O\left(\mathcal{L}p^{2}\right) $. For the second part, according to this matter that in each memory load of points, the most expensive operations belong to the clustering algorithms, DBSCAN and K-means, and regarding the linear space complexity of these methods, hence, the overall space complexity will be $O\left(n_\mathfrak{C}+\mathcal{L}p^{2}\right) $.

\section{Experimental evaluation}\label{sec_ExprEval}
This section supplies details on the efficacy, efficiency, and other specific tests employed in our experiments. For the effectiveness test, we analyze the accuracy and stability of the proposed method on some real-life and synthetic datasets, compared to some other state-of-the-art and well-known distance-, density-, and clustering-based methods, which are all subsets of the proximity-based methods for outlier identification. The representative competing strategies are namely ORCA, DOLPHIN, and \textit{S\textsubscript{p}} as the distance-based methods; LOF, LoOP, and an ensemble version of LOF (introduced in \cite{bandaragoda2018isolation}) called EnLOF\footnote{It should be noted that another ensemble version of LOF has been presented by Zimek et al. \cite{zimek2013subsampling}, though with different parameterization. However, this approach produces similar outputs to EnLOF, it is empirically proven in \cite{bandaragoda2018isolation} that this peculiar assembly adaptation of LOF suffers from a much higher arithmetical cost, even in most cases in comparison with the original LOF method.} as the density-based techniques; and finally, a fast K-means variant optimized for clustering large-scale data, named X-means \cite{pelleg2000x} as the only clustering-based approach in our analysis\footnote{Actually, X-means is not an anomaly recognition technique in essence as it assumes the input data free of noise. Therefore, after obtaining the final clustering outcome through this method, the Euclidean distance of every object to its closest centroid is assigned to it as an outlier score; hence, the following assessment measures could be calculated.}, for which we will report a special benchmarking against SDCOR w.r.t. some exclusive clustering robustness criteria.

Moreover, for the competing methods that are based on some sort of random sampling and present non-deterministic results over various iterations \textemdash~namely \textit{S\textsubscript{p}}, EnLOF, and SDCOR \textemdash~we provide some statistical significance test to find differences among them across several attempts. Consequently, for the efficiency test, experimentation is conducted on some artificial datasets to assess how the final accuracy varies when the number of outliers increases. Besides, more investigations are carried out to appraise the scalability of the proposed algorithm and the effect of random sampling rate on the final detection validness.

All the experiments were executed on a laptop having a 2.5 GHz Intel Core i5 processor, 6 GB of memory, and a Windows 7 Ultimate (Service Pack 1) operating system. Furthermore, in what follows, first of all, we explain in detail the utilized evaluation metrics and the analyzed real-world and artificial datasets in our analysis. Moreover, we provide some ins and outs on the algorithms implementations and the user-defined parameters for all contending methods.

\subsection{Evaluation metrics}\label{subSec_EvalMetr}

\subsubsection{General-purpose metrics}\label{subSubSec_UnivMetr}
We evaluate the functionality of the competing outlier detection algorithms in general by cross-checking two widely-used metrics in this area \cite{campos2016evaluation,domingues2018comparative}. One is the Receiver Operating Characteristic, also the Receiver Operating Curve (ROC), or the same curve of detection rate and false alarm rate (true positive rate against false positive rate), calculated concerning Eqs. (\ref{equ_TPR-Recall}) and (\ref{equ_FPR}); the other is the Precision-Recall (PR) curve, computed in respect of Eqs. (\ref{equ_Prec}) and (\ref{equ_TPR-Recall}). In both metrics, the positive class (class of interest) stands for the outliers, while the negative class typifies the normal samples. The examination is then established on the Area Under the Curve (AUC) for both metrics, which are called, in order, AUROC (Area Under Receiver Operating Curve) and AUPRC (Area Under Precision-Recall Curve); more importantly, it should be pointed out that throughout the literature, AUROC is much more common and popular than AUPRC for comparing the anomaly identification tasks. The related equations for computing the referred measures are as follows:

\begin{equation}
	\label{equ_TPR-Recall}
	\textit{TPR (Recall)} = \frac{\textit{true positives}}{\textit{true positives} + \textit{false negatives}}
\end{equation}

\begin{equation}
	\label{equ_FPR}
	\textit{FPR} = \frac{\textit{false positives}}{\textit{false positives} + \textit{true negatives}}
\end{equation}

\begin{equation}
	\label{equ_Prec}
	\textit{Precision} = \frac{\textit{true positives}}{\textit{true positives} + \textit{false positives}}
\end{equation}

\subsubsection{Clustering validity metrics}\label{subSubSec_ClustValidMetrc}
Besides the above-mentioned two popular assessment measures for unsupervised outlier detection methodologies, as the proposed method is essentially founded upon the cluster analysis, and also, X-means is the only clustering-based competing method in our experiments, hence we desire to compare these two techniques, in particular, utilizing some external clustering validity measures employed in \cite{aliguliyev2009performance,alguliyev2017anomaly}\footnote{The distinction between the internal and the external clustering validity metrics is that the internal indices assess the inherent unification/separation of the output clusters, while the external indices appraise the similarity/dissimilarity between the clustering solution out of a peculiar clustering function and the expected results or the same ``ground truth'' of the corresponding dataset.}.

Here, we exert five specific measures: Purity \cite{rubinov2006classes}, the Mirkin metric \cite{mirkin1996mathematical}, F-measure \cite{van2004geometry}, Entropy \cite{boutin2004cluster}, and Variation of Information (VI) \cite{meilua2003comparing} to weigh up the difference among the clustering results of SDCOR and X-means. These measures are divided into two separate collections. In one category \textemdash~containing Purity, the Mirkin metric, and F-measure \textemdash~it depends on computing the number of pairs of points on which two clustering outcomes coincide/collide with each other. The other category \textemdash~including Entropy and VI \textemdash~is information-reliant, and more precisely, it is the homogeneity of a clustering solution that goes under evaluation. Furthermore, each one of these measures happens at a certain value as its best condition; some fall into a minimum equal to 0, and some drive into a maximum equal to 1 in the ideal case. Consider that dataset $ \mathcal{X} $ with $ n $ objects composes, in reality, of classes $ D = \left( D_1,\cdots D_l\right) $, and the result of applying an arbitrary clustering approach on $ \mathcal{X} $ is as $ C = \left( C_1,\cdots C_k\right) $; in both the real situation and the clustering output, anomalies, like a group of normal objects, are categorized as a single group of elements\footnote{SDCOR and X-means both provide anomaly scores for all the normal and abnormal elements in the input data, and even anomalies are finally assigned to a discovered cluster. Therefore, for distinctly defining the resultant outliers cluster, we utilize the existing ground truth on the actual number of outliers (o) in the correspondent dataset; scores are sorted in descending order, and then the top-o indices will be separated as the anomaly cluster.}. More details on the noted external clustering accuracy metrics are presented in the following.

\paragraph{Purity}\label{parg_CVMpurity}
Every output cluster $ C_r, 1\le r \le k $ may include elements associated with multiple target classes. The Purity of an arbitrary cluster is the ratio of the size of the domineering class in the cluster to the cluster cardinality, delineated as follows:

\begin{equation}
	\label{equ_Cpurity}
	\textit{Purity}\left( C_r\right) = \frac{1}{\left| C_r \right|} \smash{\displaystyle\max_{s = 1,\cdots,l}} \left| C_r \cap D_s \right|,\quad 1 \le r \le k
\end{equation}

The Purity value of a single cluster always falls in the interval $ \left[ \frac{1}{l},1\right] $; higher Purity rates denote that the relevant cluster is a more ``pure'' version of the domineering class. The Purity of the entire assortment of the detected clusters is computed as a weighted aggregation of the respective clusters purities, defined as follows:

\begin{equation}
	\label{equ_purity}
	\begin{aligned}
		\textit{Purity}\left( C\right) = \displaystyle\sum\limits_{r=1}^k \frac{\left| C_r\right|}{n} \textit{Purity}\left( C_r\right) = \\ \frac{1}{n}\displaystyle\sum\limits_{r=1}^k \max_{s = 1,\cdots,l} \left| C_r \cap D_s \right|
	\end{aligned}
\end{equation}

In the best condition, Purity is equal to 1. Greater values for this measure imply that the equivalent clustering result is more accurate; terrible clusterings possess Purity values near 0. However, Purity is not totally dependable for the accuracy evaluation of a clustering method, as a deficient result cannot absolutely lead to the inefficacy of the related method.

\paragraph{Mirkin metric}\label{parg_CVMmirkin}
The Mirkin metric is computed as follows:

\begin{equation}
	\label{equ_mirkinMetric}
	\begin{aligned}
		M \left( C,D\right) = \frac{1}{n^2}
		\biggl( \displaystyle\sum\limits_{r=1}^k{\left| C_r \right|^2} +
		\displaystyle\sum\limits_{s=1}^l{\left| D_s \right|^2} \\
		- 2\displaystyle\sum\limits_{r=1}^k \displaystyle\sum\limits_{s=1}^l{\left| C_r \cap D_s \right|^2} \biggr)
	\end{aligned}
\end{equation}

Concerning this formula, the Mirkin metric is limited to the range of $ \left[ 0,1 \right] $, and in the best case where the clustering solution and the ground truth are identical, it is equal to 0.

\paragraph{F-measure}\label{parg_CVMfmeasure}
F-measure is a widely-used measure in the clustering accuracy appraisal tasks, which is intrinsically dependent on the information-retrieval concepts. For comparing two partitions of clusters and classes through this metric, an easy way would be calculating the Precision (P), Recall (R), F-value (F), and ultimately the consequent F-measure for every cluster w.r.t. the regular classes, and at last, we just need to compute the total F-measure for the entire clusters collection. The formulation of this metric is as follows:

\begin{equation}
	\label{equ_FmeasPrec}
	\begin{aligned}
		P\left( C_r,D_s \right) = \frac{\left| C_r \cap D_s \right|}{\left| C_r \right|}, \quad 1 \le r \le k, 1 \le s \le l
	\end{aligned}
\end{equation}

\begin{equation}
	\label{equ_FmeasRecl}
	\begin{aligned}
		R\left( C_r,D_s \right) = \frac{\left| C_r \cap D_s \right|}{\left| D_s \right|}, \quad 1 \le r \le k, 1 \le s \le l
	\end{aligned}
\end{equation}

Concerning the already-mentioned formulas, it can be understood that Precision and Recall describe, in order, how homogeneous and complete the cluster $ C_r $ is with regard to the class $ D_s $. After computing the Precision and Recall values for a specific cluster with reference to various classes, the correspondent F-values will be the harmonic mean of the respective Precisions and Recalls, in the following terms:

\begin{equation}
	\label{equ_FmeasFval}
	\begin{aligned}
		& F\left( C_r,D_s \right) = \biggl( \frac{\frac{1}{P\left( C_r,D_s \right)} + \frac{1}{R\left( C_r,D_s \right)}}{2} \biggr)^{-1} = \\
		& \frac{2 P\left( C_r,D_s \right) R\left( C_r,D_s \right)}{P\left( C_r,D_s \right) + R\left( C_r,D_s \right)}, \quad 1 \le r \le k, 1 \le s \le l
	\end{aligned}
\end{equation}

The succeeding F-measure\footnote{Here as we employ the traditional form of F-measure which is the harmonic mean of Precision and Recall, we denote it as $ F_1 $.} quantity for the cluster $ C_r $ is the maximal F-value attained at the whole category of classes $ D = \left( D_1,\cdots D_l\right) $. That is:

\begin{equation}
	\label{equ_FmeasCr}
	\begin{aligned}
		F_1 \left( C_r \right) = \smash{\displaystyle\max_{s = 1,\cdots,l}} F \left( C_r,D_s \right),\quad 1 \le r \le k
	\end{aligned}
\end{equation}

Finally, the F-measure value for the entire clusters collection is calculated as the weighted sum of the distinctive clusters F-measures, denoted in this way:

\begin{equation}
	\label{equ_Fmeas}
	\begin{aligned}
		F_1 \left( C\right) = \displaystyle\sum\limits_{r=1}^k \frac{\left| C_r\right|}{n} F_1 \left( C_r\right)
	\end{aligned}
\end{equation}

F-measure is always positive, and at the worst condition, it falls into 0; it happens at 1 as its best case, and high values for it connote better clusterings. This measure has an essential superiority over the Purity and the Entropy criteria, as it takes into account both the homogeneity (conformity) and the completeness (integrity) of the clustering solution.

\paragraph{Entropy}\label{parg_CVMentropy}
The Entropy metric is an information-theoretic benchmark, and particularly, it is more reliable than the Purity metric. The Entropy of the cluster $ C_r $ is defined as follows:

\begin{equation}
	\label{equ_entrpCr}
	\begin{aligned}
		E\left( C_r \right) = -\frac{1}{\log\left( l \right)} \displaystyle\sum\limits_{s=1}^l \frac{\left| C_r \cap D_s \right|}{\left| C_r \right|} \log \biggl( \frac{\left| C_r \cap D_s \right|}{\left| C_r \right|} \biggr), \\	1 \le r \le k
	\end{aligned}
\end{equation}

Note that for conditions that $ x $ equals 0 in $ x \log \left( x \right) $, we consider the result as 0. Furthermore, as it was remarked, Entropy is a more all-embracing measure than Purity because it considers the distribution of the entire classes in a certain cluster. The precedent term in Eq. (\ref{equ_entrpCr}), $ -\frac{1}{\log\left( l \right)} $, is a normalization term to restrict the Entropy values to the interval $ \left[ 0,1 \right] $; however, the Entropy of a random cluster is contrary to its Purity, as a Purity of 1 indicates that the cluster is identical to the respective class, but for Entropy, it is reversed. Finally, the global Entropy of the corresponding clustering product is computed as below:

\begin{equation}
	\label{equ_entrp}
	\begin{aligned}
		E\left( C\right) = \displaystyle\sum\limits_{r=1}^k \frac{\left| C_r\right|}{n} E\left( C_r\right)
	\end{aligned}
\end{equation}

The global Entropy quantities also fall in the range of $ \left[ 0,1 \right] $. As it is obvious, a flawless clustering is the one in which every single cluster incorporates elements from only a unique class; the global Entropy in such an ideal condition is equivalent to 0. Generally, the lower the value of Entropy, the more unerring the clustering outcome is.

\paragraph{Variation of Information (VI)}\label{parg_CVMvi}
This measure is another information-theoretic-based tool like Entropy to evaluate clustering solutions. VI appraises the proportion of information that one gains and loses while transferring from one partition to another. VI is delineated in the literature in various forms, though we use the following formula:

\begin{equation}
	\label{equ_VI}
	\begin{aligned}
		& \textit{VI}\left( C,D \right) = \\
		& \frac{1}{n \log\left( n \right)} \displaystyle\sum\limits_{r=1}^k \displaystyle\sum\limits_{s=1}^l \left| C_r \cap D_s \right| \log \biggl( \frac{\left| C_r \right| \left| D_s \right|}{\left| C_r \cap D_s \right|^2} \biggr)
	\end{aligned}
\end{equation}

The maximum value of VI relies on $ n $, although in the optimum state, it comes to 0.

\subsection{Datasets description}\label{subSec_DataDscr}
Our evaluation employs various real-world and artificial data in diverse situations. These data collections range in cardinality from about 11,000 to more than 2,000,000 points and in dimensionality from 3 to 60 features. In the following, all of the datasets utilized in our analysis are explained in detail.

\begin{table}[!ht]
	\caption{Properties of the datasets used in the efficacy experiments}
	\label{table-DataProp}
	\centering
	\arrayrulecolor{black}
	\setlength{\tabcolsep}{4.6pt}
	\begin{tabular}{llllll}
		\toprule
		& \multicolumn{1}{l}{Dataset} & \multicolumn{1}{l}{\#n} & \multicolumn{1}{l}{\#p} & \multicolumn{1}{l}{\#o} & \multicolumn{1}{l}{\%o} \\
		\midrule
		\multicolumn{1}{l}{\multirow{9}[2]{*}{\parbox[l]{1cm}{Real Datasets}}} & Mammography & 11,183 & 6     & 260   & 2.32 \\
		& Adult & 38,323 & 6     & 1,168 & 3.05 \\
		& Shuttle & 49,097 & 9     & 3,511 & 7.15 \\
		& Smtp  & 95,156 & 3     & 30    & 0.03 \\
		& Skin  & 199,283 & 3     & 5,085 & 2.55 \\
		& CreditCardFraud & 284,807 & 29    & 492   & 0.17 \\
		& ForestCover & 286,048 & 10    & 2,747 & 0.96 \\
		& Http  & 567,498 & 3     & 2,211 & 0.39 \\
		& Hepc  & 2,003,171 & 7     & 5,123 & 0.26 \\
		\midrule
		\multicolumn{1}{l}{\multirow{4}[2]{*}{\parbox[l]{1cm}{Synth. Datasets}}} & Data1 & 500,000 & 30    & 5,000 & 1.00 \\
		& Data2 & 1,000,000 & 40    & 10,000 & 1.00 \\
		& Data3 & 1,500,000 & 50    & 15,000 & 1.00 \\
		& Data4 & 2,000,000 & 60    & 20,000 & 1.00 \\
		\bottomrule
	\end{tabular}%
\end{table}%

\subsubsection{Real-life datasets}\label{subSubSec_RealDataDscr}
Some public and large-scale real benchmark datasets, mostly taken from the UCI repository \cite{Dua:2019}, and some others from the OpenML \cite{vanschoren2014openml} and Kaggle \cite{kaggle} libraries are used in our experiments; they are representatives of different domains in science, society, and humanities. The upper part of Table~\ref{table-DataProp} gives a summary of the properties of these real-world data series. This table shows the characteristics of all the important test datasets, namely the numbers of objects (\#n), attributes (\#p), and outliers (\#o); in addition, for outliers in each dataset, their share of total objects in the corresponding dataset is reported in percentage terms (\%o). Among the noted nine real datasets, except for \textit{Mammography} and \textit{CreditCardFraud} which can be attained through OpenML and Kaggle respectively, the other seven can be directly reached through UCI\footnote{\textit{Smtp} and \textit{Http} are derived from the \textit{KDD Cup 1999} dataset, available on UCI.}. Moreover, \textit{Mammography}, \textit{Shuttle}, \textit{Smtp}, \textit{ForestCover}, and \textit{Http} are preprocessed and labeled by ODDS library \cite{Rayana:2016}; the other 4 have been curated under our reflections which will be clarified in subsequent.

The \textit{Adult} dataset is extracted out of the 1994 Census database, and the prediction problem is to define whether a person earns over \$50 thousand a year. This dataset contains 14 attributes in total which we consider the six continuous ones here to represent the data; the >\$50K class is down-sampled to 10\% to exemplify the outlier class.

The ``Skin Segmentation Data Set'', which herein is referred to as the \textit{Skin} dataset, is collected out of various face images of diverse age and race groups, and different genders. Two classes are available, namely skin and non-skin, in which the skin class is down-sampled to 10\% to typify the anomaly class.

The ``Credit Card Fraud Detection'' data collection\footnote{The dataset has been gathered and evaluated over a research teamwork of Worldline and the Machine Learning Group of ULB (Université Libre de Bruxelles) on big data mining and fraud detection. It can be directly acquired through the Kaggle library, although the original source is this link: \href{www.ulb.ac.be/di/map/adalpozz/data/creditcard.Rdata}{www.ulb.ac.be/di/map/adalpozz/data/creditcard.Rdata}.} which here is denoted as the \textit{CreditCardFraud} dataset, comprises transactions made in two days under credit cards in September 2013 by European cardholders. The dataset is exceedingly unbalanced, and the anomaly class (frauds) stands for 0.17\% of all transactions; furthermore, we have discarded the first attribute, which represents the time of the transaction occurrence w.r.t. the first transaction in the dataset.

The ``Individual household electric power consumption Data Set'' which herein appears as the \textit{Hepc} dataset, holds mensuration of some electrical equipment in a house beside the date and time of the measurement. The dataset consists of 9 features, including the two temporal ones, which we omit here; about 1.25\% of the samples suffer from missing values which are discarded consequently. This dataset does not possess any labeling; hence we follow a subtle heuristic as employed in \cite{zimek2013subsampling} to label data objects as inliers/outliers. For an arbitrary dataset with $ p $ dimensions, the Mahalanobis distances of all points in each cluster w.r.t. the corresponding sample mean and sample covariance matrix, follow a $ \chi ^2 $ distribution with $ p $ degrees of freedom. The points exhibiting a distance more than the theoretical 0.975 quantile are indicated as candidate outliers; then, the nominated outliers pack is down-sampled to 10\% to represent the anomaly class.

\subsubsection{Synthetic datasets}\label{subSubSec_SynthDataDscr}
Various experiments have been conducted on variant simulated datasets in our study. All of these artificial data series have been created under an ideal setting and follow the strong assumptions of our algorithm, including the one that the structure of existing clusters should be Gaussian due to the use of the Mahalanobis distance measure, and also, the meaningful proportion between the dimensionality and cardinality of every dataset to evade the singularity problem throughout the clustering course of action. Moreover, the generated outliers are typically more outstanding than those in the real data, and the outliers ``truth'' can be utilized to evaluate whether an outlier algorithm is eligible to locate them.

In more detail, for each dataset having $ p $ dimensions, we build Gaussian clusters with arbitrary mean vectors so that they are quite far away from each other to hinder possible overlappings among the multidimensional clusters. As for the covariance matrix, first, we create a matrix ${\left[\mathbb{A}\right]}_{p\times p} $, whose elements are uniformly distributed in the interval [0,1]. Then, we randomly select half the elements in $ \mathbb{A} $ and make them negative. Finally, the corresponding covariance matrix is obtained in the form of ${\left[\sum\right]}_{p\times p}=\mathbb{A}^{T}\mathbb{A} $. Now, w.r.t. the mean vector (location) of each cluster and its covariance matrix (shape of the cluster), we can generate an arbitrary number of data points from a $ p $-variate Gaussian distribution. Moreover, to eliminate marginal noisy objects in each cluster, we can exploit the Mahalanobis distance criterion and eliminate those objects outside the Mahalanobis radius, set to, e.g., 1.

For injecting local outliers to each cluster, first, we consider a hypercube covering the boundaries of the corresponding cluster in every dimension, with the same centroid as of the cluster and having a specific amount of vacant space around it. Then, we randomly generate records in this space and accept them as local outliers if and only if they fall in the accepted Mahalanobis distance interval of the cluster, e.g., $\left[4\sqrt p,6\sqrt p\right] $. Since the hypercube volume increases so rapidly by $ p $ for high dimensions, we need to extend the accepted interval for generated points to save time. Therefore, some of the synthesized outliers in this way could be global. However, as mentioned earlier, the concept of local outliers covers global ones; hence, each global outlier could be presumed as a local outlier too, but not vice versa.

\subsection{Algorithms implementations and parameters}\label{subSec_ParamSelec}
Except for ORCA, DOLPHIN, and X-means which are implemented by the genuine authors in C/C++, we have implemented all of the other five competing methods, including the proposed method, in MATLAB R2016b (version 9.1)\footnote{For the sake of reproducibility, our code is published on GitHub: \href{https://github.com/sana33/SDCOR}{https://github.com/sana33/SDCOR}. Moreover, we were cautious about the efficient array-based implementation techniques in MATLAB, especially in the case of the FOR/WHILE loops, to avoid excessive execution runtimes; in such a case, it could be meaningfully compared to other fast and cost-effective implementations like C/C++ or Java.}. In what follows, for every method in our analysis, some essential matters on the parameter setting are presented.

\subsubsection{ORCA}\label{subSubSec_ORCAparamSetng}
In ORCA, the parameter $ k $ denotes the number of nearest neighbors, which by using higher values for that, the execution time will also increase. Here, the suggested value for $ k $, equal to $ 5 $, is utilized in all of our experiments. The parameter $ N $ specifies the maximum number of anomalies to be reported. If $ N $ is set to a small value, then ORCA increases the running cut-off quickly, and therefore, more searches will be pruned off, which will result in a much faster runtime. Hence, as the correct number of anomalies is not assumed to be foreknown in the algorithm, we set $ N=\frac{n}{8} $, as a reasonable value, where $ n $ stands for the cardinality of the input data; in the case of the efficiency test where the outliers proportion is foreknown as much higher than usual, we set $ N=\frac{n}{2} $.

However, ORCA does not report an anomaly score for the rest of the data, and this is as long as for computing the AUC values, there is an indispensable need to report an anomaly score for every instance; thus, we set a score equal to zero for other non-anomaly reported objects.

\subsubsection{DOLPHIN}\label{subSubSec_DOLPHINparamSetng}
For executing the DOLPHIN method on any query data, we need two distinct input parameters specialized for every dataset and two other general parameters that could be established globally in all experiments. The two specific parameters are, namely, $ R $, the neighborhood radius, and $ k $, the minimum neighborhood cardinality required for an object to be identified as an inlier. The other two general parameters are viz \textit{p\textsubscript{inliers}}, the fraction of granted inliers to be maintained in the indexing structure, and \textit{h}, the number of histogram bins used to approximate the nearest neighbors distribution for every point; these parameters are in connection to the pruning rules employed by the method and are determined equal to 0.05 and 16, respectively.

To define $ k $ under each dataset, we have set it to 1\% of the dataset size. However, for $ R $, we followed the \textit{DolphinParamEstim} procedure stipulated in the original paper. Concerning this procedure, the parameter $ R $ directly correlates with the expected ratio of outliers, \textit{alpha}, which is anticipated to be detected by DOLPHIN.

It should be noted that DOLPHIN is the only method in our evaluations that does not provide any anomaly scores for the data elements, and its output is all and solely the definite list of potential outliers. In such a case, the ROC and PR curves will not be appealingly smooth, and the following AUC values will not be very reliable either\footnote{In fact, in more convenient computational conditions, DOLPHIN mostly leads to non-promising detection results. On the other hand, in more compelling parameter settings, i.e., lower values for $ R $ and greater values for $ k $, in spite of higher data-processing costs, the DOLPHIN algorithm outputs incline to be quite deterministic and auspicious in all cases; thus, the subsequent detection accuracy outcomes will be more dependable.}. For this reason, we decided to run the method by various $ R $ values, which are in accordance with different \textit{alpha} values\footnote{In all assessments, \textit{alpha} takes ten different values; 1 to 10 with the step length of 1 for the effectiveness experimentation, and 5 to 50 with the step length of 5 in the case of the efficiency test where the outliers ratio in the input data is much higher than usual.}, and rank the detected outliers in the entire iterations w.r.t. the sum of their appearance times in diverse iterations. This heuristic strategy would lead to some sort of outlier ranking, in which every potential outlier gains a positive integer score with a direct relationship to its anomalousness degree, while non-outlier objects attain a score of zero.

\subsubsection{\textit{S\textsubscript{p}}}\label{subSubSec_SpParamSetng}
\textit{S\textsubscript{p}} is the simplest distance-based method in our analysis and only requires one single parameter, which even with its default value proposed by the authors, promising outcomes could be achieved over multiple datasets. For this reason, we follow the same procedure as suggested in the original paper and set the sample size, $ s $, equal to 20 in our experiments.

\subsubsection{LOF and LoOP}\label{subSubSec_LOF-LoOPparamSetng}
These two traditional state-of-the-art density-based methods follow a unique structure to build the premise of their calculations. The first thing they need to compute the outlier scores is the materialization matrix, which contains the \textit{k}NN category for every data instance and the associated distances\footnote{It better be noted that while working with LOF and LoOP, for the sake of saving time and space, we decided to omit ties among the nearest neighbors; ties or the same overlapping neighbors, are those with a distance from the query point equal to the \textit{k}-th smallest distance. Only one from such category will be chosen as the \textit{k}-th nearest neighbor. Usually, this does not hurt the final detection results too much.}.

For LOF, there are two input parameters determining the lower and upper proximity bounds in \textit{k}NN evaluation. As stated by the authors, unfortunately, the LOF values attained over variant $ k $ values do not demonstrate a fixed and predictable behavior; i.e., LOF neither rises nor falls monotonically under the neighborhood parameter. Therefore, in all experiments, we follow the implied manner in the original article and set these two parameters as $ \textit{MinPtsLB} = 10 $ and $ \textit{MinPtsUB} = 50 $.

For LoOP, as it is inspired by LOF, there is solely one proximity parameter which we consider here equal to 30, as the average value of the LOF related lower and upper vicinity limits. There is another input parameter, $ \lambda $, which is in accordance with the empirical 68-95-99.7 rule and specifies the strictness level on defining anomalies; we set it equal to 3, as suggested.

\subsubsection{EnLOF}\label{subSubSec_EnLOFparamSetng}
EnLOF is an ensemble derivative of LOF, which is mainly influenced by the nearest neighbor of every object among the sampled instances. This method, like \textit{i}NNE, is essentially inspired by the \textit{i}Forest method and thus enjoys an adequate number of subsamples ($ t $) with a specific size ($ \psi $) to determine the anomaly scores for every object. Here, we follow the same premise as \textit{i}Forest and set the two parameters as suggested, i.e., $ t = 100 $ and $ \psi = 256 $.

\subsubsection{X-means}\label{subSubSec_XmeansParamSetng}
For X-means, the minimum and the maximum number of clusters are set to 1 and 15, respectively; the number of times to split a cluster and the maximum number of iterations are set to 6 and 50, respectively, as well. The other interesting parameters for building the KD-tree data structure are \textit{max\_leaf\_size} and \textit{min\_box\_width} that are set as suggested by the authors; for the datasets with a cardinality less than 100 thousand points, values equal to 40 and 0.03 are employed, respectively, and for those with greater size, in order, 80 and 0.1 are utilized.

\subsubsection{SDCOR}\label{subSubSec_SDCORparamSetng}
SDCOR enjoys a small number of input parameters that could be either set as recommended or meaningfully acquired through simple operations. For the DBSCAN parameters, in Appendix \ref{appSec_DBSCANoptmlParamSampl}, it is explained well in detail how to practically obtain the optimum values for the \textit{Eps} and \textit{MinPts} parameters through two different approaches; one, heuristic, and the other, evolutionary.

More importantly, as long as we are trying not to let outliers take part in the process of forming and updating miniclusters during scalable clustering, two of the input parameters of the proposed algorithm are more critical, listed in the following. The random sampling rate, $ \eta $, which influences the parameter $ \vec{\delta} $, the boundary on the volume of miniclusters at the time of creation; the membership threshold, $ \alpha $, which is useful to restrain, again, the volume of miniclusters, but while they are progressively growing over time.

Note that the sampling rate should not be set too low, as by which the singularity problem might happen during the ``Sampling'' phase, or even some original clusters may not take the initial density-based form, for the lack of enough data points. Hence, one can state that there is a linear relationship between the random sampling rate and the actual number of original clusters in data. In other words, for datasets with a high frequency and variety of clusters, we are forced to take higher ratios of random sampling to avoid both problems of singularity and misclustering in the ``Sampling'' stage. Here, regarding our preknowledge about the real and synthetic data, we have set $ \eta $ to 0.5\%, unless specified otherwise; e.g., the \textit{Adult} dataset was an exception as its inherent manifold forced us to exercise a larger ratio for random sampling equal to 3\%, because, for the lower rates, the singularity issue was bothering us while processing various individual runs\footnote{The random sampling rate could be much greater and equal to, e.g., the chunk size. Nonetheless, in the present report, to substantiate the excellent and stable performance of our proposed method, we decided to set it to minimal values.}.

The same problem concerns $ \alpha $, as a too low value could bring to the problem that the number of subclusters created over scalable clustering will become too large, leading to a much higher computational load. In addition, a too high value for $ \alpha $ brings the risk of outliers getting joined to normal subclusters, and it gets worse when this misclassification escalates over time, subsequently increasing the ``False Negative'' rate. Herein, for all experiments, $ \alpha $ is set to 2\footnote{However, with the expense of higher arithmetical stress, one can be more conservative and set $ \alpha $ to a lower value, e.g., 1, which is reasonable too.}. Moreover, the pruning threshold, $ \beta $, could be determined the same as $ \alpha $, and as observed over our analysis, modifying it in the range of meaningful values does not significantly affect the final output quality.

Finally, for $ \Lambda $, the PC share of total energy (variance), we decided to exert the full capacity in the case of real datasets, as none of them are, in essence, high-dimensional\footnote{Besides, the distribution of variance among the real attributes may not be fair enough, which could cause some unexpected results if fewer PCs are taken into account; furthermore, it is not easy to notice as it requires intensive computing, especially in the case of large-scale data. Therefore, it would be more prudent to avoid opting for lower shares of the total variance while working with not very high dimensions.}. However, for the synthetic datasets, in the case of more than 30 dimensions, lower quantities like 90\% or 80\% were employed.

\section{Experimental results}\label{sec_ExprResl}

\subsection{Accuracy and stability analysis}\label{subSec_AccrStabAnal}
Here, the evaluation results of the experiments conducted on various real-life and synthetic datasets, with a diversity of size and dimensionality, are presented in order to demonstrate the accuracy and stability of the proposed approach.

For SDCOR, every dataset is divided into ten chunks to simulate the algorithm capability for scalable processing. Furthermore, as we know the true structural characteristics of all the real and synthetic data, we compute the accurate anomaly score, based on the Mahalanobis distance criterion, for each object and report the following optimal AUC next to the attained result by SDCOR. Moreover, for the algorithms that have random elements, including \textit{S\textsubscript{p}}, EnLOF, and the proposed method, their detection results are reported in the format of $ \mu\pm\sigma $, over 40 independent runs, as $ \mu $ and $ \sigma $ stand for the average and the standard deviation of the correspondent AUC values.

The AUROC and AUPRC results, and the subsequent runtimes of different methods are summarized in Table~\ref{table-ROCresRealSynthDS}, Table~\ref{table-PRresRealSynthDS} and Table~\ref{table-ExecTimeResRealSynthDS}, respectively. Every \textbf{bold-faced} AUC and runtime denote the best method for a specific dataset.

\begin{table*}[!ht]
	\caption{AUROC results for SDCOR and its competitors on the real and synthetic datasets}
	\label{table-ROCresRealSynthDS}
	\centering
	\arrayrulecolor{black}
	\setlength{\tabcolsep}{4pt}
	\begin{tabular}{rlcccccccc}
		\toprule
		& Dataset & ORCA  & DOLPHIN & \textit{S\textsubscript{p}}    & LOF   & LoOP  & EnLOF & X-means & SDCOR (Optimal) \\
		\midrule
		\multicolumn{1}{l}{\multirow{9}[2]{*}{\parbox[l]{1cm}{Real Datasets}}} & Mammography & 0.703 & 0.723 & 0.818±0.040 & 0.772 & 0.741 & 0.775±0.018 & 0.803 & \textbf{0.820}±0.037 (0.879) \\
		& Adult & 0.625 & 0.622 & \textbf{0.644}±0.029 & 0.557 & 0.566 & 0.509±0.016 & 0.642 & 0.590±0.036 (0.665) \\
		& Shuttle & 0.606 & 0.721 & 0.877±0.095 & 0.614 & 0.568 & 0.703±0.053 & 0.780 & \textbf{0.969}±0.009 (0.994) \\
		& Smtp  & 0.860 & 0.847 & 0.864±0.025 & 0.881 & \textbf{0.939} & 0.835±0.031 & 0.884 & 0.779±0.008 (0.815) \\
		& Skin  & 0.692 & 0.659 & 0.881±0.065 & 0.798 & 0.060 & 0.136±0.083 & 0.862 & \textbf{0.882}±0.009 (0.889) \\
		& CreditCardFraud & 0.855 & 0.809 & 0.822±0.025 & 0.731 & 0.695 & 0.940±0.006 & 0.824 & \textbf{0.960}±0.001 (0.958) \\
		& ForestCover & 0.743 & 0.471 & 0.568±0.111 & 0.598 & 0.557 & 0.912±0.017 & 0.779 & \textbf{0.935}±0.006 (0.950) \\
		& Http  & 0.459 & 0.989 & 0.924±0.255 & 0.066 & 0.259 & 0.771±0.028 & 0.055 & \textbf{0.999}±0.001 (0.999) \\
		& Hepc  & 0.975 & 0.961 & 0.964±0.011 & 0.830 & 0.816 & 0.805±0.033 & 0.971 & \textbf{0.989}±0.003 (0.997) \\
		\midrule
		& Real AVG & 0.724 & 0.756 & 0.818±0.073 & 0.650 & 0.578 & 0.710±0.032 & 0.733 & \textbf{0.880}±0.012 (0.905) \\
		\midrule
		\multicolumn{1}{l}{\multirow{4}[2]{*}{\parbox[l]{1cm}{Synth. Datasets}}} & Data1 & 1.000 & 1.000 & 1.000±0.000 & 0.992 & 0.984 & 1.000±0.000 & 1.000 & 1.000±0.000 (1.000) \\
		& Data2 & 1.000 & 1.000 & 1.000±0.000 & 0.997 & 0.993 & 1.000±0.000 & 1.000 & 1.000±0.000 (1.000) \\
		& Data3 & 1.000 & 0.996 & 1.000±0.000 & 1.000 & 0.999 & 1.000±0.000 & 1.000 & 1.000±0.000 (1.000) \\
		& Data4 & 1.000 & 1.000 & 1.000±0.000 & 1.000 & 1.000 & 1.000±0.000 & 1.000 & 1.000±0.000 (1.000) \\
		\midrule
		& Synth. AVG & 1.000 & 0.999 & 1.000±0.000 & 0.997 & 0.994 & 1.000±0.000 & 1.000 & 1.000±0.000 (1.000) \\
		\bottomrule
	\end{tabular}%
\end{table*}%

\begin{table*}[!ht]
	\caption{AUPRC results for SDCOR and its competitors on the real and synthetic datasets}
	\label{table-PRresRealSynthDS}
	\centering
	\arrayrulecolor{black}
	\setlength{\tabcolsep}{4pt}
	\begin{tabular}{rlcccccccc}
		\toprule
		& Dataset & ORCA  & DOLPHIN & \textit{S\textsubscript{p}}    & LOF   & LoOP  & EnLOF & X-means & SDCOR (Optimal) \\
		\midrule
		\multicolumn{1}{l}{\multirow{9}[2]{*}{\parbox[l]{1cm}{Real Datasets}}} & Mammography & 0.149 & \textbf{0.151} & 0.148±0.056 & 0.136 & 0.093 & 0.095±0.013 & 0.123 & 0.148±0.067 (0.277) \\
		& Adult & 0.099 & 0.083 & \textbf{0.110}±0.018 & 0.038 & 0.037 & 0.033±0.002 & 0.109 & 0.073±0.020 (0.230) \\
		& Shuttle & 0.178 & 0.314 & 0.608±0.182 & 0.132 & 0.115 & 0.251±0.035 & 0.328 & \textbf{0.617}±0.049 (0.942) \\
		& Smtp  & 0.353 & 0.001 & \textbf{0.496}±0.076 & 0.054 & 0.013 & 0.001±0.001 & 0.293 & 0.245±0.083 (0.252) \\
		& Skin  & 0.073 & 0.063 & \textbf{0.115}±0.043 & 0.060 & 0.037 & 0.014±0.001 & 0.083 & 0.096±0.007 (0.100) \\
		& CreditCardFraud & 0.044 & 0.010 & 0.006±0.001 & 0.016 & 0.033 & 0.035±0.018 & 0.007 & \textbf{0.708}±0.003 (0.489) \\
		& ForestCover & 0.067 & 0.007 & 0.011±0.003 & 0.019 & 0.015 & \textbf{0.090}±0.027 & 0.020 & 0.089±0.009 (0.109) \\
		& Http  & 0.006 & 0.100 & 0.231±0.084 & 0.002 & 0.002 & 0.008±0.001 & 0.003 & \textbf{0.463}±0.063 (0.500) \\
		& Hepc  & 0.249 & 0.251 & 0.370±0.081 & 0.016 & 0.018 & 0.032±0.018 & 0.322 & \textbf{0.426}±0.070 (0.658) \\
		\midrule
		& Real AVG & 0.136 & 0.109 & 0.233±0.060 & 0.053 & 0.040 & 0.062±0.013 & 0.143 & \textbf{0.318}±0.041 (0.395) \\
		\midrule
		\multicolumn{1}{l}{\multirow{4}[2]{*}{\parbox[l]{1cm}{Synth. Datasets}}} & Data1 & 1.000 & 0.426 & 1.000±0.000 & 0.939 & 0.874 & 1.000±0.000 & 1.000 & 1.000±0.000 (1.000) \\
		& Data2 & 1.000 & 0.000 & 1.000±0.000 & 0.966 & 0.888 & 1.000±0.000 & 1.000 & 1.000±0.000 (1.000) \\
		& Data3 & 1.000 & 0.390 & 1.000±0.000 & 0.992 & 0.871 & 1.000±0.000 & 1.000 & 1.000±0.000 (1.000) \\
		& Data4 & 1.000 & 0.000 & 1.000±0.000 & 1.000 & 0.999 & 1.000±0.000 & 1.000 & 1.000±0.000 (1.000) \\
		\midrule
		& Synth. AVG & 1.000 & 0.204 & 1.000±0.000 & 0.974 & 0.908 & 1.000±0.000 & 1.000 & 1.000±0.000 (1.000) \\
		\bottomrule
	\end{tabular}%
\end{table*}%

\begin{table*}[!ht]
	\caption{Execution time (secs) results for SDCOR and its competitors on the real and synthetic datasets}
	\label{table-ExecTimeResRealSynthDS}
	\centering
	\arrayrulecolor{black}
	\setlength{\tabcolsep}{4.1pt}
	\begin{threeparttable}
		\begin{tabular}{rlllllllll}
			\toprule
			& Dataset & ORCA  & DOLPHIN\tnote{1} & \textit{S\textsubscript{p}}    & LOF   & LoOP  & EnLOF & X-means & SDCOR\tnote{2} \\
			\midrule
			\multicolumn{1}{l}{\multirow{9}[2]{*}{\parbox[l]{1cm}{Real Datasets}}} & Mammography & 5.909 & 2.750 & \textbf{0.020} & 0.972 & 0.423 & 1.956 & 1.141 & 0.701 \\
			& Adult & 33.860 & 90.010 & \textbf{0.050} & 6.087 & 4.507 & 11.007 & 6.926 & 0.855 \\
			& Shuttle & 80.367 & 99.670 & \textbf{0.061} & 10.058 & 7.836 & 15.277 & 7.385 & 0.907 \\
			& Smtp  & 142.526 & 162.410 & \textbf{0.110} & 6.225 & 1.927 & 8.448 & 38.663 & 0.895 \\
			& Skin  & 405.904 & 673.170 & \textbf{0.209} & 14.625 & 5.165 & 12.915 & 8.564 & 1.307 \\
			& CreditCardFraud & 5,396.614 & 1,281.840 & \textbf{0.528} & 539.153 & 524.688 & 169.969 & 374.642 & 7.811 \\
			& ForestCover & 3,618.686 & 3,557.930 & \textbf{0.489} & 22.865 & 9.577 & 40.152 & 22.246 & 1.643 \\
			& Http  & 5,570.813 & 5,879.130 & \textbf{0.686} & 55.660 & 25.305 & 56.319 & 793.285 & 1.848 \\
			& Hepc  & 92,882.000 & 243,583.580 & \textbf{3.247} & 45,098.631 & 37,145.707 & 643.060 & 7,078.436 & 17.632 \\
			\midrule
			& Real AVG & 12,015.187 & 28,370.054 & \textbf{0.600} & 5,083.808 & 72.428 & 106.567 & 925.698 & 3.733 \\
			\midrule
			\multicolumn{1}{l}{\multirow{4}[2]{*}{\parbox[l]{1cm}{Synth. Datasets}}} & Data1 & 18,838.446 & 55,411.820 & \textbf{0.918} & 5,924.641 & 5,458.474 & 324.351 & 461.955 & 12.863 \\
			& Data2 & 101,361.263 & 262,633.230 & \textbf{2.075} & 33,806.998 & 32,672.413 & 856.039 & 1,641.932 & 41.162 \\
			& Data3 & 224,559.345 & 647,534.410 & \textbf{3.479} & 83,647.072 & 82,139.098 & 1,520.316 & 3,528.569 & 89.455 \\
			& Data4 & 389,826.856 & 1,210,724.637 & \textbf{4.997} & 133,835.315 & 132,388.373 & 2,267.934 & 5,952.274 & 161.932 \\
			\midrule
			& Synth. AVG & 183,646.478 & 544,076.024 & \textbf{2.867} & 64,303.506 & 63,164.590 & 1,242.160 & 2,896.183 & 76.353 \\
			\bottomrule
		\end{tabular}%
		\begin{tablenotes}
			\item[1]\footnotesize{The reported quantity is the highest execution time over various DOLPHIN runs w.r.t. different $ R $ values.}
			\item[2]\footnotesize{The required time for finding the optimal parameters of DBSCAN is not taken into account as a part of the total runtime.}
		\end{tablenotes}
	\end{threeparttable}
\end{table*}%

\subsubsection{Real datasets}\label{subSubSec_TestRealDats}
As for the real datasets, Table~\ref{table-ROCresRealSynthDS} and Table~\ref{table-PRresRealSynthDS} show that in terms of both AUROC and AUPRC, SDCOR is more accurate than all the other competing methods. Further elaborated, in terms of AUROC, SDCOR outperforms ORCA, DOLPHIN, \textit{S\textsubscript{p}}, and X-means in seven out of the nine real datasets, and compares favorably with them in the other two viz \textit{Adult} and \textit{Smtp}; for LOF, LoOP, and EnLOF, SDCOR excels them in eight out of the nine real-life data, and only in the case of the \textit{Smtp} dataset, it falls behind these three density-based methods, though not with a remarkable difference. In terms of AUPRC, SDCOR surpasses all the other techniques in all datasets, except for ORCA in \textit{Mammography}, \textit{Adult}, and \textit{Smtp}; for DOLPHIN in \textit{Mammography} and \textit{Adult}; for \textit{S\textsubscript{p}} in \textit{Adult}, \textit{Smtp}, and \textit{Skin}; for EnLOF in only \textit{ForestCover}; for X-means in \textit{Adult} and \textit{Smtp}. However, the superior methods over SDCOR concerning AUPRC have either a quite equivalence or an insignificant difference with it in all the superiority cases.

Moreover, it is obvious that in the context of AUROC, the attained results by the proposed approach are almost the same as the optimal ones. However, in the case of AUPRC, the deviation from the optimum case is not trivial in most cases\footnote{Despite this inescapable deviation, it is crystal clear that even the optimum values for AUPRC are in general far away from the perfect state, and this is quite natural while analyzing anomaly identification approaches. Further detailed, anomalies are typically in the minority, thus making the correspondent dataset highly-skewed in the class distribution; i.e., the number of negative elements (inliers) substantially outmatches the number of positive elements (outliers). On the other hand, w.r.t. the ROC and PR formulations, the impact of the number of false positives is inferior in the ROC analysis, though gains great superiority while estimating PR; such distinction mainly originates in the incorporated Precision criterion by PR. As a result, the referred data imbalance could incur, as a rule, considerably lower values for AUPRC than the related AUROC \cite{davis2006relationship}.}, but it is satisfactory as SDCOR generally stays at the top w.r.t. the other contending methods. Additionally, the \textit{CreditCardFraud} dataset is an exception concerning the optimal AUC levels, in which both the AUROC and AUPRC outcomes outstrip the corresponding ideal conditions; for AUROC, the distinction is superficial, though, for AUPRC, it is considerable. Such an unusual condition could be caused by some adopted suboptimal clustering parameters; moreover, sometimes, it could come off because of the closeness of some outliers to the original normal clusters in certain real situations, along with the slight variations in the acquired ultimate Mahalanobis contour lines on account of the minor randomness in the functionality of the proposed method.

In addition, the average lines indicate that SDCOR performs overall much better than all the other methods. More importantly, SDCOR is effective on the largest datasets, \textit{Http} and \textit{Hepc}, with rather perfect results. In addition, by considering the standard deviations of AUC values for \textit{S\textsubscript{p}}, EnLOF, and SDCOR, it is evident that SDCOR is much more stable than the other two random-based outlier identification methods\footnote{In the context of AUROC which is more prevalent than AUPRC for analyzing anomaly detection approaches, the \textit{Mammography} and \textit{Adult} datasets come down with further variances, and this is as a consequence of the adopted low sampling rates for them. These two data series have the smallest cardinality among all, and thus the utilized sampling ratios are not effective in the best way. Furthermore, as stipulated earlier, we do this to manifest the capability of the proposed method to operate fairly well even with shallow levels of random sampling; however, for higher sampling ratios, more consistent results were observed.}, except for EnLOF in terms of AUPRC, which suffers somewhat from a lower average variance than SDCOR. For \textit{S\textsubscript{p}}, it should be pointed out that the outstanding values of standard deviation for it are due to using only one tiny sample in every execution, which causes the algorithm to go through large variations on the final accuracy.

Besides, Table~\ref{table-ExecTimeResRealSynthDS} reveals that SDCOR performs much better than the other competing methods in terms of execution time, except for \textit{S\textsubscript{p}} that accomplishes the tasks slightly faster than the proposed algorithm. Although \textit{S\textsubscript{p}} is the fastest among the compared algorithms, as it was noted, its AUC results suffer from large variations, and overall is lower than SDCOR. Moreover, w.r.t. the two conventional and one novel density-based methods, it is evident that there is a huge difference in the consuming time between SDCOR and these counterparts from the density realm; it is due to the fact that differently from LOF and LoOP, in SDCOR, it is not required to compute the pairwise distances of the total objects in a dataset, and in contrast to EnLOF, SDCOR is not established on some ensemble calculations which despite being efficient in certain aspects, could incur high computational expenses.

Finally, it is worthwhile to mention some valuable comments here. As stated earlier at the beginning of this paper, the strong assumption of SDCOR is on the structure of existing clusters in the input dataset, which should have Gaussian distribution. In practice, though, w.r.t. \cite{shlens2014tutorial}, with appreciation to the Central Limit Theorem, a fair amount of the real-world data follows the Gaussian distribution. Furthermore, w.r.t. \cite{jolliffe2011principal,hawkins1980identification,barnett1974outliers}, even when the original variables are not Normal (Gaussian), employing the properties of PCs for detecting outliers is possible, and the corresponding results will be reliable. Since, given that the PCs are linear functions of $ p $ random variables, a call for the Central Limit Theorem may vindicate the approximate Normality for the PCs, even in the cases that the original variables are not Normal.

Regarding this concern, it is plausible to set up more official statistical tests for outliers based on the PCs, supposing that the PCs have Gaussian distributions. Moreover, the exercise of the Mahalanobis distance criterion for outlier detection is only viable for the convex-shaped clusters. Otherwise, outliers could be assigned to an irregular (density-based) cluster under the masking effect, thus will be misclassified.

\subsubsection{Synthetic datasets}\label{subSubSec_TestSynthDats}
The experiments are carried out on four artificial datasets with specific details presented in the lower part of Table~\ref{table-DataProp}; each dataset consists of 6 Gaussian clusters, and the outliers take up 1 percent of its volume.

The excellent results in Table~\ref{table-ROCresRealSynthDS} and Table~\ref{table-PRresRealSynthDS} verify that the synthetic datasets are overall too simple for SDCOR and also for all the other competing methods except for DOLPHIN in the case of AUPRC. The chief reason for this almost-common ideal performance is that, as stipulated beforehand, in our dedicated setting for building these synthetic data collections, normal objects are in very dense areas, and outliers reside in very sparse zones far enough away from the normal clusters.

However, despite the absolute functioning of most of the other rivaling techniques, regarding the time consumption results in Table~\ref{table-ExecTimeResRealSynthDS}, apart from \textit{S\textsubscript{p}}, the other runtimes are remarkably and incompetently higher than SDCOR. Besides, for \textit{S\textsubscript{p}}, which is performing slightly quicker than our proposed method, first, considering all runtime results, its distinction from SDCOR is not significant, and second, its accuracy in the real situations, which is the most challenging contest, is still quite far away from SDCOR.

\subsection{Significance Analysis}\label{subSec_FriedTest}
As it was clarified beforehand, through our assessment, some of the compared methods are non-deterministic, namely \textit{S\textsubscript{p}}, EnLOF, and SDCOR; i.e., they present different results over diverse executions, even if the same input parameters are exerted globally. Therefore, in this subsection, we perform a significance analysis to realize whether there are major differences in the erratic results out of these three random-based methods. Here, we adopt the Friedman test \cite{friedman1937use,demsar06} for such statistical analysis.

The Friedman test is a distribution-free statistical hypothesis examination that arranges the significance of the compared algorithms w.r.t. every dataset independently; as it is free of any input parameters, it is also called a non-parametric randomized black analysis of variance. The null hypothesis of this test is that all of the populations of the experimental results \textemdash~each population corresponding to the application of one of the analyzed methods in different repetitions \textemdash~have the same median value. Once the Friedman test rejects the null hypothesis, a post-hoc test is required to make pairwise comparisons and detect the couples of significantly distinctive methods.

A critical value obtained from a chi-squared distribution with two degrees of freedom and a significance level of 5\% is used for this test. Furthermore, the \textit{p}-value was adjusted by using the Holm methodology \cite{Garcia09}, and also, as for the post-hoc test, to verify the differences between each couple of methods, the Nemenyi test \cite{demsar06} was adopted.

\begin{table*}[!ht]
	\caption{Friedman test results for the real datasets, on the AUROC and AUPRC outputs out of the \textit{S\textsubscript{p}}, EnLOF, and SDCOR methods}
	\label{table-FriedTest}
	\centering
	\arrayrulecolor{black}
	\setlength{\tabcolsep}{4.5pt}
	\begin{tabular}{lcccccccc}
		\toprule
		&       & \multicolumn{3}{c}{AUROC} &       & \multicolumn{3}{c}{AUPRC} \\
		\midrule
		Dataset &   ~~~~~~    & \textit{S\textsubscript{p}}    & EnLOF & SDCOR &    ~~~~~~   & \textit{S\textsubscript{p}}    & EnLOF & SDCOR \\
		\midrule
		Mammography &       & \textbf{0.818} & 0.775 & \textbf{0.820} &       & \textbf{0.148} & 0.095 & \textbf{0.148} \\
		Adult &       & \textbf{0.644} & 0.509 & 0.590 &       & \textbf{0.110} & 0.033 & \textbf{0.073} \\
		Shuttle &       & 0.877 & 0.703 & \textbf{0.969} &       & \textbf{0.608} & 0.251 & \textbf{0.617} \\
		Smtp  &       & \textbf{0.864} & \textbf{0.835} & 0.779 &       & \textbf{0.496} & 0.001 & 0.245 \\
		Skin  &       & \textbf{0.881} & 0.136 & \textbf{0.882} &       & \textbf{0.115} & 0.014 & \textbf{0.096} \\
		CreditCardFraud &       & 0.822 & 0.940 & \textbf{0.960} &       & 0.006 & 0.035 & \textbf{0.708} \\
		ForestCover &       & 0.568 & \textbf{0.912} & \textbf{0.935} &       & 0.011 & \textbf{0.090} & \textbf{0.089} \\
		Http  &       & 0.924 & 0.771 & \textbf{0.999} &       & 0.231 & 0.008 & \textbf{0.463} \\
		Hepc  &       & 0.964 & 0.805 & \textbf{0.989} &       & \textbf{0.370} & 0.032 & \textbf{0.426} \\
		\bottomrule
	\end{tabular}%
	\label{tab:addlabel}%
\end{table*}%

The Friedman test results for the real-life data series, over the AUROC and AUPRC measurements, are reported in Table~\ref{table-FriedTest}\footnote{It should be noted that for this statistical appraisal, the same results out of the 40 iterative runs mentioned in Subsection~\ref{subSec_AccrStabAnal} are exerted. Furthermore, the following stipulated average AUC ranks in this subsection are utilized in Table~\ref{table-FriedTest} as representatives for the non-deterministic methods on different datasets.}. In particular, in every row, if only one method is reported in \textbf{bold}, it is determined as the significantly best method among all w.r.t. the corresponding dataset; if two values are reported in \textbf{bold}, it means that there is no significant difference between the related methods, though both of them are performing better than the remaining one. Moreover, for the four synthetic datasets in the efficacy experimentation, all of the acquired results out of the three random-based methods are perfect; thus, there is no place for more evaluation.

It is evident that in terms of AUROC, SDCOR is the only winner in 4, the co-winner along with \textit{S\textsubscript{p}} and EnLOF among respectively 2 and 1, out of the nine real-life datasets; hence, SDCOR is the only winner or is amid the winners in 7 out of the total. However, \textit{S\textsubscript{p}} is the sole winner only in the case of the \textit{Adult} dataset, while EnLOF never stands in the first place without any rivals. In terms of AUPRC, SDCOR conquers others only in two datasets, viz \textit{CreditCardFraud} and \textit{Http}, and in the other 6 instances, it is among the conquerors mostly along with \textit{S\textsubscript{p}} and exclusively with EnLOF just in the case of the \textit{ForestCover} dataset; \textit{S\textsubscript{p}} excels others only in \textit{Smtp}, whereas EnLOF never gains the absolute superiority. Overall, it is apparent that SDCOR is the conqueror and the most precise technique among all non-deterministic methods in our analysis.

\subsection{SDCOR vs. X-means}\label{subSec_SDCORvsXmeans}
The evaluation results over the real-world data collections for the two clustering-based techniques, SDCOR and X-means, obtained out of the five external clustering validity measures, namely Purity, the Mirkin metric, F-measure, Entropy, and VI, along with the following averages are presented in Table~\ref{table-ClustValidMeas}. Moreover, for every metric, concerning its ideal condition mentioned beneath the metric name, the best method(s) among two for a particular dataset is/are indicated in \textbf{bold-faced} print. Besides, for the four synthetic datasets in the effectiveness evaluations, as both SDCOR and X-means attain perfect results, we omit demonstrating them here.

\begin{table*}[!ht]
	\caption{Results of the clustering validity measures for SDCOR and X-means on the real datasets}
	\label{table-ClustValidMeas}
	\centering
	\arrayrulecolor{black}
	\setlength{\tabcolsep}{1.8pt}
	\begin{tabular}{lrccrccccccccccc}
		\toprule
		&   ~~~    & \multicolumn{2}{c}{Purity} &   ~~~    & \multicolumn{2}{c}{Mirkin} &   ~~~    & \multicolumn{2}{c}{F-measure} &   ~~~    & \multicolumn{2}{c}{Entropy} &   ~~~    & \multicolumn{2}{c}{VI} \\
		&       & \multicolumn{2}{c}{(max=1)} &       & \multicolumn{2}{c}{(min=0)} &       & \multicolumn{2}{c}{(max=1)} &       & \multicolumn{2}{c}{(min=0)} &       & \multicolumn{2}{c}{(min=0)} \\
		\midrule
		Dataset &       & SDCOR & X-means &       & SDCOR & X-means &       & SDCOR & X-means &       & SDCOR & X-means &       & SDCOR & X-means \\
		\midrule
		Mammography &       & 0.977 & \textbf{0.978} &       & \textbf{0.071} & 0.784 &       & \textbf{0.963} & 0.299 &       & 0.149 & \textbf{0.107} &       & \textbf{0.022} & 0.223 \\
		Adult &       & \textbf{0.970} & \textbf{0.970} &       & \textbf{0.105} & 0.836 &       & \textbf{0.945} & 0.207 &       & 0.195 & \textbf{0.169} &       & \textbf{0.026} & 0.232 \\
		Shuttle &       & 0.954 & \textbf{0.976} &       & \textbf{0.088} & 0.091 &       & \textbf{0.954} & 0.918 &       & 0.221 & \textbf{0.076} &       & 0.028 & \textbf{0.027} \\
		Smtp  &       & \textbf{1.000} & \textbf{1.000} &       & \textbf{0.001} & 0.894 &       & \textbf{1.000} & 0.190 &       & \textbf{0.002} & \textbf{0.002} &       & \textbf{0.000} & 0.207 \\
		Skin  &       & 0.974 & \textbf{0.975} &       & \textbf{0.097} & 0.831 &       & \textbf{0.950} & 0.232 &       & 0.170 & \textbf{0.073} &       & \textbf{0.019} & 0.188 \\
		CreditCardFraud &       & \textbf{0.999} & 0.998 &       & \textbf{0.001} & 0.654 &       & \textbf{0.999} & 0.473 &       & \textbf{0.006} & 0.018 &       & \textbf{0.001} & 0.116 \\
		ForestCover &       & \textbf{0.990} & \textbf{0.990} &       & \textbf{0.036} & 0.906 &       & \textbf{0.982} & 0.144 &       & 0.078 & \textbf{0.061} &       & \textbf{0.009} & 0.212 \\
		Http  &       & 0.999 & \textbf{1.000} &       & \textbf{0.002} & 0.814 &       & \textbf{0.999} & 0.305 &       & 0.009 & \textbf{0.001} &       & \textbf{0.001} & 0.133 \\
		Hepc  &       & \textbf{0.997} & \textbf{0.997} &       & \textbf{0.005} & 0.901 &       & \textbf{0.997} & 0.173 &       & 0.017 & \textbf{0.014} &       & \textbf{0.002} & 0.172 \\
		\midrule
		Real AVG &       & 0.984 & \textbf{0.987} &       & \textbf{0.045} & 0.746 &       & \textbf{0.977} & 0.327 &       & 0.094 & \textbf{0.058} &       & \textbf{0.012} & 0.168 \\
		\bottomrule
	\end{tabular}%
\end{table*}%

It is perceptible from Table~\ref{table-ClustValidMeas} that in all comparison cases, SDCOR is either the undoubted winner or compares favorably well with X-means. In the case of Purity and Entropy, SDCOR fails to conquer X-means overall, although with a negligible difference. However, for the Mirkin metric, F-measure, and VI, the superiority is achieved by SDCOR.

More detailed, it was noticed that in contrast to Purity and Entropy, which alone consider the conformity of the output, F-measure enjoys an essential advantage over them, as it takes into account both the conformity and integrity of the clustering solution. Now w.r.t. the X-means fundamentals in which there is no noise taken for granted beforehand, and more crucial, it is tried to split the input data to multiple clusters until it reaches a reasonable quantity for the frequency of the clusters, hence an arbitrary original cluster (class), as well as the outliers pack, can be eventually divided into several subclusters; this is the exact point which makes the major deviation between X-means and our proposed approach in which the number of output clusters is the same as the ground truth.

Concerning the conformity and integrity concepts, when a cluster contains objects solely from a unique class but does not wholly encompass it, we will have perfect conformity though imperfect integrity as output. A similar conception is intelligently embedded in the Mirkin metric and VI mathematical formulas. In the Mirkin metric, the term corresponding to the intersection of the related clusters and classes will be shallow in the case of high-conformity/low-integrity, thus producing Mirkin values more distant from 0, the optimal condition; this makes the Mirkin metric quite the same as F-measure in quality. On the other hand, in VI, as an enhanced version of Entropy, it is stipulated that the gained/lost portion of information while moving from one partition to another is computed. More precisely, the input to the logarithm term, which in the ideal state equals 1, will tend to greater values in non-paradigmatic conditions. This causes the following measured quantity for VI to become farther away from the optimum case equal to 0. Such quality of VI addresses both the conformity and integrity concepts.

Finally, for the real case of the evaluated data series, as in all instances, there is merely a normal cluster and some outliers around it, X-means, w.r.t. the employed parameter setting, inevitably breaks the normal cluster along with the outliers pack down to various subclusters; however, as stated earlier, as required in our analysis, identified outliers, despite being scattered over various subclusters, are after all distinguished as a single cluster, and this is carried out by sorting the related outlier scores and cutting off the topmost indices. This cluster chopping procedure done by X-means leads us to the high-conformity/low-integrity situation, and the following deficiencies in the final clustering cogency evaluations through the Mirkin metric, F-measure, and VI; for Purity and Entropy though, as the outliers are separated well from the inliers in the final clustering solution, hence the acquired measurements for these two conformity-based metrics are generally satisfactory. However, for the four artificial datasets, each of them containing 6 clusters and some distant anomalies dispersedly in the space, it is lucky that the normal clusters are not broken into smaller pieces, and it has occurred just for the outliers pack due to their much greater dispersion; therefore, regarding the flawless anomaly detection results for these synthetic datasets, the following clustering validness measures are all perfect too.

\subsection{Tolerance to a high number of outliers}\label{subSec_TolrHighNumbOutl}
To analyze the accuracy when the number of outliers is increased, we follow the same procedure used to generate the synthetic datasets, described in detail in Subsection~\ref{subSubSec_SynthDataDscr}. More precisely, every dataset is established upon a fixed manifold concerning the normal elements, comprising 4 pruned Gaussian clusters with a total of 20,000 objects, in a predefined dimensionality of 2. For the sake of adding noise, the percentage of injected outliers is increased from 50 to 150 percent of the volume of the normal objects, with a step length of 10 percent; alternatively stated, with a fixed number of 20,000 normal elements in every dataset, the number of outliers is raised from 10,000 to 30,000 with the step length of 2,000. Moreover, as we observe this presumption that these data series hold very high rates of noise, sizeable \textit{MinPts} values for finding the DBSCAN optimum parameterization, along with a random sampling rate equal to 10\% in all tests, are exerted.

\begin{figure*}[pos=t!]
	\centering
	\includegraphics[width=1\linewidth]{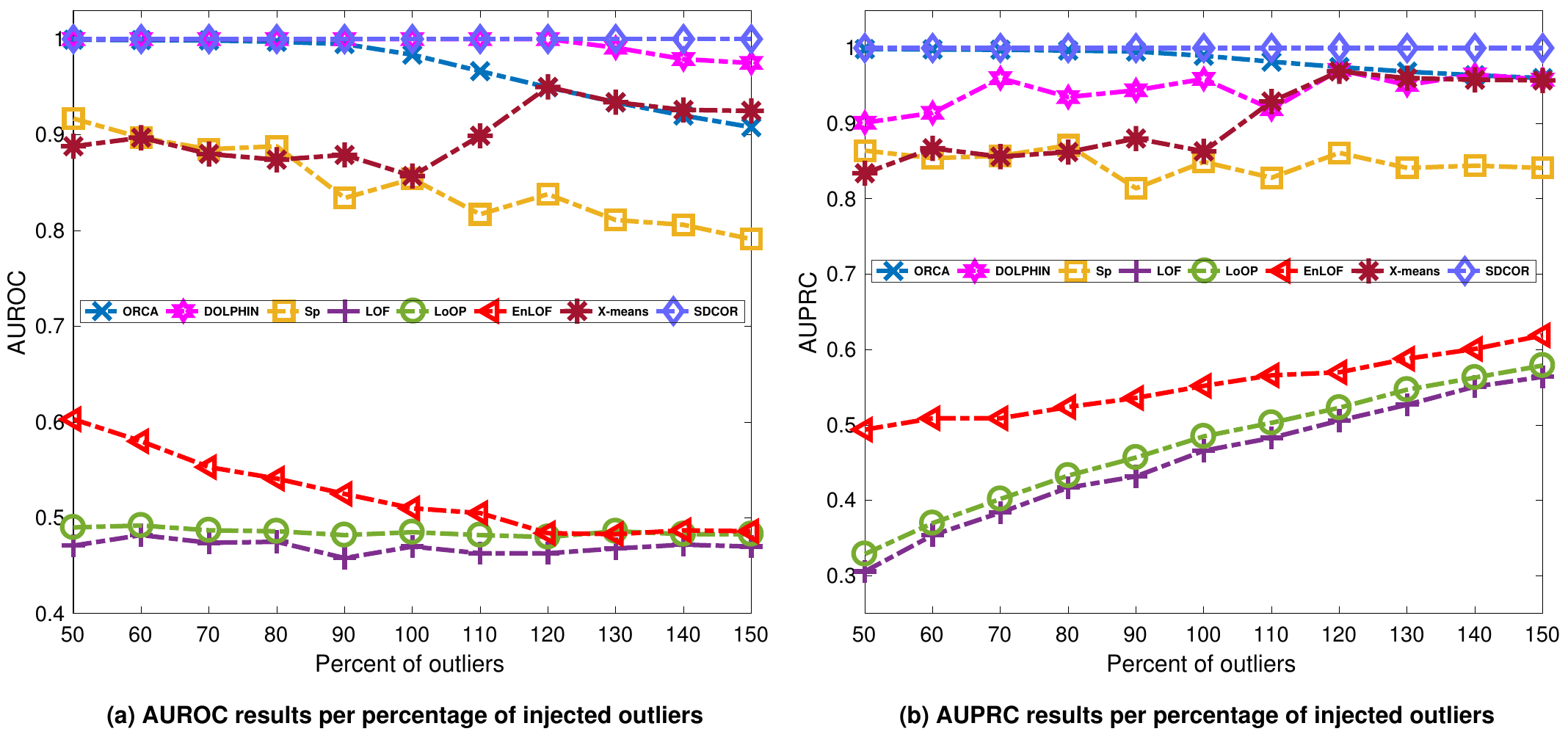}
	\caption{Robustness results out of the competing methods, over the AUROC and AUPRC measures}
	\label{figure-RobustTest}
\end{figure*}

\begin{figure}[pos=h!]
	\centering
	\includegraphics[width=1\linewidth]{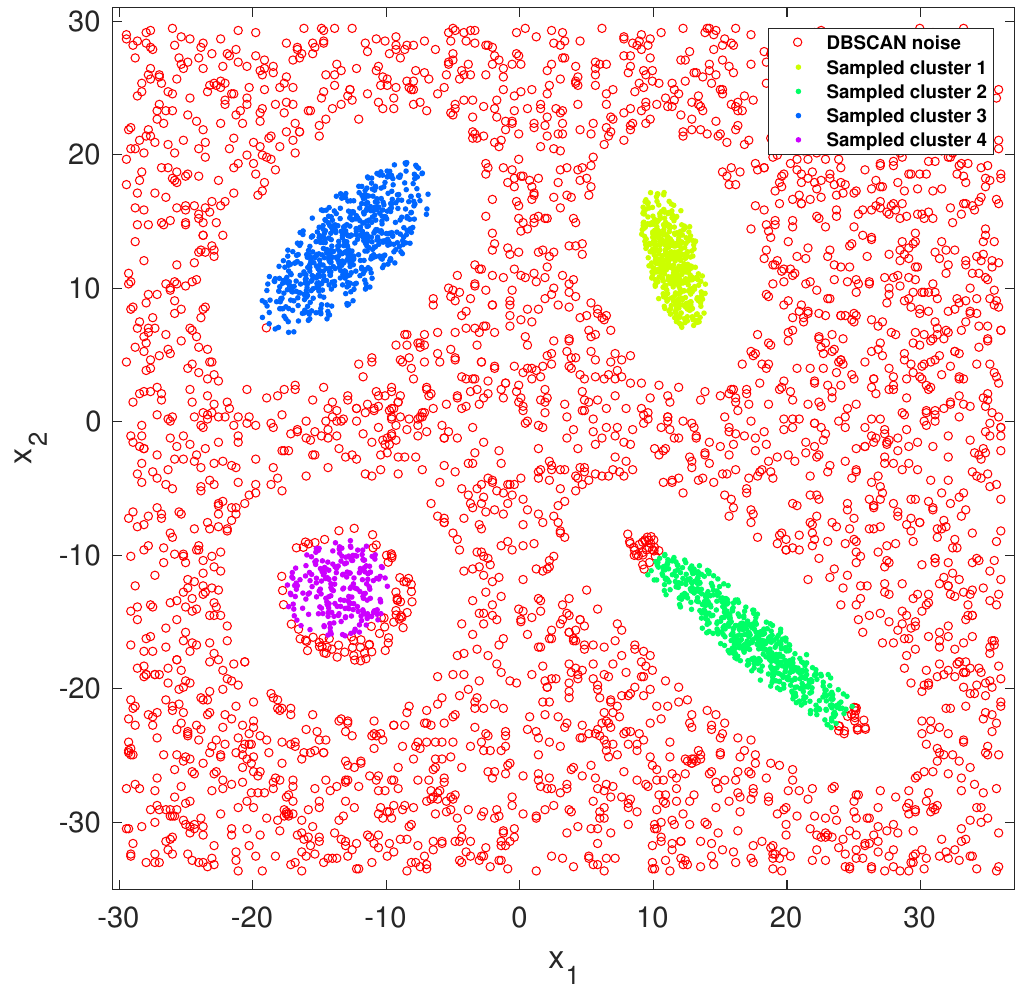}
	\caption{DBSCAN result with the optimal parameters on the high-noisy sampled data}
	\label{figure-DBSCANnois}
\end{figure}

Fig.~\ref{figure-RobustTest} reveals that, in terms of both AUROC and AUPRC, SDCOR is noise-tolerant, as by increasing the percentage of outliers, always perfect AUC results are achieved out of our method. Concerning the other methods, their behavior under both AUROC and AUPRC is somewhat the same. ORCA and DOLPHIN are the closest rivals to the proposed method; \textit{S\textsubscript{p}} and X-means generally perform at a lower level but are relatively good. However, for LOF, LoOP, and EnLOF, they are entirely misclassifying outliers in all complex conditions\footnote{Actually, except for EnLOF which is predefined to work with only the nearest neighbor of every object, for the other two state-of-the-art density-based techniques, such deficient outcomes are quite related to the lack of the optimal parameters. More detailed, it was noted beforehand that in LOF and LoOP, locating the optimum values for the proximity options is not easy and demands a high amount of trial and error w.r.t. the existing outliers ground truth, which is not fair concerning our analytical context.}.

The reason for the favorable functionality of SDCOR in bearing much noise is that the basis for forming miniclusters during scalable clustering is the fulfillment of the DBSCAN requirements, which are obtained out of the ``Sampling'' stage. Simply put, as all of the normal objects follow a Gaussian distribution and outliers are injected using a continuous uniform distribution, hence the local density of the normal points is much higher than that of the outliers; therefore, the acquired optimal parameters for the density-based clustering are obtained proportional to the dense regions containing only the sampled inliers. For this reason, in every memory process, the probability of a subcluster being formed by the outliers is much lesser than that of the normal objects, and thus, our approach based on DBSCAN obtains very satisfactory results.

Fig.~\ref{figure-DBSCANnois} shows the DBSCAN result on the sampled data from the test dataset with 150 percent of injected outliers. Four discovered Gaussian clusters and noises are represented with dots in different colors and empty red circles, respectively. As it is evident, even in this highly noisy situation, outliers cannot satisfy DBSCAN constraints on forming a minicluster.

Here, it is worth remarking that w.r.t. our non-reported evaluations on other efficiency tests, when the numbers of normal objects and attributes are increasing (like what we did in the ``percentage of outliers'' test), under the strong assumptions of the proposed method, one can still anticipate significant detection results out of SDCOR. However, for some of the methods in the contest, w.r.t. their corresponding high computational load, it would be a tedious task to obtain the accuracy results out of them for the very large datasets employed in our experiments.

\subsection{Scalability}\label{subSec_ScalTest}
To assess the scalability of the proposed method, we measure the time consumption with the increasing number of objects. For this purpose, first, a synthetic dataset containing 4 Gaussian clusters with 200,000 elements in 10 dimensions is generated. Then, we conduct random sampling with the rates of 10 to 100 percent with a step length of 10 percent to build ten datasets for the scalability test; for each resulting dataset, we inject 200 outliers into it. In other words, we want to analyze the execution time using datasets having very similar essential characteristics, namely the location (centroid) and the shape (covariance matrix). Moreover, in all experiments, parameters were set as suggested, and the following detection results were all perfect.

\begin{figure}[pos=b!]
	\centering
	\includegraphics[width=1\linewidth]{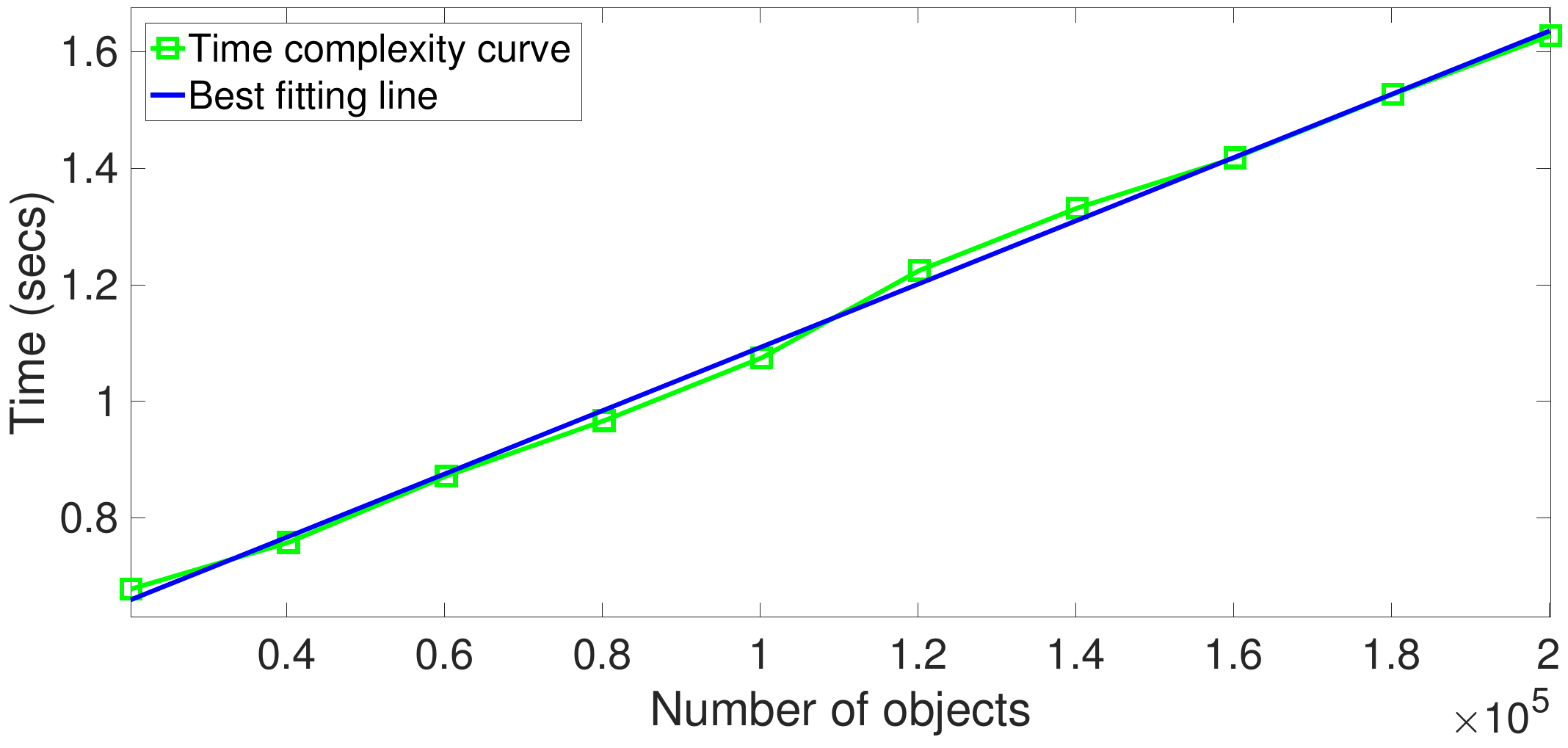}
	\caption{Scalability test result for SDCOR}
	\label{figure-ResTimComplxTest}
\end{figure}

Fig.~\ref{figure-ResTimComplxTest} demonstrates that the run time of SDCOR is close to a linear function of the number of objects; this is a confirmation of the result of the analysis conducted in Subsection~\ref{subSec_AlgoCmplx} on the algorithm complexity. In particular, for the dataset with the minimum size equal to 20,200, the run time is 0.678 seconds. However, when the number of objects reaches the maximum value equal to 200,200 \textemdash~i.e., about ten times the minimum value \textemdash~the processing time increases by only 2.4 times to 1.628 seconds. Accordingly, SDCOR has a low constant in its runtime complexity, and hence, is scalable.

\subsection{Effect of sampling rate}\label{subSec_SampRatTest}
Here, we examine the variation of the covariance determinant belonging to the sampled data from a unique cluster per various random sampling rates. Therefore, first of all, we create an arbitrary Gaussian cluster with 10,000 objects and two attributes. Then, we start sampling with the rate of 0.5 percent and proceed to 100 percent with the step length of 0.5 percent; subsequently, for each of these resulting sampled clusters, we calculate the corresponding covariance determinant to be incorporated in our analysis.

\begin{figure*}[pos=t!]
	\centering
	\includegraphics[width=1\linewidth]{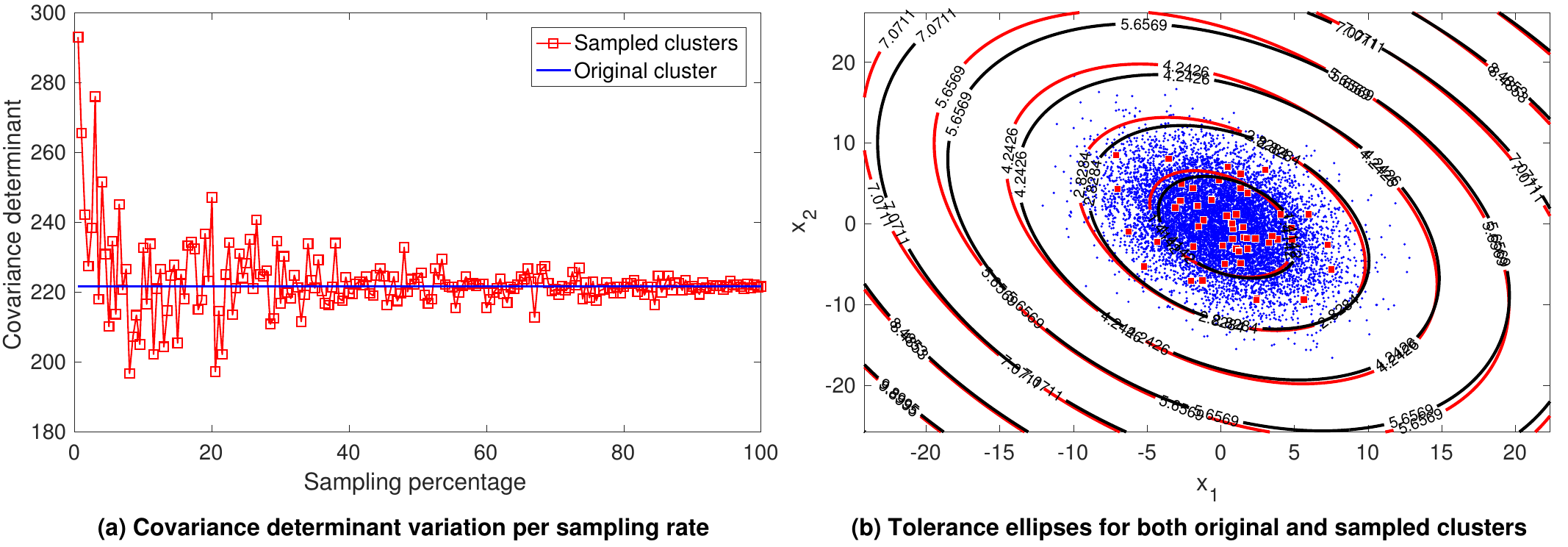}
	\caption{Covariance determinant and tolerance ellipses variation per random sampling rate}
	\label{figure-CovDetAndTolEllpVarRandSampRat}
\end{figure*}

As Fig.~\ref{figure-CovDetAndTolEllpVarRandSampRat}a displays, with the increase in the random sampling rate, the corresponding covariance determinant is approaching that of the original cluster. However, as it is evident, even covariance determinants associated with very shallow sampling rates are quite close to that of the main cluster. For example, for the lowest sampling rate equal to 0.5 percent, the corresponding covariance determinant is approximately 300, which is close enough to that of the original cluster, roughly equal to 220.

Now, if we plot the tolerance ellipses for both the main cluster and the sampled cluster with the sampling rate of 0.5 percent, we observe that they are so similar to each other. Fig.~\ref{figure-CovDetAndTolEllpVarRandSampRat}b illustrates such a situation, in which the objects belonging to the original cluster and those associated with the sampled one are illustrated with blue dots and red squares, respectively. Moreover, tolerance ellipses of the main and sampled clusters are shown, in order, in red and black.

\section{Conclusion}\label{sec_Conc}
In this paper, a new scalable density-based clustering approach for local outlier detection in massive data is proposed, which processes the input data in chunks. First of all, by obtaining a random sample of the entire dataset and applying a density-based clustering algorithm to it, the initial temporary clustering model is built, which contains the rough information of the original clusters in data. Then, this model is progressively updated by loading successive chunks of data into the memory. Ultimately, after processing the whole chunks, the final clustering model is acquired, which w.r.t. that and conducting another scan of the entire dataset, each object is given an outlying score equal to its local Mahalanobis distance.

A complete evaluation, conducted on both real-world and synthetic datasets, demonstrates the appealing performance of SDCOR in comparison with different state-of-the-art density-based outlier algorithms, which need the data to be resident in the memory; and also, with some other rapid distance-based anomaly detection methods, which can operate well on the disk-resident data. Moreover, the efficiency outcomes confirm the robustness of the proposed method comparing to other methods in very noisy conditions. In addition, the experiments substantiate that the algorithm has a linear time complexity with a low constant and that, even with a meager rate of random sampling, it is still able to satisfactorily approximate the shape of the real clusters. For future work, we would like to enhance our proposed approach to be able to cope also with the density-based non-convex clusters having various distributions.

\section*{Appendices}\label{sec_appndx}%\addcontentsline{toc}{section}{Appendices}
\appendix

\section{DBSCAN optimal parameters}\label{appSec_DBSCANoptmlParamSampl}
This Appendix describes the ways employed in this paper to determine the optimal values for the DBSCAN parameters to be applied throughout the ``Sampling'' and the ``Scalable Clustering'' phases.

\subsection{\textit{k}-distance graph}\label{appSubSec_kDistGraph}
The first approach is a heuristic suggested by the original paper of DBSCAN \cite[Section 4.2]{ester1996density}, the sorted \textit{k}-dist or the same \textit{k}-distance graph, which is a mapping from the corresponding dataset to the distance of every point to its \textit{k}-th nearest neighbor plotted in descending order. Alternatively stated, in this manner, we fix the \textit{MinPts} parameter, as it is more convenient than the \textit{Eps} parameter to be established, and its main purpose is to polish the density estimation; its variation often has an insignificant influence on the clustering outcomes as they ``do not substantially conflict with each other'' \cite{ester1996density,schubert2017dbscan}. For most of datasets, one can keep this parameter at the predetermined value of $ MinPts = 4 $ \cite{ester1996density} \textemdash~which is mostly recommended for two-dimensional data \textemdash~or $ MinPts = \lfloor\ln \left( n\right)\rfloor $ \cite{birant2007st} which is in accordance with the size of the corresponding input data; furthermore, Sander et al. \cite{sander1998density} suggest to set it to double the number of dimensions, i.e., $ MinPts = 2p $. Besides, it would be wiser, more considerate, and helpful to set this parameter to much higher values to acquire improved results while working on datasets that are very noisy, or large-scale, or high-dimensional, or suffer from many duplicate values; any of the mentioned special conditions or more than one can be enough reason to employ a large value for \textit{MinPts} \cite{schubert2017dbscan}.

Finally, after determining the \textit{MinPts} value, the best value for \textit{Eps} is in consonance with the maximal \textit{k}-dist value in the ``thinnest'' cluster \textemdash~the cluster with the lowest density. For identifying this threshold, we need to visually locate the first point in the first ``valley'' \textemdash~``knee'' or ``elbow'' \textemdash~in the related sorted \textit{k}-distance graph. By utilizing this quantity as the distribution threshold, every point with a higher \textit{k}-dist value that stands at the left of the threshold will be considered as noise or a marginal point; on the contrary, every point with a lower \textit{k}-dist value residing at the right of the threshold will be a core point, thus attributed to some cluster.

Besides the challenges behind selecting the optimal parameter values for DBSCAN, it ought to be contemplated that the distribution of the sampled data is different from the original one, and much more critical, DBSCAN shall be applied on both; hence, it will be required to set distinct values for each situation. As \textit{MinPts} is a less sensitive parameter than \textit{Eps}, we decide to utilize the same quantity for it concerning both the sampled and the original distributions. However, about \textit{Eps}, sometimes even slight variations could incur significant ups and downs in the final results. One exact point is that after sampling, the distance between every point and its \textit{k}-th nearest neighbor expands, and this is a sufficient cause to say that the optimal \textit{Eps} value for the sampled data is meaningfully larger than the required one for the original distribution. Now, if we plot the sorted \textit{k}-distance plot for both the sampled and original data, it could be observed that the first valley for the sampling \textit{k}-dist plot happens earlier and with a greater value than the corresponding valley for the original plot.

\begin{figure*}[pos=t!]
	\centering
	\includegraphics[width=1\linewidth]{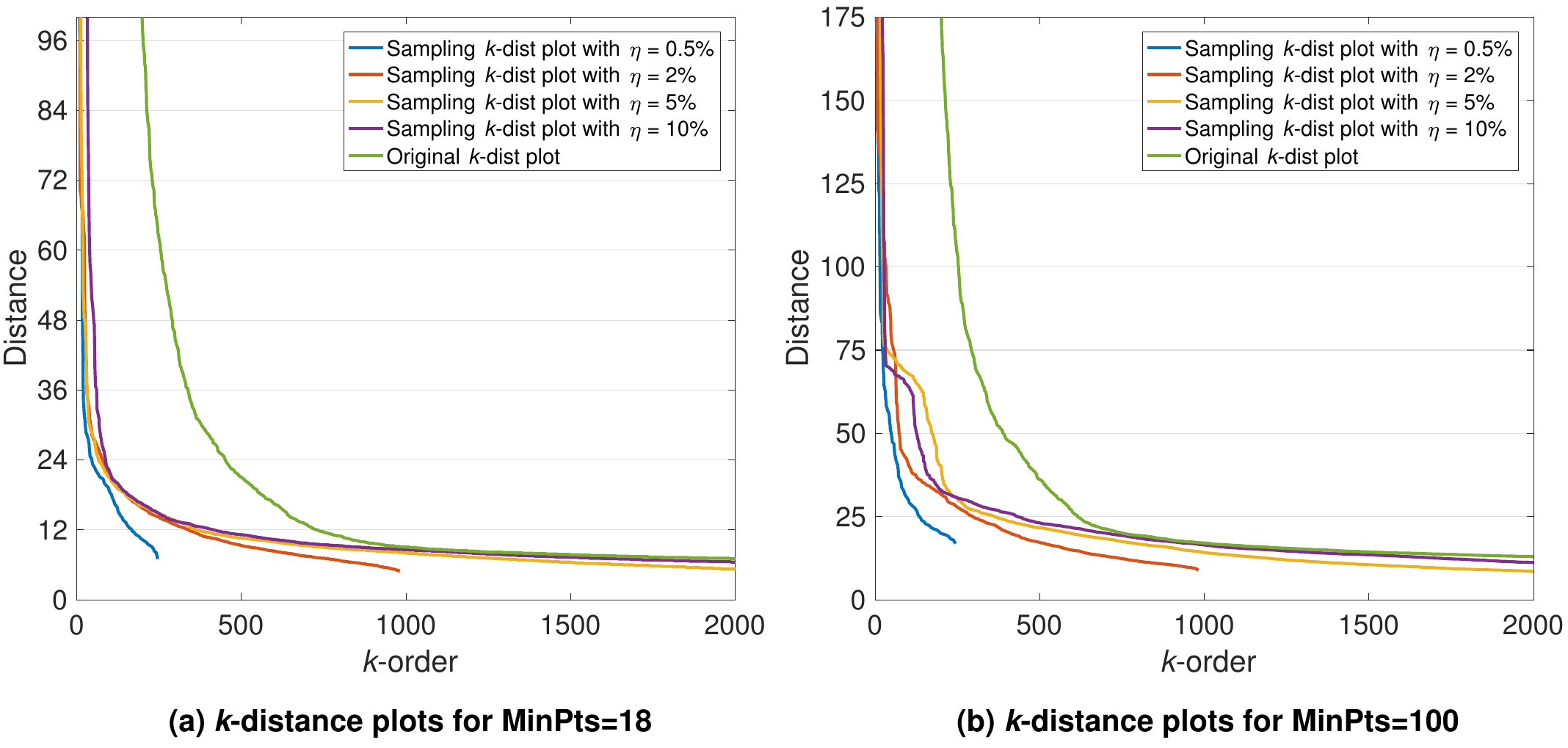}
	\caption{Sampling and original sorted \textit{k}-distance plots for the \textit{Shuttle} dataset with $ k = 17 $ ($ MinPts = 2 \times 9 = 18 $) and $ k = 99 $ ($ MinPts = 100 $)}
	\label{figure-kDistGraph}
\end{figure*}

Fig.~\ref{figure-kDistGraph} illustrates the zoomed-in views of the \textit{k}-distance plots concerning several sampling rates for the \textit{Shuttle} dataset; the blue, red, yellow, and violet lines indicate, in order, the sampling \textit{k}-dist plots for the sampling rates equal to $ 0.5\% $, $ 2\% $, $ 5\% $, and $ 10\% $, while the green line denotes the correspondent \textit{k}-dist graph for the original distribution. Two distinct quantities for \textit{MinPts} w.r.t. the above heuristics are supposed as one in Fig.~\ref{figure-kDistGraph}a, equal to double the \textit{Shuttle} dimensionality ($ 2 \times 9 = 18 $), and the other in Fig.~\ref{figure-kDistGraph}b, established as a much higher value ($ 100 $) to evade the noise\footnote{The proximity parameter, \textit{k}, in the \textit{k}NN search does not include the subject point, but the range query search employed in DBSCAN considers the point itself too for the density evaluation; thus, \textit{k}, here, complies with $ MinPts = k+1 $.} \textemdash~as \textit{Shuttle} contains the most ratio of outliers among real and artificial datasets exerted in our accuracy and stability analysis.

As it is discerned from the illustration, for both \textit{MinPts} specific values, the sampling \textit{k}-dist plots are following almost the same pattern and can be easily distinguished from the relevant original plot; furthermore, in both situations, the first valley for the sampling plots occurs at a higher value than the original one. In Fig.~\ref{figure-kDistGraph}a, for $ MinPts = 18 $, it seems most proper to define the cut around $ 24 $ for the sampled distribution, and about half of this value for the original one\footnote{To avoid the singularity problem while operating on diverse iterations of the proposed method, it is more discreet to set the sampling cut a bit greater, as different random samplings could have slightly different distributions, thus demanding varying requisites.}; in Fig.~\ref{figure-kDistGraph}b, for $ MinPts = 100 $, the best cuts for the sampling and the primary data appears to be approximately $ 50 $ and $ 25 $, respectively\footnote{Higher integers for \textit{MinPts}, naturally incur larger values for \textit{Eps} to satisfy the range query search carried out by DBSCAN.}. As a consequence, as it is not possible to attain the original \textit{k}-dist graph in the usual case because of the following computational expenses, we prefer to set the correspondent \textit{Eps} value for the original distribution during scalable clustering to half of the optimal cut for the sampled data.

\subsection{PSO evolutionary algorithm}\label{appSubSec_PSO}
Despite the advantages connected to utilizing the \textit{k}-distance graph for finding the DBSCAN optimum parameters, sometimes, it is not that easy to interactively locate the first valley position in the related sorted \textit{k}-dist graph as it might be very smooth and without any visual steps (artifacts); furthermore, in some cases, setting the most suitable value for \textit{MinPts} requires a high amount of trial and error. Therefore, one can resort to an optimization algorithm like PSO to conquer the mentioned difficulties. PSO looks for the optimal cases in a search space having a dimensionality on a par with the number of demanded parameters by the particular problem; \textit{Eps} and \textit{MinPts} are the required parameters here, and their lower and upper bounds need to be determined for building the search zone employed by the PSO algorithm while being applied to the sampled data. For the original distribution requisite parameters, regarding the impossibility of putting PSO into practice for the original massive data, the same scenario will be followed as in the \textit{k}-distance graph strategy; half of the acquired sampling optimum \textit{Eps} for the primary data during the batch-wise clustering, and the same \textit{MinPts} for both situations.

Inspired by the approximate analogousness of the plots in Fig.~\ref{figure-kDistGraph}, and that the optimal quantity for \textit{Eps} lies between the minimum and maximum of any of these plots \textemdash~which are almost the same in various sampling conditions, and also the original graph \textemdash~we decide to establish the lower and upper survey bounds for the \textit{Eps} parameter as the minimum and maximum values of the \textit{k}-dist graph associated with the corresponding sampled data. For the \textit{MinPts} parameter, w.r.t. the already mentioned heuristics, we prefer to set the lower search bound to $ \lfloor\ln \left( n\right)\rfloor $ \textemdash~as it is a dynamic number according to every certain dataset \textemdash~and the subsequent upper bound could be asked through the user as a high-enough integer to avoid anomalies; the most suitable \textit{MinPts} value will be lying in this range. In the following, an explanation of the PSO algorithm and the peculiar evaluating manner for its optimization plan exerted in this study will be provided.

PSO is an evolutionary population-based algorithm oriented from the corporate behavior of some animal gatherings, in particular, flocks and shoals; it is totally founded upon mathematical concepts and can solve intricate arithmetical optimization problems \cite{nickabadi2011novel,de2019particle}. One of the greatest virtues of utilizing PSO among various optimization strategies is its dependence on a fewer number of parameters; besides, the requisite ones are extensively debated in the scientific literature.

Here, we laconically explain the classical or the same inertial version of the algorithm. Like any optimization problem, the target is to figure out a variable vector \textemdash~also known as the position vector \textemdash~represented as $ X = \left( x_1,\cdots ,x_n\right) $ which is about to minimize or maximize a specific optimization function denoted as $ f\left( X\right) $ \textemdash~also called the fitness, the objective or the cost function. $ X $ is an $ n $-dimensional vector which stands for the $ n $ distinct unknown variables in the intended problem\footnote{The related variables for DBSCAN are its two input parameters, \textit{Eps} and \textit{MinPts}, as it was already discussed.}, and $ f\left( X\right) $ is a multi-criteria evaluation metric that appraises the reliability of a particular condition specified by the position vector $ X $. If we consider a swarm with $ P $ particles, there would be a position vector along with a velocity vector for each particle; every position vector is a capable solution to the objective function and is updated through successive iterations according to the knowledge of the particle itself and other participating ones in the group. We illustrate particle $ i $ at iteration $ t $ of the solving procedure as $ X_{i}^{t} = \left( x_{i1},\cdots x_{in}\right) $, and the corresponding velocity vector as $ V_{i}^{t} = \left( v_{i1},\cdots v_{in}\right) $. Each particle is bounded in a distinct interval, which in total, all of these intervals build the search space required for resolving the optimization function. At each iteration, the velocity and position vectors for particle $ i $ at the $ j $-th dimension are updated as follows:

\begin{equation}
	\label{equ_PSOvelVec}
	\begin{aligned}
		V_{ij}^{t+1} = wV_{ij}^{t} + c_1r_1^t\left( pbest_{ij} - X_{ij}^t \right) \\
		+ c_2r_2^t\left( gbest_j - X_{ij}^t \right)
	\end{aligned}
\end{equation}

\begin{equation}
	\label{equ_PSOposVec}
	X_{ij}^{t+1} = X_{ij}^{t} + V_{ij}^{t+1}
\end{equation}

where $ i = 1,\cdots ,P $ and $ j = 1,\cdots ,n $.

Regarding Eq. (\ref{equ_PSOvelVec}), the updating formula of the velocity vector in every dimension, three different terms contribute to a particle motion over any iteration. In the first term, parameter $ w $ is the inertia weight constant with a positive value. This parameter handles the balance between the global search (exploration) and the local search (exploitation) of the algorithm. The second and the third terms are, in order, the individual perception term and the collective learning term. The individual perception term takes into account the difference between the particle's own previous best location ($ pbest_i $) and its current position ($ X_i^t $); in contrast, the collective learning term addresses the inclination towards the previous best-identified location among all of the involved particles ($ gbest $). Parameters $ c_1 $ and $ c_2 $ are acceleration constants that weigh the personal and social understanding of every particle, respectively. Parameters $ r_1 $ and $ r_2 $ are two random quantities in the range of $ \left[0,1\right] $ which help the optimization problem not to fall early in a local optimum resolution; these random values are reset at every iteration. Finally, the corresponding position of every particle in each dimension will be updated w.r.t. Eq. (\ref{equ_PSOposVec}), and in the case that it is out of the specified searching limits, it will get the equivalent violated bound quantity.

PSO consists of two separate sections, one initialization part, and one iterative component. In the initialization section, firstly, the position and velocity vectors for every particle get primary values, and then after evaluating the alignment of the existing positions through the fitness function, the associated best local and global positions are established. Consequently, in the iterative section, at every repetition, the velocity and position vectors are updated in all of their dimensions according to Eqs. (\ref{equ_PSOvelVec}) and (\ref{equ_PSOposVec}); then, by employing the objective function again, the related best local and global locations are modified until a termination condition is satisfied.

For the PSO cost function in our experiments, we have utilized the aggregation of two internal clustering validity indices \cite{desgraupes2013clustering}, namely the Davies-Bouldin index \cite{davies1979cluster} and the CS index \cite{chou2003new,chou2004new}, along with the ratio of detected outliers \cite{guan2019particle}, which requires to be minimized. To be noted, the CS index incurs a high calculational load, especially on large datasets, yet it was very fruitful for our analysis. By the way, other indices with various arithmetic expenses \textemdash~including but not limited to the Dunn index \cite{dunn1974well}, the Silhouette index \cite{rousseeuw1987silhouettes}, the Banfeld-Raftery index \cite{banfield1993model}, and the Maulik-Bandyopadhyay index (Index \textit{I}) \cite{maulik2002performance} \textemdash~could be employed too, although it is not guaranteed that better outcomes will undoubtedly be achieved by applying a more computationally expensive index and vice versa.

Moreover, we have defined three special conditions for the cost function as the infinite value to evade them. One is when after applying DBSCAN to the sampled data with the specific parameters out of a particular particle, no cluster is detected, and, i.e., every object is introduced as noise; such parameters are out of order w.r.t. the accepted sampling terms. The other condition is when every sampled point is assigned to some cluster without any noise being identified; this is against our presumption that the input data always contains some outliers that need to be separated. Finally, there is another unacceptable situation that happens when at least one of the discovered sampled clusters through DBSCAN is suffering from the singularity problem; with the optimum parameters in a condition conforming with the proposed method strong assumptions, there should not be any detected sampled cluster with the singularity issue. Besides, while we are applying internal cluster validation indices to a clustering result out of a particle parameters, detected noisy objects are assumed as a unique cluster, too, as they should play an active role in the arrangement evaluation of the DBSCAN outcome \cite{saitta2007bounded}.

% CRediT authorship contribution statement
\printcredits

\section*{Declaration of competing interest}The authors declare that they have no known competing financial interests or personal relationships that could have appeared
to influence the work reported in this paper.

\section*{Acknowledgments}\label{sec_Acknlg}Special thanks to Dr. Victoria J. Hodge (University of York, UK) for her clever advice on the title of this paper, and Dr. Mohammad Mehdi Ebadzadeh (Amirkabir University of Technology, Iran), for his valuable comments on some of the algorithms. Moreover, we are grateful to Dr. Fabrizio Angiulli (University of Calabria) and Dr. Dan Pelleg (Yahoo Labs) for providing us with the DOLPHIN and X-means implementation codes, respectively. We also appreciate Dr. Ali-Mohammad Saghiri and Dr. Ehsan Nazerfard (Amirkabir University of Technology, Iran) for their generous review of the paper before submission.

\bibliographystyle{cas-model2-names}

\bibliography{SDCOR-refs}

\begin{thebibliography}{140}
\expandafter\ifx\csname natexlab\endcsname\relax\def\natexlab#1{#1}\fi
\providecommand{\url}[1]{\texttt{#1}}
\providecommand{\href}[2]{#2}
\providecommand{\path}[1]{#1}
\providecommand{\DOIprefix}{doi:}
\providecommand{\ArXivprefix}{arXiv:}
\providecommand{\URLprefix}{URL: }
\providecommand{\Pubmedprefix}{pmid:}
\providecommand{\doi}[1]{\href{http://dx.doi.org/#1}{\path{#1}}}
\providecommand{\Pubmed}[1]{\href{pmid:#1}{\path{#1}}}
\providecommand{\bibinfo}[2]{#2}
\ifx\xfnm\relax \def\xfnm[#1]{\unskip,\space#1}\fi
%Type = Inproceedings
\bibitem[{Achlioptas(2001)}]{achlioptas2001database}
\bibinfo{author}{Achlioptas, D.}, \bibinfo{year}{2001}.
\newblock \bibinfo{title}{Database-friendly random projections}, in:
  \bibinfo{booktitle}{Proceedings of the twentieth ACM SIGMOD-SIGACT-SIGART
  symposium on Principles of database systems}, \bibinfo{organization}{ACM}.
  pp. \bibinfo{pages}{274--281}.
%Type = Book
\bibitem[{Aggarwal(2015a)}]{aggarwal2015data}
\bibinfo{author}{Aggarwal, C.C.}, \bibinfo{year}{2015}a.
\newblock \bibinfo{title}{Data mining: the textbook}.
\newblock \bibinfo{publisher}{Springer}.
%Type = Inproceedings
\bibitem[{Aggarwal(2015b)}]{aggarwal2015outlier}
\bibinfo{author}{Aggarwal, C.C.}, \bibinfo{year}{2015}b.
\newblock \bibinfo{title}{Outlier analysis}, in: \bibinfo{booktitle}{Data
  mining}, \bibinfo{organization}{Springer}. pp. \bibinfo{pages}{237--263}.
%Type = Article
\bibitem[{Agyemang et~al.(2006)Agyemang, Barker and
  Alhajj}]{agyemang2006comprehensive}
\bibinfo{author}{Agyemang, M.}, \bibinfo{author}{Barker, K.},
  \bibinfo{author}{Alhajj, R.}, \bibinfo{year}{2006}.
\newblock \bibinfo{title}{A comprehensive survey of numeric and symbolic
  outlier mining techniques}.
\newblock \bibinfo{journal}{Intelligent Data Analysis} \bibinfo{volume}{10},
  \bibinfo{pages}{521--538}.
%Type = Article
\bibitem[{Akoglu et~al.(2015)Akoglu, Tong and Koutra}]{akoglu2015graph}
\bibinfo{author}{Akoglu, L.}, \bibinfo{author}{Tong, H.},
  \bibinfo{author}{Koutra, D.}, \bibinfo{year}{2015}.
\newblock \bibinfo{title}{Graph based anomaly detection and description: a
  survey}.
\newblock \bibinfo{journal}{Data mining and knowledge discovery}
  \bibinfo{volume}{29}, \bibinfo{pages}{626--688}.
%Type = Article
\bibitem[{Alguliyev et~al.(2017)Alguliyev, Aliguliyev and
  Sukhostat}]{alguliyev2017anomaly}
\bibinfo{author}{Alguliyev, R.}, \bibinfo{author}{Aliguliyev, R.},
  \bibinfo{author}{Sukhostat, L.}, \bibinfo{year}{2017}.
\newblock \bibinfo{title}{Anomaly detection in big data based on clustering}.
\newblock \bibinfo{journal}{Statistics, Optimization \& Information Computing}
  \bibinfo{volume}{5}, \bibinfo{pages}{325--340}.
%Type = Article
\bibitem[{Aliguliyev(2009)}]{aliguliyev2009performance}
\bibinfo{author}{Aliguliyev, R.M.}, \bibinfo{year}{2009}.
\newblock \bibinfo{title}{Performance evaluation of density-based clustering
  methods}.
\newblock \bibinfo{journal}{Information Sciences} \bibinfo{volume}{179},
  \bibinfo{pages}{3583--3602}.
%Type = Incollection
\bibitem[{de~Almeida and Leite(2019)}]{de2019particle}
\bibinfo{author}{de~Almeida, B.S.G.}, \bibinfo{author}{Leite, V.C.},
  \bibinfo{year}{2019}.
\newblock \bibinfo{title}{Particle swarm optimization: A powerful technique for
  solving engineering problems}, in: \bibinfo{booktitle}{Swarm
  Intelligence-Recent Advances, New Perspectives and Applications}.
  \bibinfo{publisher}{IntechOpen}.
%Type = Article
\bibitem[{Amil~Marletti et~al.(2019)Amil~Marletti, Almeira and
  Masoller~Alonso}]{amil2019outlier}
\bibinfo{author}{Amil~Marletti, P.}, \bibinfo{author}{Almeira, N.},
  \bibinfo{author}{Masoller~Alonso, C.}, \bibinfo{year}{2019}.
\newblock \bibinfo{title}{Outlier mining methods based on graph structure
  analysis}.
\newblock \bibinfo{journal}{Frontiers in Physics} \bibinfo{volume}{7},
  \bibinfo{pages}{1--11}.
%Type = Article
\bibitem[{Angiulli et~al.(2012)Angiulli, Basta, Lodi and
  Sartori}]{angiulli2012distributed}
\bibinfo{author}{Angiulli, F.}, \bibinfo{author}{Basta, S.},
  \bibinfo{author}{Lodi, S.}, \bibinfo{author}{Sartori, C.},
  \bibinfo{year}{2012}.
\newblock \bibinfo{title}{Distributed strategies for mining outliers in large
  data sets}.
\newblock \bibinfo{journal}{IEEE Transactions on Knowledge and Data
  Engineering} \bibinfo{volume}{25}, \bibinfo{pages}{1520--1532}.
%Type = Article
\bibitem[{Angiulli and Fassetti(2009)}]{angiulli2009dolphin}
\bibinfo{author}{Angiulli, F.}, \bibinfo{author}{Fassetti, F.},
  \bibinfo{year}{2009}.
\newblock \bibinfo{title}{Dolphin: An efficient algorithm for mining
  distance-based outliers in very large datasets}.
\newblock \bibinfo{journal}{ACM Transactions on Knowledge Discovery from Data
  (TKDD)} \bibinfo{volume}{3}, \bibinfo{pages}{1--57}.
%Type = Article
\bibitem[{Ayy{\i}ld{\i}z et~al.(2012)Ayy{\i}ld{\i}z, Purut{\c{c}}uoglu and
  Wit}]{ayyildiz2012short}
\bibinfo{author}{Ayy{\i}ld{\i}z, E.}, \bibinfo{author}{Purut{\c{c}}uoglu, V.},
  \bibinfo{author}{Wit, E.}, \bibinfo{year}{2012}.
\newblock \bibinfo{title}{A short note on resolving singularity problems in
  covariance matrices}.
\newblock \bibinfo{journal}{International Journal of Statistics and
  Probability} \bibinfo{volume}{1}, \bibinfo{pages}{113--118}.
%Type = Inproceedings
\bibitem[{Bandaragoda et~al.(2014)Bandaragoda, Ting, Albrecht, Liu and
  Wells}]{bandaragoda2014efficient}
\bibinfo{author}{Bandaragoda, T.R.}, \bibinfo{author}{Ting, K.M.},
  \bibinfo{author}{Albrecht, D.}, \bibinfo{author}{Liu, F.T.},
  \bibinfo{author}{Wells, J.R.}, \bibinfo{year}{2014}.
\newblock \bibinfo{title}{Efficient anomaly detection by isolation using
  nearest neighbour ensemble}, in: \bibinfo{booktitle}{2014 IEEE International
  Conference on Data Mining Workshop}, \bibinfo{organization}{IEEE}. pp.
  \bibinfo{pages}{698--705}.
%Type = Article
\bibitem[{Bandaragoda et~al.(2018)Bandaragoda, Ting, Albrecht, Liu, Zhu and
  Wells}]{bandaragoda2018isolation}
\bibinfo{author}{Bandaragoda, T.R.}, \bibinfo{author}{Ting, K.M.},
  \bibinfo{author}{Albrecht, D.}, \bibinfo{author}{Liu, F.T.},
  \bibinfo{author}{Zhu, Y.}, \bibinfo{author}{Wells, J.R.},
  \bibinfo{year}{2018}.
\newblock \bibinfo{title}{Isolation-based anomaly detection using
  nearest-neighbor ensembles}.
\newblock \bibinfo{journal}{Computational Intelligence} \bibinfo{volume}{34},
  \bibinfo{pages}{968--998}.
%Type = Article
\bibitem[{Banfield and Raftery(1993)}]{banfield1993model}
\bibinfo{author}{Banfield, J.D.}, \bibinfo{author}{Raftery, A.E.},
  \bibinfo{year}{1993}.
\newblock \bibinfo{title}{Model-based gaussian and non-gaussian clustering}.
\newblock \bibinfo{journal}{Biometrics} , \bibinfo{pages}{803--821}.
%Type = Book
\bibitem[{Barnett and Lewis(1974)}]{barnett1974outliers}
\bibinfo{author}{Barnett, V.}, \bibinfo{author}{Lewis, T.},
  \bibinfo{year}{1974}.
\newblock \bibinfo{title}{Outliers in statistical data}.
\newblock \bibinfo{publisher}{Wiley}.
%Type = Inproceedings
\bibitem[{Bay and Schwabacher(2003)}]{bay2003mining}
\bibinfo{author}{Bay, S.D.}, \bibinfo{author}{Schwabacher, M.},
  \bibinfo{year}{2003}.
\newblock \bibinfo{title}{Mining distance-based outliers in near linear time
  with randomization and a simple pruning rule}, in:
  \bibinfo{booktitle}{Proceedings of the ninth ACM SIGKDD international
  conference on Knowledge discovery and data mining}, pp.
  \bibinfo{pages}{29--38}.
%Type = Article
\bibitem[{Bentley(1975)}]{bentley1975multidimensional}
\bibinfo{author}{Bentley, J.L.}, \bibinfo{year}{1975}.
\newblock \bibinfo{title}{Multidimensional binary search trees used for
  associative searching}.
\newblock \bibinfo{journal}{Communications of the ACM} \bibinfo{volume}{18},
  \bibinfo{pages}{509--517}.
%Type = Article
\bibitem[{Bie et~al.(2016)Bie, Mehmood, Ruan, Sun and Dawood}]{bie2016adaptive}
\bibinfo{author}{Bie, R.}, \bibinfo{author}{Mehmood, R.},
  \bibinfo{author}{Ruan, S.}, \bibinfo{author}{Sun, Y.},
  \bibinfo{author}{Dawood, H.}, \bibinfo{year}{2016}.
\newblock \bibinfo{title}{Adaptive fuzzy clustering by fast search and find of
  density peaks}.
\newblock \bibinfo{journal}{Personal and Ubiquitous Computing}
  \bibinfo{volume}{20}, \bibinfo{pages}{785--793}.
%Type = Article
\bibitem[{Birant and Kut(2007)}]{birant2007st}
\bibinfo{author}{Birant, D.}, \bibinfo{author}{Kut, A.}, \bibinfo{year}{2007}.
\newblock \bibinfo{title}{St-dbscan: An algorithm for clustering
  spatial--temporal data}.
\newblock \bibinfo{journal}{Data \& knowledge engineering}
  \bibinfo{volume}{60}, \bibinfo{pages}{208--221}.
%Type = Inproceedings
\bibitem[{Boutin and Hasco{\"e}t(2004)}]{boutin2004cluster}
\bibinfo{author}{Boutin, F.}, \bibinfo{author}{Hasco{\"e}t, M.},
  \bibinfo{year}{2004}.
\newblock \bibinfo{title}{Cluster validity indices for graph partitioning}, in:
  \bibinfo{booktitle}{Proceedings. Eighth International Conference on
  Information Visualisation, 2004. IV 2004.}, \bibinfo{organization}{IEEE}. pp.
  \bibinfo{pages}{376--381}.
%Type = Inproceedings
\bibitem[{Bradley et~al.(1998)Bradley, Fayyad, Reina
  et~al.}]{bradley1998scaling}
\bibinfo{author}{Bradley, P.S.}, \bibinfo{author}{Fayyad, U.M.},
  \bibinfo{author}{Reina, C.}, et~al., \bibinfo{year}{1998}.
\newblock \bibinfo{title}{Scaling clustering algorithms to large databases.},
  in: \bibinfo{booktitle}{KDD}, pp. \bibinfo{pages}{9--15}.
%Type = Article
\bibitem[{Breiman(2001)}]{breiman2001random}
\bibinfo{author}{Breiman, L.}, \bibinfo{year}{2001}.
\newblock \bibinfo{title}{Random forests}.
\newblock \bibinfo{journal}{Machine learning} \bibinfo{volume}{45},
  \bibinfo{pages}{5--32}.
%Type = Inproceedings
\bibitem[{Breunig et~al.(2000)Breunig, Kriegel, Ng and Sander}]{breunig2000lof}
\bibinfo{author}{Breunig, M.M.}, \bibinfo{author}{Kriegel, H.P.},
  \bibinfo{author}{Ng, R.T.}, \bibinfo{author}{Sander, J.},
  \bibinfo{year}{2000}.
\newblock \bibinfo{title}{Lof: identifying density-based local outliers}, in:
  \bibinfo{booktitle}{ACM sigmod record}, \bibinfo{organization}{ACM}. pp.
  \bibinfo{pages}{93--104}.
%Type = Article
\bibitem[{Bryant and Cios(2017)}]{bryant2017rnn}
\bibinfo{author}{Bryant, A.}, \bibinfo{author}{Cios, K.}, \bibinfo{year}{2017}.
\newblock \bibinfo{title}{Rnn-dbscan: A density-based clustering algorithm
  using reverse nearest neighbor density estimates}.
\newblock \bibinfo{journal}{IEEE Transactions on Knowledge and Data
  Engineering} \bibinfo{volume}{30}, \bibinfo{pages}{1109--1121}.
%Type = Article
\bibitem[{Cabras and Morales(2007)}]{cabras2007extreme}
\bibinfo{author}{Cabras, S.}, \bibinfo{author}{Morales, J.},
  \bibinfo{year}{2007}.
\newblock \bibinfo{title}{Extreme value analysis within a parametric outlier
  detection framework}.
\newblock \bibinfo{journal}{Applied Stochastic Models in Business and Industry}
  \bibinfo{volume}{23}, \bibinfo{pages}{157--164}.
%Type = Article
\bibitem[{Campos et~al.(2016)Campos, Zimek, Sander, Campello, Micenkov{\'a},
  Schubert, Assent and Houle}]{campos2016evaluation}
\bibinfo{author}{Campos, G.O.}, \bibinfo{author}{Zimek, A.},
  \bibinfo{author}{Sander, J.}, \bibinfo{author}{Campello, R.J.},
  \bibinfo{author}{Micenkov{\'a}, B.}, \bibinfo{author}{Schubert, E.},
  \bibinfo{author}{Assent, I.}, \bibinfo{author}{Houle, M.E.},
  \bibinfo{year}{2016}.
\newblock \bibinfo{title}{On the evaluation of unsupervised outlier detection:
  measures, datasets, and an empirical study}.
\newblock \bibinfo{journal}{Data mining and knowledge discovery}
  \bibinfo{volume}{30}, \bibinfo{pages}{891--927}.
%Type = Article
\bibitem[{Chandola et~al.(2009)Chandola, Banerjee and
  Kumar}]{chandola2009anomaly}
\bibinfo{author}{Chandola, V.}, \bibinfo{author}{Banerjee, A.},
  \bibinfo{author}{Kumar, V.}, \bibinfo{year}{2009}.
\newblock \bibinfo{title}{Anomaly detection: A survey}.
\newblock \bibinfo{journal}{ACM computing surveys (CSUR)} \bibinfo{volume}{41},
  \bibinfo{pages}{15}.
%Type = Article
\bibitem[{Chen et~al.(2018)Chen, Azer and Zhang}]{chen2018practical}
\bibinfo{author}{Chen, J.}, \bibinfo{author}{Azer, E.S.},
  \bibinfo{author}{Zhang, Q.}, \bibinfo{year}{2018}.
\newblock \bibinfo{title}{A practical algorithm for distributed clustering and
  outlier detection}.
\newblock \bibinfo{journal}{arXiv preprint arXiv:1805.09495} .
%Type = Article
\bibitem[{Chen et~al.(2011)Chen, Liu, Qiu and Lai}]{chen2011apscan}
\bibinfo{author}{Chen, X.}, \bibinfo{author}{Liu, W.}, \bibinfo{author}{Qiu,
  H.}, \bibinfo{author}{Lai, J.}, \bibinfo{year}{2011}.
\newblock \bibinfo{title}{Apscan: A parameter free algorithm for clustering}.
\newblock \bibinfo{journal}{Pattern Recognition Letters} \bibinfo{volume}{32},
  \bibinfo{pages}{973--986}.
%Type = Inproceedings
\bibitem[{Chou et~al.(2003)Chou, Su and Lai}]{chou2003new}
\bibinfo{author}{Chou, C.H.}, \bibinfo{author}{Su, M.C.}, \bibinfo{author}{Lai,
  E.}, \bibinfo{year}{2003}.
\newblock \bibinfo{title}{A new cluster validity measure for clusters with
  different densities}, in: \bibinfo{booktitle}{IASTED International Conference
  on Intelligent Systems and Control}, pp. \bibinfo{pages}{276--281}.
%Type = Article
\bibitem[{Chou et~al.(2004)Chou, Su and Lai}]{chou2004new}
\bibinfo{author}{Chou, C.H.}, \bibinfo{author}{Su, M.C.}, \bibinfo{author}{Lai,
  E.}, \bibinfo{year}{2004}.
\newblock \bibinfo{title}{A new cluster validity measure and its application to
  image compression}.
\newblock \bibinfo{journal}{Pattern Analysis and Applications}
  \bibinfo{volume}{7}, \bibinfo{pages}{205--220}.
%Type = Article
\bibitem[{Cook and Holder(2000)}]{cook2000graph}
\bibinfo{author}{Cook, D.J.}, \bibinfo{author}{Holder, L.B.},
  \bibinfo{year}{2000}.
\newblock \bibinfo{title}{Graph-based data mining}.
\newblock \bibinfo{journal}{IEEE Intelligent Systems and Their Applications}
  \bibinfo{volume}{15}, \bibinfo{pages}{32--41}.
%Type = Book
\bibitem[{Cover(1999)}]{cover1999elements}
\bibinfo{author}{Cover, T.M.}, \bibinfo{year}{1999}.
\newblock \bibinfo{title}{Elements of information theory}.
\newblock \bibinfo{publisher}{John Wiley \& Sons}.
%Type = Inproceedings
\bibitem[{Dang et~al.(2013)Dang, Micenkov{\'a}, Assent and Ng}]{dang2013local}
\bibinfo{author}{Dang, X.H.}, \bibinfo{author}{Micenkov{\'a}, B.},
  \bibinfo{author}{Assent, I.}, \bibinfo{author}{Ng, R.T.},
  \bibinfo{year}{2013}.
\newblock \bibinfo{title}{Local outlier detection with interpretation}, in:
  \bibinfo{booktitle}{Joint European Conference on Machine Learning and
  Knowledge Discovery in Databases}, \bibinfo{organization}{Springer}. pp.
  \bibinfo{pages}{304--320}.
%Type = Article
\bibitem[{Dasgupta and Gupta(1999)}]{dasgupta1999elementary}
\bibinfo{author}{Dasgupta, S.}, \bibinfo{author}{Gupta, A.},
  \bibinfo{year}{1999}.
\newblock \bibinfo{title}{An elementary proof of the johnson-lindenstrauss
  lemma}.
\newblock \bibinfo{journal}{International Computer Science Institute, Technical
  Report} \bibinfo{volume}{22}, \bibinfo{pages}{1--5}.
%Type = Article
\bibitem[{Davies and Bouldin(1979)}]{davies1979cluster}
\bibinfo{author}{Davies, D.L.}, \bibinfo{author}{Bouldin, D.W.},
  \bibinfo{year}{1979}.
\newblock \bibinfo{title}{A cluster separation measure}.
\newblock \bibinfo{journal}{IEEE transactions on pattern analysis and machine
  intelligence} , \bibinfo{pages}{224--227}.
%Type = Inproceedings
\bibitem[{Davis and Goadrich(2006)}]{davis2006relationship}
\bibinfo{author}{Davis, J.}, \bibinfo{author}{Goadrich, M.},
  \bibinfo{year}{2006}.
\newblock \bibinfo{title}{The relationship between precision-recall and roc
  curves}, in: \bibinfo{booktitle}{Proceedings of the 23rd international
  conference on Machine learning}, pp. \bibinfo{pages}{233--240}.
%Type = Inproceedings
\bibitem[{De~Vries et~al.(2010)De~Vries, Chawla and Houle}]{de2010finding}
\bibinfo{author}{De~Vries, T.}, \bibinfo{author}{Chawla, S.},
  \bibinfo{author}{Houle, M.E.}, \bibinfo{year}{2010}.
\newblock \bibinfo{title}{Finding local anomalies in very high dimensional
  space}, in: \bibinfo{booktitle}{2010 IEEE International Conference on Data
  Mining}, \bibinfo{organization}{IEEE}. pp. \bibinfo{pages}{128--137}.
%Type = Article
\bibitem[{Demsar(2006)}]{demsar06}
\bibinfo{author}{Demsar, J.}, \bibinfo{year}{2006}.
\newblock \bibinfo{title}{Statistical comparisons of classifiers over multiple
  data sets}.
\newblock \bibinfo{journal}{Journal of Machine Learning Research}
  \bibinfo{volume}{7}, \bibinfo{pages}{1--30}.
%Type = Article
\bibitem[{Desgraupes(2013)}]{desgraupes2013clustering}
\bibinfo{author}{Desgraupes, B.}, \bibinfo{year}{2013}.
\newblock \bibinfo{title}{Clustering indices}.
\newblock \bibinfo{journal}{University of Paris Ouest-Lab Modal’X}
  \bibinfo{volume}{1}, \bibinfo{pages}{34}.
%Type = Article
\bibitem[{Domingues et~al.(2018)Domingues, Filippone, Michiardi and
  Zouaoui}]{domingues2018comparative}
\bibinfo{author}{Domingues, R.}, \bibinfo{author}{Filippone, M.},
  \bibinfo{author}{Michiardi, P.}, \bibinfo{author}{Zouaoui, J.},
  \bibinfo{year}{2018}.
\newblock \bibinfo{title}{A comparative evaluation of outlier detection
  algorithms: Experiments and analyses}.
\newblock \bibinfo{journal}{Pattern Recognition} \bibinfo{volume}{74},
  \bibinfo{pages}{406--421}.
%Type = Misc
\bibitem[{Dua and Graff(2017)}]{Dua:2019}
\bibinfo{author}{Dua, D.}, \bibinfo{author}{Graff, C.}, \bibinfo{year}{2017}.
\newblock \bibinfo{title}{{UCI} machine learning repository}.
\newblock \URLprefix \url{http://archive.ics.uci.edu/ml}.
%Type = Article
\bibitem[{Duan et~al.(2009)Duan, Xu, Liu and Lee}]{duan2009cluster}
\bibinfo{author}{Duan, L.}, \bibinfo{author}{Xu, L.}, \bibinfo{author}{Liu,
  Y.}, \bibinfo{author}{Lee, J.}, \bibinfo{year}{2009}.
\newblock \bibinfo{title}{Cluster-based outlier detection}.
\newblock \bibinfo{journal}{Annals of Operations Research}
  \bibinfo{volume}{168}, \bibinfo{pages}{151--168}.
%Type = Article
\bibitem[{Dunn(1974)}]{dunn1974well}
\bibinfo{author}{Dunn, J.C.}, \bibinfo{year}{1974}.
\newblock \bibinfo{title}{Well-separated clusters and optimal fuzzy
  partitions}.
\newblock \bibinfo{journal}{Journal of cybernetics} \bibinfo{volume}{4},
  \bibinfo{pages}{95--104}.
%Type = Inproceedings
\bibitem[{Ester et~al.(1996)Ester, Kriegel, Sander, Xu
  et~al.}]{ester1996density}
\bibinfo{author}{Ester, M.}, \bibinfo{author}{Kriegel, H.P.},
  \bibinfo{author}{Sander, J.}, \bibinfo{author}{Xu, X.}, et~al.,
  \bibinfo{year}{1996}.
\newblock \bibinfo{title}{A density-based algorithm for discovering clusters in
  large spatial databases with noise.}, in: \bibinfo{booktitle}{Kdd}, pp.
  \bibinfo{pages}{226--231}.
%Type = Article
\bibitem[{Filzmoser et~al.(2008)Filzmoser, Maronna and
  Werner}]{filzmoser2008outlier}
\bibinfo{author}{Filzmoser, P.}, \bibinfo{author}{Maronna, R.},
  \bibinfo{author}{Werner, M.}, \bibinfo{year}{2008}.
\newblock \bibinfo{title}{Outlier identification in high dimensions}.
\newblock \bibinfo{journal}{Computational Statistics \& Data Analysis}
  \bibinfo{volume}{52}, \bibinfo{pages}{1694--1711}.
%Type = Article
\bibitem[{Forgey(1965)}]{forgey1965cluster}
\bibinfo{author}{Forgey, E.}, \bibinfo{year}{1965}.
\newblock \bibinfo{title}{Cluster analysis of multivariate data: Efficiency vs.
  interpretability of classification}.
\newblock \bibinfo{journal}{Biometrics} \bibinfo{volume}{21},
  \bibinfo{pages}{768--769}.
%Type = Article
\bibitem[{Friedman(1937)}]{friedman1937use}
\bibinfo{author}{Friedman, M.}, \bibinfo{year}{1937}.
\newblock \bibinfo{title}{The use of ranks to avoid the assumption of normality
  implicit in the analysis of variance}.
\newblock \bibinfo{journal}{Journal of the american statistical association}
  \bibinfo{volume}{32}, \bibinfo{pages}{675--701}.
%Type = Article
\bibitem[{García and Herrera(2009)}]{Garcia09}
\bibinfo{author}{García, S.}, \bibinfo{author}{Herrera, F.},
  \bibinfo{year}{2009}.
\newblock \bibinfo{title}{An extension on "statistical comparisons of
  classifiers over multiple data sets" for all pairwise comparisons}.
\newblock \bibinfo{journal}{Journal of Machine Learning Research}
  \bibinfo{volume}{9}, \bibinfo{pages}{2677--2694}.
%Type = Misc
\bibitem[{Goldbloom and Hamner(2010)}]{kaggle}
\bibinfo{author}{Goldbloom, A.}, \bibinfo{author}{Hamner, B.},
  \bibinfo{year}{2010}.
\newblock \bibinfo{title}{{Kaggle} data science company}.
\newblock \URLprefix \url{https://www.kaggle.com}.
%Type = Article
\bibitem[{Guan et~al.(2019)Guan, Yuen and Coenen}]{guan2019particle}
\bibinfo{author}{Guan, C.}, \bibinfo{author}{Yuen, K.K.F.},
  \bibinfo{author}{Coenen, F.}, \bibinfo{year}{2019}.
\newblock \bibinfo{title}{Particle swarm optimized density-based clustering and
  classification: Supervised and unsupervised learning approaches}.
\newblock \bibinfo{journal}{Swarm and evolutionary computation}
  \bibinfo{volume}{44}, \bibinfo{pages}{876--896}.
%Type = Book
\bibitem[{Han et~al.(2011)Han, Pei and Kamber}]{han2011data}
\bibinfo{author}{Han, J.}, \bibinfo{author}{Pei, J.}, \bibinfo{author}{Kamber,
  M.}, \bibinfo{year}{2011}.
\newblock \bibinfo{title}{Data mining: concepts and techniques}.
\newblock \bibinfo{publisher}{Elsevier}.
%Type = Book
\bibitem[{Hawkins(1980)}]{hawkins1980identification}
\bibinfo{author}{Hawkins, D.M.}, \bibinfo{year}{1980}.
\newblock \bibinfo{title}{Identification of outliers}.
  volume~\bibinfo{volume}{11}.
\newblock \bibinfo{publisher}{Springer}.
%Type = Article
\bibitem[{He et~al.(2002)He, Xu and Deng}]{he2002squeezer}
\bibinfo{author}{He, Z.}, \bibinfo{author}{Xu, X.}, \bibinfo{author}{Deng, S.},
  \bibinfo{year}{2002}.
\newblock \bibinfo{title}{Squeezer: an efficient algorithm for clustering
  categorical data}.
\newblock \bibinfo{journal}{Journal of Computer Science and Technology}
  \bibinfo{volume}{17}, \bibinfo{pages}{611--624}.
%Type = Article
\bibitem[{He et~al.(2003)He, Xu and Deng}]{he2003discovering}
\bibinfo{author}{He, Z.}, \bibinfo{author}{Xu, X.}, \bibinfo{author}{Deng, S.},
  \bibinfo{year}{2003}.
\newblock \bibinfo{title}{Discovering cluster-based local outliers}.
\newblock \bibinfo{journal}{Pattern Recognition Letters} \bibinfo{volume}{24},
  \bibinfo{pages}{1641--1650}.
%Type = Article
\bibitem[{Hodge and Austin(2004)}]{hodge2004survey}
\bibinfo{author}{Hodge, V.}, \bibinfo{author}{Austin, J.},
  \bibinfo{year}{2004}.
\newblock \bibinfo{title}{A survey of outlier detection methodologies}.
\newblock \bibinfo{journal}{Artificial intelligence review}
  \bibinfo{volume}{22}, \bibinfo{pages}{85--126}.
%Type = Article
\bibitem[{Hou et~al.(2016)Hou, Gao and Li}]{hou2016dsets}
\bibinfo{author}{Hou, J.}, \bibinfo{author}{Gao, H.}, \bibinfo{author}{Li, X.},
  \bibinfo{year}{2016}.
\newblock \bibinfo{title}{Dsets-dbscan: A parameter-free clustering algorithm}.
\newblock \bibinfo{journal}{IEEE Transactions on Image Processing}
  \bibinfo{volume}{25}, \bibinfo{pages}{3182--3193}.
%Type = Article
\bibitem[{Hou and Liu(2017)}]{hou2017parameter}
\bibinfo{author}{Hou, J.}, \bibinfo{author}{Liu, W.}, \bibinfo{year}{2017}.
\newblock \bibinfo{title}{A parameter-independent clustering framework}.
\newblock \bibinfo{journal}{IEEE Transactions on Industrial Informatics}
  \bibinfo{volume}{13}, \bibinfo{pages}{1825--1832}.
%Type = Article
\bibitem[{Huang et~al.(2017)Huang, Zhu, Yang, Cheng and Wu}]{huang2017novel}
\bibinfo{author}{Huang, J.}, \bibinfo{author}{Zhu, Q.}, \bibinfo{author}{Yang,
  L.}, \bibinfo{author}{Cheng, D.}, \bibinfo{author}{Wu, Q.},
  \bibinfo{year}{2017}.
\newblock \bibinfo{title}{A novel outlier cluster detection algorithm without
  top-n parameter}.
\newblock \bibinfo{journal}{Knowledge-Based Systems} \bibinfo{volume}{121},
  \bibinfo{pages}{32--40}.
%Type = Article
\bibitem[{Huang et~al.(2016)Huang, Zhu, Yang and Feng}]{huang2016non}
\bibinfo{author}{Huang, J.}, \bibinfo{author}{Zhu, Q.}, \bibinfo{author}{Yang,
  L.}, \bibinfo{author}{Feng, J.}, \bibinfo{year}{2016}.
\newblock \bibinfo{title}{A non-parameter outlier detection algorithm based on
  natural neighbor}.
\newblock \bibinfo{journal}{Knowledge-Based Systems} \bibinfo{volume}{92},
  \bibinfo{pages}{71--77}.
%Type = Article
\bibitem[{Hubert and Debruyne(2010)}]{hubert2010minimum}
\bibinfo{author}{Hubert, M.}, \bibinfo{author}{Debruyne, M.},
  \bibinfo{year}{2010}.
\newblock \bibinfo{title}{Minimum covariance determinant}.
\newblock \bibinfo{journal}{Wiley interdisciplinary reviews: Computational
  statistics} \bibinfo{volume}{2}, \bibinfo{pages}{36--43}.
%Type = Article
\bibitem[{Hubert et~al.(2005)Hubert, Rousseeuw and
  Vanden~Branden}]{hubert2005robpca}
\bibinfo{author}{Hubert, M.}, \bibinfo{author}{Rousseeuw, P.J.},
  \bibinfo{author}{Vanden~Branden, K.}, \bibinfo{year}{2005}.
\newblock \bibinfo{title}{Robpca: a new approach to robust principal component
  analysis}.
\newblock \bibinfo{journal}{Technometrics} \bibinfo{volume}{47},
  \bibinfo{pages}{64--79}.
%Type = Inproceedings
\bibitem[{Januzaj et~al.(2004)Januzaj, Kriegel and
  Pfeifle}]{januzaj2004scalable}
\bibinfo{author}{Januzaj, E.}, \bibinfo{author}{Kriegel, H.P.},
  \bibinfo{author}{Pfeifle, M.}, \bibinfo{year}{2004}.
\newblock \bibinfo{title}{Scalable density-based distributed clustering}, in:
  \bibinfo{booktitle}{European Conference on Principles of Data Mining and
  Knowledge Discovery}, \bibinfo{organization}{Springer}. pp.
  \bibinfo{pages}{231--244}.
%Type = Inproceedings
\bibitem[{Jin et~al.(2006)Jin, Tung, Han and Wang}]{jin2006ranking}
\bibinfo{author}{Jin, W.}, \bibinfo{author}{Tung, A.K.}, \bibinfo{author}{Han,
  J.}, \bibinfo{author}{Wang, W.}, \bibinfo{year}{2006}.
\newblock \bibinfo{title}{Ranking outliers using symmetric neighborhood
  relationship}, in: \bibinfo{booktitle}{Pacific-Asia Conference on Knowledge
  Discovery and Data Mining}, \bibinfo{organization}{Springer}. pp.
  \bibinfo{pages}{577--593}.
%Type = Article
\bibitem[{Jobe and Pokojovy(2015)}]{jobe2015cluster}
\bibinfo{author}{Jobe, J.M.}, \bibinfo{author}{Pokojovy, M.},
  \bibinfo{year}{2015}.
\newblock \bibinfo{title}{A cluster-based outlier detection scheme for
  multivariate data}.
\newblock \bibinfo{journal}{Journal of the American Statistical Association}
  \bibinfo{volume}{110}, \bibinfo{pages}{1543--1551}.
%Type = Article
\bibitem[{Johnson and Lindenstrauss(1984)}]{johnson1984extensions}
\bibinfo{author}{Johnson, W.B.}, \bibinfo{author}{Lindenstrauss, J.},
  \bibinfo{year}{1984}.
\newblock \bibinfo{title}{Extensions of lipschitz mappings into a hilbert
  space}.
\newblock \bibinfo{journal}{Contemporary mathematics} \bibinfo{volume}{26},
  \bibinfo{pages}{1}.
%Type = Article
\bibitem[{Johnstone and Lu(2009)}]{johnstone2009sparse}
\bibinfo{author}{Johnstone, I.M.}, \bibinfo{author}{Lu, A.Y.},
  \bibinfo{year}{2009}.
\newblock \bibinfo{title}{Sparse principal components analysis}.
\newblock \bibinfo{journal}{arXiv preprint arXiv:0901.4392} .
%Type = Book
\bibitem[{Jolliffe(2011)}]{jolliffe2011principal}
\bibinfo{author}{Jolliffe, I.}, \bibinfo{year}{2011}.
\newblock \bibinfo{title}{Principal component analysis}.
\newblock \bibinfo{publisher}{Springer}.
%Type = Inproceedings
\bibitem[{Kennedy and Eberhart(1995)}]{kennedy1995particle}
\bibinfo{author}{Kennedy, J.}, \bibinfo{author}{Eberhart, R.},
  \bibinfo{year}{1995}.
\newblock \bibinfo{title}{Particle swarm optimization}, in:
  \bibinfo{booktitle}{Proceedings of ICNN'95-International Conference on Neural
  Networks}, \bibinfo{organization}{IEEE}. pp. \bibinfo{pages}{1942--1948}.
%Type = Inproceedings
\bibitem[{Knox and Ng(1998)}]{knox1998algorithms}
\bibinfo{author}{Knox, E.M.}, \bibinfo{author}{Ng, R.T.}, \bibinfo{year}{1998}.
\newblock \bibinfo{title}{Algorithms for mining distancebased outliers in large
  datasets}, in: \bibinfo{booktitle}{Proceedings of the international
  conference on very large data bases}, \bibinfo{organization}{Citeseer}. pp.
  \bibinfo{pages}{392--403}.
%Type = Article
\bibitem[{Kollios et~al.(2003)Kollios, Gunopulos, Koudas and
  Berchtold}]{kollios2003efficient}
\bibinfo{author}{Kollios, G.}, \bibinfo{author}{Gunopulos, D.},
  \bibinfo{author}{Koudas, N.}, \bibinfo{author}{Berchtold, S.},
  \bibinfo{year}{2003}.
\newblock \bibinfo{title}{Efficient biased sampling for approximate clustering
  and outlier detection in large data sets}.
\newblock \bibinfo{journal}{IEEE Transactions on knowledge and data
  engineering} \bibinfo{volume}{15}, \bibinfo{pages}{1170--1187}.
%Type = Inproceedings
\bibitem[{Kriegel et~al.(2009)Kriegel, Kr{\"o}ger, Schubert and
  Zimek}]{kriegel2009loop}
\bibinfo{author}{Kriegel, H.P.}, \bibinfo{author}{Kr{\"o}ger, P.},
  \bibinfo{author}{Schubert, E.}, \bibinfo{author}{Zimek, A.},
  \bibinfo{year}{2009}.
\newblock \bibinfo{title}{Loop: local outlier probabilities}, in:
  \bibinfo{booktitle}{Proceedings of the 18th ACM conference on Information and
  knowledge management}, \bibinfo{organization}{ACM}. pp.
  \bibinfo{pages}{1649--1652}.
%Type = Article
\bibitem[{Ledoit and Wolf(2004)}]{ledoit2004honey}
\bibinfo{author}{Ledoit, O.}, \bibinfo{author}{Wolf, M.}, \bibinfo{year}{2004}.
\newblock \bibinfo{title}{Honey, i shrunk the sample covariance matrix}.
\newblock \bibinfo{journal}{The Journal of Portfolio Management}
  \bibinfo{volume}{30}, \bibinfo{pages}{110--119}.
%Type = Book
\bibitem[{Leskovec et~al.(2014)Leskovec, Rajaraman and
  Ullman}]{leskovec2014mining}
\bibinfo{author}{Leskovec, J.}, \bibinfo{author}{Rajaraman, A.},
  \bibinfo{author}{Ullman, J.D.}, \bibinfo{year}{2014}.
\newblock \bibinfo{title}{Mining of massive datasets}.
\newblock \bibinfo{publisher}{Cambridge university press}.
%Type = Inproceedings
\bibitem[{Liu et~al.(2008)Liu, Ting and Zhou}]{liu2008isolation}
\bibinfo{author}{Liu, F.T.}, \bibinfo{author}{Ting, K.M.},
  \bibinfo{author}{Zhou, Z.H.}, \bibinfo{year}{2008}.
\newblock \bibinfo{title}{Isolation forest}, in: \bibinfo{booktitle}{2008
  Eighth IEEE International Conference on Data Mining},
  \bibinfo{organization}{IEEE}. pp. \bibinfo{pages}{413--422}.
%Type = Article
\bibitem[{Liu et~al.(2012)Liu, Ting and Zhou}]{liu2012isolation}
\bibinfo{author}{Liu, F.T.}, \bibinfo{author}{Ting, K.M.},
  \bibinfo{author}{Zhou, Z.H.}, \bibinfo{year}{2012}.
\newblock \bibinfo{title}{Isolation-based anomaly detection}.
\newblock \bibinfo{journal}{ACM Transactions on Knowledge Discovery from Data
  (TKDD)} \bibinfo{volume}{6}, \bibinfo{pages}{1--39}.
%Type = Article
\bibitem[{Liu et~al.(2019a)Liu, Huang, Fei, Wang and Liang}]{liu2019constraint}
\bibinfo{author}{Liu, R.}, \bibinfo{author}{Huang, W.}, \bibinfo{author}{Fei,
  Z.}, \bibinfo{author}{Wang, K.}, \bibinfo{author}{Liang, J.},
  \bibinfo{year}{2019}a.
\newblock \bibinfo{title}{Constraint-based clustering by fast search and find
  of density peaks}.
\newblock \bibinfo{journal}{Neurocomputing} \bibinfo{volume}{330},
  \bibinfo{pages}{223--237}.
%Type = Article
\bibitem[{Liu et~al.(2018)Liu, Wang and Yu}]{liu2018shared}
\bibinfo{author}{Liu, R.}, \bibinfo{author}{Wang, H.}, \bibinfo{author}{Yu,
  X.}, \bibinfo{year}{2018}.
\newblock \bibinfo{title}{Shared-nearest-neighbor-based clustering by fast
  search and find of density peaks}.
\newblock \bibinfo{journal}{Information Sciences} \bibinfo{volume}{450},
  \bibinfo{pages}{200--226}.
%Type = Article
\bibitem[{Liu et~al.(2019b)Liu, Li and Zhao}]{liu2019clustering}
\bibinfo{author}{Liu, T.}, \bibinfo{author}{Li, H.}, \bibinfo{author}{Zhao,
  X.}, \bibinfo{year}{2019}b.
\newblock \bibinfo{title}{Clustering by search in descending order and
  automatic find of density peaks}.
\newblock \bibinfo{journal}{IEEE Access} \bibinfo{volume}{7},
  \bibinfo{pages}{133772--133780}.
%Type = Article
\bibitem[{Lotfi et~al.(2020)Lotfi, Moradi and Beigy}]{lotfi2020density}
\bibinfo{author}{Lotfi, A.}, \bibinfo{author}{Moradi, P.},
  \bibinfo{author}{Beigy, H.}, \bibinfo{year}{2020}.
\newblock \bibinfo{title}{Density peaks clustering based on density backbone
  and fuzzy neighborhood}.
\newblock \bibinfo{journal}{Pattern Recognition} \bibinfo{volume}{107},
  \bibinfo{pages}{107449}.
%Type = Inproceedings
\bibitem[{Mahalanobis(1936)}]{mahalanobis1936generalized}
\bibinfo{author}{Mahalanobis, P.C.}, \bibinfo{year}{1936}.
\newblock \bibinfo{title}{On the generalized distance in statistics},
  \bibinfo{organization}{National Institute of Science of India}.
%Type = Inproceedings
\bibitem[{Mao et~al.(2018)Mao, Sun, Jin and Zhou}]{mao2018outlier}
\bibinfo{author}{Mao, J.}, \bibinfo{author}{Sun, P.}, \bibinfo{author}{Jin,
  C.}, \bibinfo{author}{Zhou, A.}, \bibinfo{year}{2018}.
\newblock \bibinfo{title}{Outlier detection over distributed trajectory
  streams}, in: \bibinfo{booktitle}{Proceedings of the 2018 SIAM international
  conference on data mining}, \bibinfo{organization}{SIAM}. pp.
  \bibinfo{pages}{64--72}.
%Type = Article
\bibitem[{Maronna and Zamar(2002)}]{maronna2002robust}
\bibinfo{author}{Maronna, R.A.}, \bibinfo{author}{Zamar, R.H.},
  \bibinfo{year}{2002}.
\newblock \bibinfo{title}{Robust estimates of location and dispersion for
  high-dimensional datasets}.
\newblock \bibinfo{journal}{Technometrics} \bibinfo{volume}{44},
  \bibinfo{pages}{307--317}.
%Type = Article
\bibitem[{Maulik and Bandyopadhyay(2002)}]{maulik2002performance}
\bibinfo{author}{Maulik, U.}, \bibinfo{author}{Bandyopadhyay, S.},
  \bibinfo{year}{2002}.
\newblock \bibinfo{title}{Performance evaluation of some clustering algorithms
  and validity indices}.
\newblock \bibinfo{journal}{IEEE Transactions on pattern analysis and machine
  intelligence} \bibinfo{volume}{24}, \bibinfo{pages}{1650--1654}.
%Type = Article
\bibitem[{Mehmood et~al.(2016)Mehmood, Zhang, Bie, Dawood and
  Ahmad}]{mehmood2016clustering}
\bibinfo{author}{Mehmood, R.}, \bibinfo{author}{Zhang, G.},
  \bibinfo{author}{Bie, R.}, \bibinfo{author}{Dawood, H.},
  \bibinfo{author}{Ahmad, H.}, \bibinfo{year}{2016}.
\newblock \bibinfo{title}{Clustering by fast search and find of density peaks
  via heat diffusion}.
\newblock \bibinfo{journal}{Neurocomputing} \bibinfo{volume}{208},
  \bibinfo{pages}{210--217}.
%Type = Incollection
\bibitem[{Meil{\u{a}}(2003)}]{meilua2003comparing}
\bibinfo{author}{Meil{\u{a}}, M.}, \bibinfo{year}{2003}.
\newblock \bibinfo{title}{Comparing clusterings by the variation of
  information}, in: \bibinfo{booktitle}{Learning theory and kernel machines}.
  \bibinfo{publisher}{Springer}, pp. \bibinfo{pages}{173--187}.
%Type = Book
\bibitem[{Mirkin(1996)}]{mirkin1996mathematical}
\bibinfo{author}{Mirkin, B.}, \bibinfo{year}{1996}.
\newblock \bibinfo{title}{Mathematical classification and clustering}.
  volume~\bibinfo{volume}{11}.
\newblock \bibinfo{publisher}{Springer Science \& Business Media}.
%Type = Inproceedings
\bibitem[{Moonesignhe and Tan(2006)}]{moonesignhe2006outlier}
\bibinfo{author}{Moonesignhe, H.}, \bibinfo{author}{Tan, P.N.},
  \bibinfo{year}{2006}.
\newblock \bibinfo{title}{Outlier detection using random walks}, in:
  \bibinfo{booktitle}{2006 18th IEEE International Conference on Tools with
  Artificial Intelligence (ICTAI'06)}, \bibinfo{organization}{IEEE}. pp.
  \bibinfo{pages}{532--539}.
%Type = Article
\bibitem[{Moonesinghe and Tan(2008)}]{moonesinghe2008outrank}
\bibinfo{author}{Moonesinghe, H.}, \bibinfo{author}{Tan, P.N.},
  \bibinfo{year}{2008}.
\newblock \bibinfo{title}{Outrank: a graph-based outlier detection framework
  using random walk}.
\newblock \bibinfo{journal}{International Journal on Artificial Intelligence
  Tools} \bibinfo{volume}{17}, \bibinfo{pages}{19--36}.
%Type = Book
\bibitem[{Newton(1802)}]{newton1802mathematical}
\bibinfo{author}{Newton, I.}, \bibinfo{year}{1802}.
\newblock \bibinfo{title}{Mathematical principles of natural philosophy}.
\newblock \bibinfo{publisher}{A. Strahan}.
%Type = Article
\bibitem[{Nickabadi et~al.(2011)Nickabadi, Ebadzadeh and
  Safabakhsh}]{nickabadi2011novel}
\bibinfo{author}{Nickabadi, A.}, \bibinfo{author}{Ebadzadeh, M.M.},
  \bibinfo{author}{Safabakhsh, R.}, \bibinfo{year}{2011}.
\newblock \bibinfo{title}{A novel particle swarm optimization algorithm with
  adaptive inertia weight}.
\newblock \bibinfo{journal}{Applied soft computing} \bibinfo{volume}{11},
  \bibinfo{pages}{3658--3670}.
%Type = Inproceedings
\bibitem[{Palmer and Faloutsos(2000)}]{palmer2000density}
\bibinfo{author}{Palmer, C.R.}, \bibinfo{author}{Faloutsos, C.},
  \bibinfo{year}{2000}.
\newblock \bibinfo{title}{Density biased sampling: An improved method for data
  mining and clustering}, in: \bibinfo{booktitle}{Proceedings of the 2000 ACM
  SIGMOD international conference on Management of data}, pp.
  \bibinfo{pages}{82--92}.
%Type = Article
\bibitem[{Pavan and Pelillo(2006)}]{pavan2006dominant}
\bibinfo{author}{Pavan, M.}, \bibinfo{author}{Pelillo, M.},
  \bibinfo{year}{2006}.
\newblock \bibinfo{title}{Dominant sets and pairwise clustering}.
\newblock \bibinfo{journal}{IEEE transactions on pattern analysis and machine
  intelligence} \bibinfo{volume}{29}, \bibinfo{pages}{167--172}.
%Type = Article
\bibitem[{Pearson(1901)}]{pearson1901liii}
\bibinfo{author}{Pearson, K.}, \bibinfo{year}{1901}.
\newblock \bibinfo{title}{Liii. on lines and planes of closest fit to systems
  of points in space}.
\newblock \bibinfo{journal}{The London, Edinburgh, and Dublin Philosophical
  Magazine and Journal of Science} \bibinfo{volume}{2},
  \bibinfo{pages}{559--572}.
%Type = Inproceedings
\bibitem[{Pelleg et~al.(2000)Pelleg, Moore et~al.}]{pelleg2000x}
\bibinfo{author}{Pelleg, D.}, \bibinfo{author}{Moore, A.W.}, et~al.,
  \bibinfo{year}{2000}.
\newblock \bibinfo{title}{X-means: Extending k-means with efficient estimation
  of the number of clusters.}, in: \bibinfo{booktitle}{Icml}, pp.
  \bibinfo{pages}{727--734}.
%Type = Article
\bibitem[{Rahman et~al.(2018)Rahman, Ang and Seng}]{rahman2018unique}
\bibinfo{author}{Rahman, M.A.}, \bibinfo{author}{Ang, K.L.M.},
  \bibinfo{author}{Seng, K.P.}, \bibinfo{year}{2018}.
\newblock \bibinfo{title}{Unique neighborhood set parameter independent
  density-based clustering with outlier detection}.
\newblock \bibinfo{journal}{IEEE Access} \bibinfo{volume}{6},
  \bibinfo{pages}{44707--44717}.
%Type = Article
\bibitem[{Rahman et~al.(2020)Rahman, Ang and Seng}]{rahman2020clustering}
\bibinfo{author}{Rahman, M.A.}, \bibinfo{author}{Ang, L.M.},
  \bibinfo{author}{Seng, K.P.}, \bibinfo{year}{2020}.
\newblock \bibinfo{title}{Clustering biomedical and gene expression datasets
  with kernel density and unique neighborhood set based vein detection}.
\newblock \bibinfo{journal}{Information Systems} \bibinfo{volume}{91},
  \bibinfo{pages}{101490}.
%Type = Inproceedings
\bibitem[{Ramaswamy et~al.(2000)Ramaswamy, Rastogi and
  Shim}]{ramaswamy2000efficient}
\bibinfo{author}{Ramaswamy, S.}, \bibinfo{author}{Rastogi, R.},
  \bibinfo{author}{Shim, K.}, \bibinfo{year}{2000}.
\newblock \bibinfo{title}{Efficient algorithms for mining outliers from large
  data sets}, in: \bibinfo{booktitle}{Proceedings of the 2000 ACM SIGMOD
  international conference on Management of data}, pp.
  \bibinfo{pages}{427--438}.
%Type = Article
\bibitem[{Ranshous et~al.(2015)Ranshous, Shen, Koutra, Harenberg, Faloutsos and
  Samatova}]{ranshous2015anomaly}
\bibinfo{author}{Ranshous, S.}, \bibinfo{author}{Shen, S.},
  \bibinfo{author}{Koutra, D.}, \bibinfo{author}{Harenberg, S.},
  \bibinfo{author}{Faloutsos, C.}, \bibinfo{author}{Samatova, N.F.},
  \bibinfo{year}{2015}.
\newblock \bibinfo{title}{Anomaly detection in dynamic networks: a survey}.
\newblock \bibinfo{journal}{Wiley Interdisciplinary Reviews: Computational
  Statistics} \bibinfo{volume}{7}, \bibinfo{pages}{223--247}.
%Type = Misc
\bibitem[{Rayana(2016)}]{Rayana:2016}
\bibinfo{author}{Rayana, S.}, \bibinfo{year}{2016}.
\newblock \bibinfo{title}{{ODDS} library}.
\newblock \URLprefix \url{http://odds.cs.stonybrook.edu}.
%Type = Article
\bibitem[{Ro et~al.(2015)Ro, Zou, Wang and Yin}]{ro2015outlier}
\bibinfo{author}{Ro, K.}, \bibinfo{author}{Zou, C.}, \bibinfo{author}{Wang,
  Z.}, \bibinfo{author}{Yin, G.}, \bibinfo{year}{2015}.
\newblock \bibinfo{title}{Outlier detection for high-dimensional data}.
\newblock \bibinfo{journal}{Biometrika} \bibinfo{volume}{102},
  \bibinfo{pages}{589--599}.
%Type = Article
\bibitem[{Rodriguez and Laio(2014)}]{rodriguez2014clustering}
\bibinfo{author}{Rodriguez, A.}, \bibinfo{author}{Laio, A.},
  \bibinfo{year}{2014}.
\newblock \bibinfo{title}{Clustering by fast search and find of density peaks}.
\newblock \bibinfo{journal}{science} \bibinfo{volume}{344},
  \bibinfo{pages}{1492--1496}.
%Type = Article
\bibitem[{Rousseeuw(1987)}]{rousseeuw1987silhouettes}
\bibinfo{author}{Rousseeuw, P.J.}, \bibinfo{year}{1987}.
\newblock \bibinfo{title}{Silhouettes: a graphical aid to the interpretation
  and validation of cluster analysis}.
\newblock \bibinfo{journal}{Journal of computational and applied mathematics}
  \bibinfo{volume}{20}, \bibinfo{pages}{53--65}.
%Type = Article
\bibitem[{Rousseeuw and Driessen(1999)}]{rousseeuw1999fast}
\bibinfo{author}{Rousseeuw, P.J.}, \bibinfo{author}{Driessen, K.V.},
  \bibinfo{year}{1999}.
\newblock \bibinfo{title}{A fast algorithm for the minimum covariance
  determinant estimator}.
\newblock \bibinfo{journal}{Technometrics} \bibinfo{volume}{41},
  \bibinfo{pages}{212--223}.
%Type = Article
\bibitem[{Rubinov et~al.(2006)Rubinov, Soukhorokova and
  Ugon}]{rubinov2006classes}
\bibinfo{author}{Rubinov, A.M.}, \bibinfo{author}{Soukhorokova, N.},
  \bibinfo{author}{Ugon, J.}, \bibinfo{year}{2006}.
\newblock \bibinfo{title}{Classes and clusters in data analysis}.
\newblock \bibinfo{journal}{European Journal of Operational Research}
  \bibinfo{volume}{173}, \bibinfo{pages}{849--865}.
%Type = Inproceedings
\bibitem[{Saitta et~al.(2007)Saitta, Raphael and Smith}]{saitta2007bounded}
\bibinfo{author}{Saitta, S.}, \bibinfo{author}{Raphael, B.},
  \bibinfo{author}{Smith, I.F.}, \bibinfo{year}{2007}.
\newblock \bibinfo{title}{A bounded index for cluster validity}, in:
  \bibinfo{booktitle}{International workshop on machine learning and data
  mining in pattern recognition}, \bibinfo{organization}{Springer}. pp.
  \bibinfo{pages}{174--187}.
%Type = Article
\bibitem[{Sander et~al.(1998)Sander, Ester, Kriegel and Xu}]{sander1998density}
\bibinfo{author}{Sander, J.}, \bibinfo{author}{Ester, M.},
  \bibinfo{author}{Kriegel, H.P.}, \bibinfo{author}{Xu, X.},
  \bibinfo{year}{1998}.
\newblock \bibinfo{title}{Density-based clustering in spatial databases: The
  algorithm gdbscan and its applications}.
\newblock \bibinfo{journal}{Data mining and knowledge discovery}
  \bibinfo{volume}{2}, \bibinfo{pages}{169--194}.
%Type = Article
\bibitem[{Sch{\"o}lkopf et~al.(2001)Sch{\"o}lkopf, Platt, Shawe-Taylor, Smola
  and Williamson}]{scholkopf2001estimating}
\bibinfo{author}{Sch{\"o}lkopf, B.}, \bibinfo{author}{Platt, J.C.},
  \bibinfo{author}{Shawe-Taylor, J.}, \bibinfo{author}{Smola, A.J.},
  \bibinfo{author}{Williamson, R.C.}, \bibinfo{year}{2001}.
\newblock \bibinfo{title}{Estimating the support of a high-dimensional
  distribution}.
\newblock \bibinfo{journal}{Neural computation} \bibinfo{volume}{13},
  \bibinfo{pages}{1443--1471}.
%Type = Article
\bibitem[{Schubert et~al.(2017)Schubert, Sander, Ester, Kriegel and
  Xu}]{schubert2017dbscan}
\bibinfo{author}{Schubert, E.}, \bibinfo{author}{Sander, J.},
  \bibinfo{author}{Ester, M.}, \bibinfo{author}{Kriegel, H.P.},
  \bibinfo{author}{Xu, X.}, \bibinfo{year}{2017}.
\newblock \bibinfo{title}{Dbscan revisited, revisited: why and how you should
  (still) use dbscan}.
\newblock \bibinfo{journal}{ACM Transactions on Database Systems (TODS)}
  \bibinfo{volume}{42}, \bibinfo{pages}{1--21}.
%Type = Article
\bibitem[{Shlens(2014)}]{shlens2014tutorial}
\bibinfo{author}{Shlens, J.}, \bibinfo{year}{2014}.
\newblock \bibinfo{title}{A tutorial on principal component analysis}.
\newblock \bibinfo{journal}{arXiv preprint arXiv:1404.1100} .
%Type = Inproceedings
\bibitem[{Sugiyama and Borgwardt(2013)}]{sugiyama2013rapid}
\bibinfo{author}{Sugiyama, M.}, \bibinfo{author}{Borgwardt, K.},
  \bibinfo{year}{2013}.
\newblock \bibinfo{title}{Rapid distance-based outlier detection via sampling},
  in: \bibinfo{booktitle}{Advances in Neural Information Processing Systems},
  pp. \bibinfo{pages}{467--475}.
%Type = Article
\bibitem[{Tang and He(2017)}]{tang2017local}
\bibinfo{author}{Tang, B.}, \bibinfo{author}{He, H.}, \bibinfo{year}{2017}.
\newblock \bibinfo{title}{A local density-based approach for outlier
  detection}.
\newblock \bibinfo{journal}{Neurocomputing} \bibinfo{volume}{241},
  \bibinfo{pages}{171--180}.
%Type = Article
\bibitem[{Tax and Duin(1999)}]{tax1999support}
\bibinfo{author}{Tax, D.M.}, \bibinfo{author}{Duin, R.P.},
  \bibinfo{year}{1999}.
\newblock \bibinfo{title}{Support vector domain description}.
\newblock \bibinfo{journal}{Pattern recognition letters} \bibinfo{volume}{20},
  \bibinfo{pages}{1191--1199}.
%Type = Article
\bibitem[{Tenenbaum et~al.(2000)Tenenbaum, De~Silva and
  Langford}]{tenenbaum2000global}
\bibinfo{author}{Tenenbaum, J.B.}, \bibinfo{author}{De~Silva, V.},
  \bibinfo{author}{Langford, J.C.}, \bibinfo{year}{2000}.
\newblock \bibinfo{title}{A global geometric framework for nonlinear
  dimensionality reduction}.
\newblock \bibinfo{journal}{science} \bibinfo{volume}{290},
  \bibinfo{pages}{2319--2323}.
%Type = Article
\bibitem[{Teng et~al.(2016)}]{teng2016scalable}
\bibinfo{author}{Teng, S.H.}, et~al., \bibinfo{year}{2016}.
\newblock \bibinfo{title}{Scalable algorithms for data and network analysis}.
\newblock \bibinfo{journal}{Foundations and Trends{\textregistered} in
  Theoretical Computer Science} \bibinfo{volume}{12}, \bibinfo{pages}{1--274}.
%Type = Book
\bibitem[{Thompson(1992)}]{thompson1992sampling}
\bibinfo{author}{Thompson, S.}, \bibinfo{year}{1992}.
\newblock \bibinfo{title}{Sampling, A Wiley-Inter-science publication}.
\newblock \bibinfo{publisher}{John Wiley \& Sons, Inc., New York}.
%Type = Book
\bibitem[{Van~Rijsbergen(2004)}]{van2004geometry}
\bibinfo{author}{Van~Rijsbergen, C.J.}, \bibinfo{year}{2004}.
\newblock \bibinfo{title}{The geometry of information retrieval}.
\newblock \bibinfo{publisher}{Cambridge University Press}.
%Type = Article
\bibitem[{Vanschoren et~al.(2014)Vanschoren, Van~Rijn, Bischl and
  Torgo}]{vanschoren2014openml}
\bibinfo{author}{Vanschoren, J.}, \bibinfo{author}{Van~Rijn, J.N.},
  \bibinfo{author}{Bischl, B.}, \bibinfo{author}{Torgo, L.},
  \bibinfo{year}{2014}.
\newblock \bibinfo{title}{Openml: networked science in machine learning}.
\newblock \bibinfo{journal}{ACM SIGKDD Explorations Newsletter}
  \bibinfo{volume}{15}, \bibinfo{pages}{49--60}.
%Type = Article
\bibitem[{Wahid and Annavarapu(2020)}]{wahid2020nanod}
\bibinfo{author}{Wahid, A.}, \bibinfo{author}{Annavarapu, C.S.R.},
  \bibinfo{year}{2020}.
\newblock \bibinfo{title}{Nanod: A natural neighbour-based outlier detection
  algorithm}.
\newblock \bibinfo{journal}{Neural Computing and Applications} ,
  \bibinfo{pages}{1--17}.
%Type = Article
\bibitem[{Wahid and Rao(2019)}]{wahid2019rkdos}
\bibinfo{author}{Wahid, A.}, \bibinfo{author}{Rao, A.C.S.},
  \bibinfo{year}{2019}.
\newblock \bibinfo{title}{Rkdos: A relative kernel density-based outlier
  score}.
\newblock \bibinfo{journal}{IETE Technical Review} , \bibinfo{pages}{1--12}.
%Type = Article
\bibitem[{Wahid and Rao(2020)}]{wahid2020odra}
\bibinfo{author}{Wahid, A.}, \bibinfo{author}{Rao, A.C.S.},
  \bibinfo{year}{2020}.
\newblock \bibinfo{title}{Odra: an outlier detection algorithm based on
  relevant attribute analysis method}.
\newblock \bibinfo{journal}{Cluster Computing} , \bibinfo{pages}{1--17}.
%Type = Article
\bibitem[{Wang et~al.(2018a)Wang, Gao, Liu and Fu}]{wang2018new}
\bibinfo{author}{Wang, C.}, \bibinfo{author}{Gao, H.}, \bibinfo{author}{Liu,
  Z.}, \bibinfo{author}{Fu, Y.}, \bibinfo{year}{2018}a.
\newblock \bibinfo{title}{A new outlier detection model using random walk on
  local information graph}.
\newblock \bibinfo{journal}{IEEE Access} \bibinfo{volume}{6},
  \bibinfo{pages}{75531--75544}.
%Type = Inproceedings
\bibitem[{Wang et~al.(2018b)Wang, Gao, Liu and Fu}]{wang2018outlier}
\bibinfo{author}{Wang, C.}, \bibinfo{author}{Gao, H.}, \bibinfo{author}{Liu,
  Z.}, \bibinfo{author}{Fu, Y.}, \bibinfo{year}{2018}b.
\newblock \bibinfo{title}{Outlier detection using diverse neighborhood graphs},
  in: \bibinfo{booktitle}{2018 15th International Computer Conference on
  Wavelet Active Media Technology and Information Processing (ICCWAMTIP)},
  \bibinfo{organization}{IEEE}. pp. \bibinfo{pages}{58--62}.
%Type = Article
\bibitem[{Wang et~al.(2019a)Wang, Liu, Gao and Fu}]{wang2019vos}
\bibinfo{author}{Wang, C.}, \bibinfo{author}{Liu, Z.}, \bibinfo{author}{Gao,
  H.}, \bibinfo{author}{Fu, Y.}, \bibinfo{year}{2019}a.
\newblock \bibinfo{title}{Vos: A new outlier detection model using virtual
  graph}.
\newblock \bibinfo{journal}{Knowledge-Based Systems} \bibinfo{volume}{185},
  \bibinfo{pages}{104907}.
%Type = Article
\bibitem[{Wang et~al.(2019b)Wang, Bah and Hammad}]{wang2019progress}
\bibinfo{author}{Wang, H.}, \bibinfo{author}{Bah, M.J.},
  \bibinfo{author}{Hammad, M.}, \bibinfo{year}{2019}b.
\newblock \bibinfo{title}{Progress in outlier detection techniques: A survey}.
\newblock \bibinfo{journal}{IEEE Access} \bibinfo{volume}{7},
  \bibinfo{pages}{107964--108000}.
%Type = Book
\bibitem[{Wang et~al.(2021)Wang, Wang and Wilkes}]{wangnew}
\bibinfo{author}{Wang, X.}, \bibinfo{author}{Wang, X.},
  \bibinfo{author}{Wilkes, M.}, \bibinfo{year}{2021}.
\newblock \bibinfo{title}{New Developments in Unsupervised Outlier Detection}.
\newblock \bibinfo{publisher}{Springer Singapore}.
%Type = Inproceedings
\bibitem[{Wu and Jermaine(2006)}]{wu2006outlier}
\bibinfo{author}{Wu, M.}, \bibinfo{author}{Jermaine, C.}, \bibinfo{year}{2006}.
\newblock \bibinfo{title}{Outlier detection by sampling with accuracy
  guarantees}, in: \bibinfo{booktitle}{Proceedings of the 12th ACM SIGKDD
  international conference on Knowledge discovery and data mining}, pp.
  \bibinfo{pages}{767--772}.
%Type = Article
\bibitem[{Wu and Wang(2011)}]{wu2011information}
\bibinfo{author}{Wu, S.}, \bibinfo{author}{Wang, S.}, \bibinfo{year}{2011}.
\newblock \bibinfo{title}{Information-theoretic outlier detection for
  large-scale categorical data}.
\newblock \bibinfo{journal}{IEEE transactions on knowledge and data
  engineering} \bibinfo{volume}{25}, \bibinfo{pages}{589--602}.
%Type = Article
\bibitem[{Xie et~al.(2016)Xie, Gao, Xie, Liu and Grant}]{xie2016robust}
\bibinfo{author}{Xie, J.}, \bibinfo{author}{Gao, H.}, \bibinfo{author}{Xie,
  W.}, \bibinfo{author}{Liu, X.}, \bibinfo{author}{Grant, P.W.},
  \bibinfo{year}{2016}.
\newblock \bibinfo{title}{Robust clustering by detecting density peaks and
  assigning points based on fuzzy weighted k-nearest neighbors}.
\newblock \bibinfo{journal}{Information Sciences} \bibinfo{volume}{354},
  \bibinfo{pages}{19--40}.
%Type = Article
\bibitem[{Xie et~al.(2020)Xie, Xiong, Dai, Wang and Zhang}]{xie2020local}
\bibinfo{author}{Xie, J.}, \bibinfo{author}{Xiong, Z.}, \bibinfo{author}{Dai,
  Q.}, \bibinfo{author}{Wang, X.}, \bibinfo{author}{Zhang, Y.},
  \bibinfo{year}{2020}.
\newblock \bibinfo{title}{A local-gravitation-based method for the detection of
  outliers and boundary points}.
\newblock \bibinfo{journal}{Knowledge-Based Systems} \bibinfo{volume}{192},
  \bibinfo{pages}{105331}.
%Type = Inproceedings
\bibitem[{Yan et~al.(2017a)Yan, Cao, Kulhman and
  Rundensteiner}]{yan2017distributed1}
\bibinfo{author}{Yan, Y.}, \bibinfo{author}{Cao, L.}, \bibinfo{author}{Kulhman,
  C.}, \bibinfo{author}{Rundensteiner, E.}, \bibinfo{year}{2017}a.
\newblock \bibinfo{title}{Distributed local outlier detection in big data}, in:
  \bibinfo{booktitle}{Proceedings of the 23rd ACM SIGKDD international
  conference on knowledge discovery and data mining}, pp.
  \bibinfo{pages}{1225--1234}.
%Type = Inproceedings
\bibitem[{Yan et~al.(2017b)Yan, Cao and Rundensteiner}]{yan2017distributed2}
\bibinfo{author}{Yan, Y.}, \bibinfo{author}{Cao, L.},
  \bibinfo{author}{Rundensteiner, E.A.}, \bibinfo{year}{2017}b.
\newblock \bibinfo{title}{Distributed top-n local outlier detection in big
  data}, in: \bibinfo{booktitle}{2017 IEEE International Conference on Big Data
  (Big Data)}, \bibinfo{organization}{IEEE}. pp. \bibinfo{pages}{827--836}.
%Type = Inproceedings
\bibitem[{Yin et~al.(2013)Yin, Ho and Xing}]{yin2013scalable}
\bibinfo{author}{Yin, J.}, \bibinfo{author}{Ho, Q.}, \bibinfo{author}{Xing,
  E.P.}, \bibinfo{year}{2013}.
\newblock \bibinfo{title}{A scalable approach to probabilistic latent space
  inference of large-scale networks}, in: \bibinfo{booktitle}{Advances in
  neural information processing systems}, pp. \bibinfo{pages}{422--430}.
%Type = Article
\bibitem[{Yu et~al.(2016)Yu, Qiu, Wen, Lin and Liu}]{yu2016survey}
\bibinfo{author}{Yu, R.}, \bibinfo{author}{Qiu, H.}, \bibinfo{author}{Wen, Z.},
  \bibinfo{author}{Lin, C.}, \bibinfo{author}{Liu, Y.}, \bibinfo{year}{2016}.
\newblock \bibinfo{title}{A survey on social media anomaly detection}.
\newblock \bibinfo{journal}{ACM SIGKDD Explorations Newsletter}
  \bibinfo{volume}{18}, \bibinfo{pages}{1--14}.
%Type = Article
\bibitem[{Zeng et~al.(2012)Zeng, Li, Duan, Lu, Shi, Wang, Wu and
  Luo}]{zeng2012distributed}
\bibinfo{author}{Zeng, L.}, \bibinfo{author}{Li, L.}, \bibinfo{author}{Duan,
  L.}, \bibinfo{author}{Lu, K.}, \bibinfo{author}{Shi, Z.},
  \bibinfo{author}{Wang, M.}, \bibinfo{author}{Wu, W.}, \bibinfo{author}{Luo,
  P.}, \bibinfo{year}{2012}.
\newblock \bibinfo{title}{Distributed data mining: a survey}.
\newblock \bibinfo{journal}{Information Technology and Management}
  \bibinfo{volume}{13}, \bibinfo{pages}{403--409}.
%Type = Inproceedings
\bibitem[{Zhang et~al.(2009)Zhang, Hutter and Jin}]{zhang2009new}
\bibinfo{author}{Zhang, K.}, \bibinfo{author}{Hutter, M.},
  \bibinfo{author}{Jin, H.}, \bibinfo{year}{2009}.
\newblock \bibinfo{title}{A new local distance-based outlier detection approach
  for scattered real-world data}, in: \bibinfo{booktitle}{Pacific-Asia
  Conference on Knowledge Discovery and Data Mining},
  \bibinfo{organization}{Springer}. pp. \bibinfo{pages}{813--822}.
%Type = Article
\bibitem[{Zhou et~al.(2018)Zhou, Si, Zhang and Zheng}]{zhou2018robust}
\bibinfo{author}{Zhou, Z.}, \bibinfo{author}{Si, G.}, \bibinfo{author}{Zhang,
  Y.}, \bibinfo{author}{Zheng, K.}, \bibinfo{year}{2018}.
\newblock \bibinfo{title}{Robust clustering by identifying the veins of
  clusters based on kernel density estimation}.
\newblock \bibinfo{journal}{Knowledge-Based Systems} \bibinfo{volume}{159},
  \bibinfo{pages}{309--320}.
%Type = Inproceedings
\bibitem[{Zimek et~al.(2013)Zimek, Gaudet, Campello and
  Sander}]{zimek2013subsampling}
\bibinfo{author}{Zimek, A.}, \bibinfo{author}{Gaudet, M.},
  \bibinfo{author}{Campello, R.J.}, \bibinfo{author}{Sander, J.},
  \bibinfo{year}{2013}.
\newblock \bibinfo{title}{Subsampling for efficient and effective unsupervised
  outlier detection ensembles}, in: \bibinfo{booktitle}{Proceedings of the 19th
  ACM SIGKDD international conference on Knowledge discovery and data mining},
  pp. \bibinfo{pages}{428--436}.
%Type = Article
\bibitem[{Zimek et~al.(2012)Zimek, Schubert and Kriegel}]{zimek2012survey}
\bibinfo{author}{Zimek, A.}, \bibinfo{author}{Schubert, E.},
  \bibinfo{author}{Kriegel, H.P.}, \bibinfo{year}{2012}.
\newblock \bibinfo{title}{A survey on unsupervised outlier detection in
  high-dimensional numerical data}.
\newblock \bibinfo{journal}{Statistical Analysis and Data Mining: The ASA Data
  Science Journal} \bibinfo{volume}{5}, \bibinfo{pages}{363--387}.

\end{thebibliography}

\vskip3pt

%\clearpage

\bio{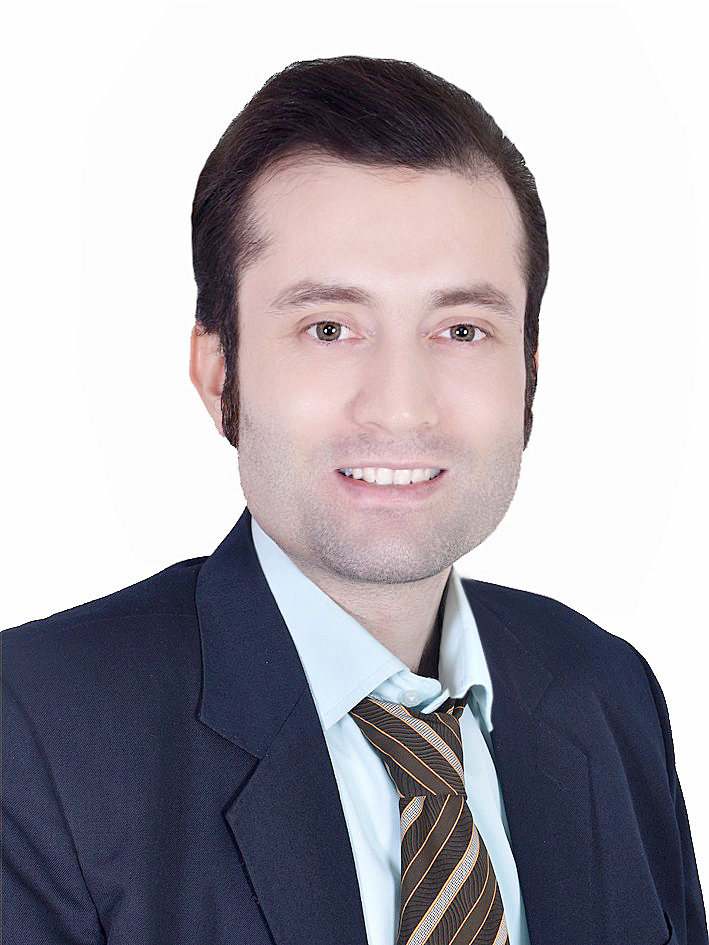}
\textbf{Sayyed Ahmad Naghavi Nozad} was born in Ghaen, South Khorasan Province, Iran, in 1988. He received the B.Sc. degree in Software Engineering from the Payame Noor University, Ghaen, Iran, in 2015, and the M.Sc. degree in Artificial Intelligence from the Amirkabir University of Technology, Tehran, Iran, in 2018. His research interests include but are not limited to Outlier Detection, Cluster Analysis, Community Detection, Novelty Detection, Information Theory, Image Processing, and Deep Learning.
\endbio

\pagebreak

\bio{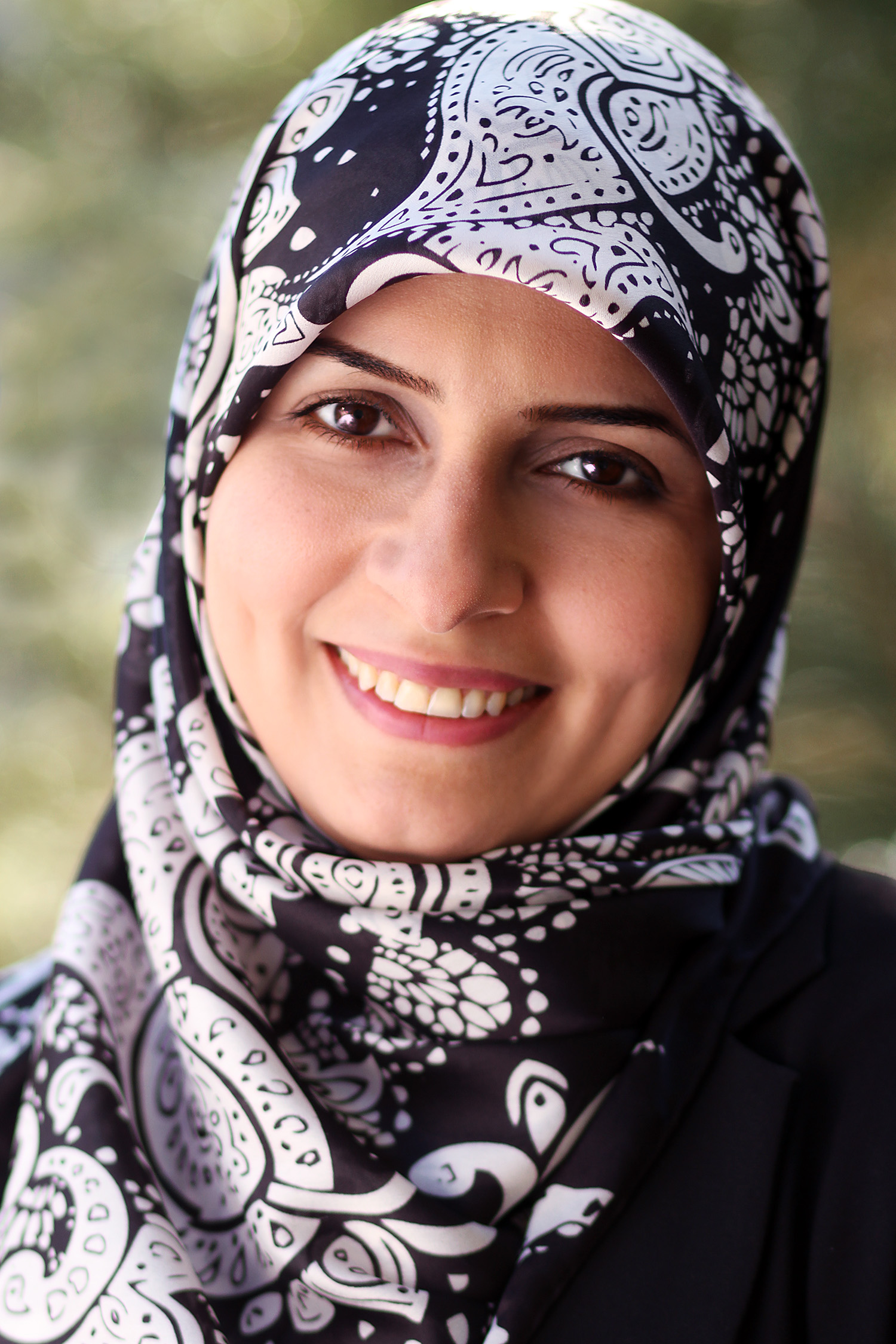}
\textbf{Maryam Amir Haeri} received her bachelor's degree in Software Engineering in 2007 and her master's degree in Information Technology in 2009, both from Sharif University of Technology, Tehran, Iran. She also received her doctorate in Artificial Intelligence from the Amirkabir University of Technology, Tehran, Iran, in 2014. She served as an assistant professor in the Computer Engineering Department, Amirkabir University of Technology, from September 2015 to January 2019. In addition, she was a research fellow at the Algorithm Accountability Lab at the University of Kaiserslautern, Germany, from February 2019 to October 2020. Since November 2020, she has been an assistant professor at the Learning, Data-Analytics and Technology Department, University of Twente, the Netherlands. Her research interests include Machine Learning, Big Data Analytics, Fairness in Machine Learning, and Complex Networks Analysis.
\endbio

%\pagebreak

\bio{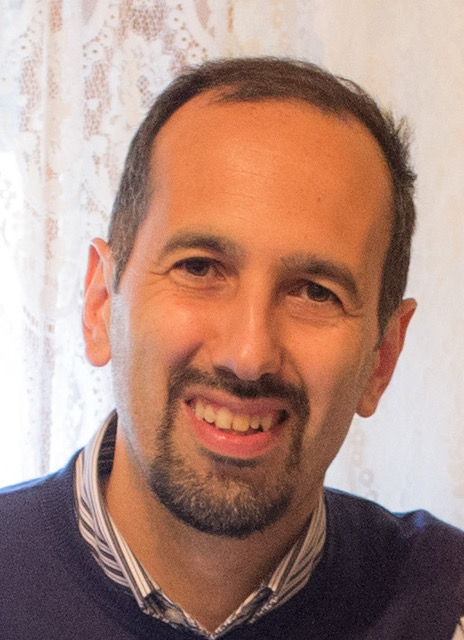}
\textbf{Gianluigi Folino} holds a Ph.D. in Physics, Mathematics, and Computer Science (Radboud University, Nijmegen, Holland, 2010). Since 2001, he works as a senior researcher at ICAR-CNR (the Institute of High Performance Computing and Networking of the Italian National Research Council). He is also a lecturer at the University of Calabria and the University Magna Graecia of Catanzaro. His research interests focus on applications of distributed computing in the area of data mining, bio-inspired algorithms (particularly genetic programming and swarm intelligence), big data, and bioinformatics. Within ICAR-CNR, he has been a contributor to several national and international research/industrial projects, and since 2013, he has been the coordinator of the project “Cyber Security – Digital and electronic payment services protection”. In addition, he was a visiting researcher at the University of Nottingham (United Kingdom) in 2007 and 2009, at Radboud University, Nijmegen (Netherlands), in 2008 and 2009, and finally at the University of California (UCLA), Los Angeles, in 2013.
\endbio

\end{document}